%% file: main.tex
\PassOptionsToPackage{table}{xcolor}
\documentclass[acmtog,nonacm]{acmart}
\acmSubmissionID{260}

\usepackage{booktabs} % For formal tables

\usepackage{mathtools}

\usepackage[ruled]{algorithm2e} % For algorithms
\usepackage{caption}
\usepackage{subcaption}

\SetAlFnt{\small}
\SetAlCapFnt{\small}
\SetAlCapNameFnt{\small}
\SetAlCapHSkip{0pt}

\usepackage{array}

\usepackage[table]{xcolor} % http://ctan.org/pkg/xcolor

\input{macros}

\begin{document}
% Title portion
\title{Bilateral Guided Radiance Field Processing}

% DO NOT ENTER AUTHOR INFORMATION FOR ANONYMOUS TECHNICAL PAPER SUBMISSIONS TO SIGGRAPH 2019!
%\author{Gang Zhou}
%\orcid{1234-5678-9012-3456}
%\affiliation{%
%  \institution{College of William and Mary}
%  \streetaddress{104 Jamestown Rd}
%  \city{Williamsburg}
%  \state{VA}
%  \postcode{23185}
%\email{gang_zhou@wm.edu}
%\affiliation{%
%  \institution{Inria Paris-Rocquencourt}
%\email{beranger@inria.fr}
%\affiliation{%
% \institution{Rajiv Gandhi University}
%\email{aprna_patel@rguhs.ac.in}
%\affiliation{%
%  \institution{Tsinghua University}
%  \streetaddress{30 Shuangqing Rd}
%  \country{China}
%}
%\email{chan0345@tsinghua.edu.cn}
%\author{Ting Yan}
%\affiliation{%
%  \institution{Eaton Innovation Center}
%  \city{Prague}
%  \country{Czech Republic}}
%\email{yanting02@gmail.com}
%\author{Tian He}
%\affiliation{%
%  \institution{University of Virginia}
%  \department{School of Engineering}
%  \city{Charlottesville}
%  \state{VA}
%  \postcode{22903}
%  \country{USA}
%}
%\affiliation{%
%  \institution{University of Minnesota}
%\email{tinghe@uva.edu}
%\author{Chengdu Huang}
%\author{John A. Stankovic}
%\author{Tarek F. Abdelzaher}
%\affiliation{%
%  \institution{University of Virginia}
%  \department{School of Engineering}
%  \city{Charlottesville}
%  \state{VA}
%  \postcode{22903}
%  \country{USA}
%}

%\renewcommand\shortauthors{Zhou, G. et al}

\author{Yuehao Wang}
\affiliation{%
 \institution{The Chinese University of Hong Kong}
 \city{Hong Kong}
 \country{China}}
\email{yhwang@link.cuhk.edu.hk}

\author{Chaoyi Wang}
\affiliation{%
 \institution{The Chinese University of Hong Kong}
 \city{Hong Kong}
 \country{China}}
\email{wc023@ie.cuhk.edu.hk}

\author{Bingchen Gong}
\affiliation{%
 \institution{The Chinese University of Hong Kong}
 \city{Hong Kong}
 \country{China}}
\email{gongbingchen@gmail.com}

\author{Tianfan Xue}
\affiliation{%
 \institution{The Chinese University of Hong Kong and Shanghai AI Laboratory}
 \city{Hong Kong}
 \country{China}}
\email{tfxue@ie.cuhk.edu.hk}

\renewcommand\shortauthors{Wang, Y. et al}

\begin{abstract}

Neural Radiance Fields (NeRF) achieves unprecedented performance in synthesizing novel view synthesis, utilizing multi-view consistency. When capturing multiple inputs, image signal processing (ISP) in modern cameras will independently enhance them, including exposure adjustment, color correction, local tone mapping, etc. While these processings greatly improve image quality, they often break the multi-view consistency assumption, leading to ``floaters'' in the reconstructed radiance fields. To address this concern without compromising visual aesthetics, we aim to first disentangle the enhancement by ISP at the NeRF training stage and re-apply user-desired enhancements to the reconstructed radiance fields at the finishing stage. Furthermore, to make the re-applied enhancements consistent between novel views, we need to perform imaging signal processing in 3D space (i.e. ``3D ISP''). For this goal, we adopt the bilateral grid, a locally-affine model, as a generalized representation of ISP processing. Specifically, we optimize per-view 3D bilateral grids with radiance fields to approximate the effects of camera pipelines for each input view. To achieve user-adjustable 3D finishing, we propose to learn a low-rank 4D bilateral grid from a given single view edit, lifting photo enhancements to the whole 3D scene. We demonstrate our approach can boost the visual quality of novel view synthesis by effectively removing floaters and performing enhancements from user retouching.
The source code and our data are available at: \clink{\url{https://bilarfpro.github.io}}.

\end{abstract}

\begin{CCSXML}
	<ccs2012>
	<concept>
	<concept_id>10010147.10010178.10010224.10010245.10010254</concept_id>
	<concept_desc>Computing methodologies~Reconstruction</concept_desc>
	<concept_significance>500</concept_significance>
	</concept>
	<concept>
	<concept_id>10010147.10010371.10010382.10010385</concept_id>
	<concept_desc>Computing methodologies~Image-based rendering</concept_desc>
	<concept_significance>300</concept_significance>
	</concept>
	<concept>
	<concept_id>10010147.10010371.10010396.10010401</concept_id>
	<concept_desc>Computing methodologies~Volumetric models</concept_desc>
	<concept_significance>300</concept_significance>
	</concept>
        <concept>
        <concept_id>10010147.10010371.10010382.10010236</concept_id>
        <concept_desc>Computing methodologies~Computational photography</concept_desc>
        <concept_significance>300</concept_significance>
        </concept>
	</ccs2012>
\end{CCSXML}

\ccsdesc[500]{Computing methodologies~Reconstruction}
\ccsdesc[300]{Computing methodologies~Image-based rendering}
\ccsdesc[300]{Computing methodologies~Volumetric models}
\ccsdesc[300]{Computing methodologies~Computational photography}


\keywords{Neural radiance fields, neural rendering, bilateral grid, 3D editing}

\begin{teaserfigure}
    \centering
    \includegraphics[width=\textwidth]{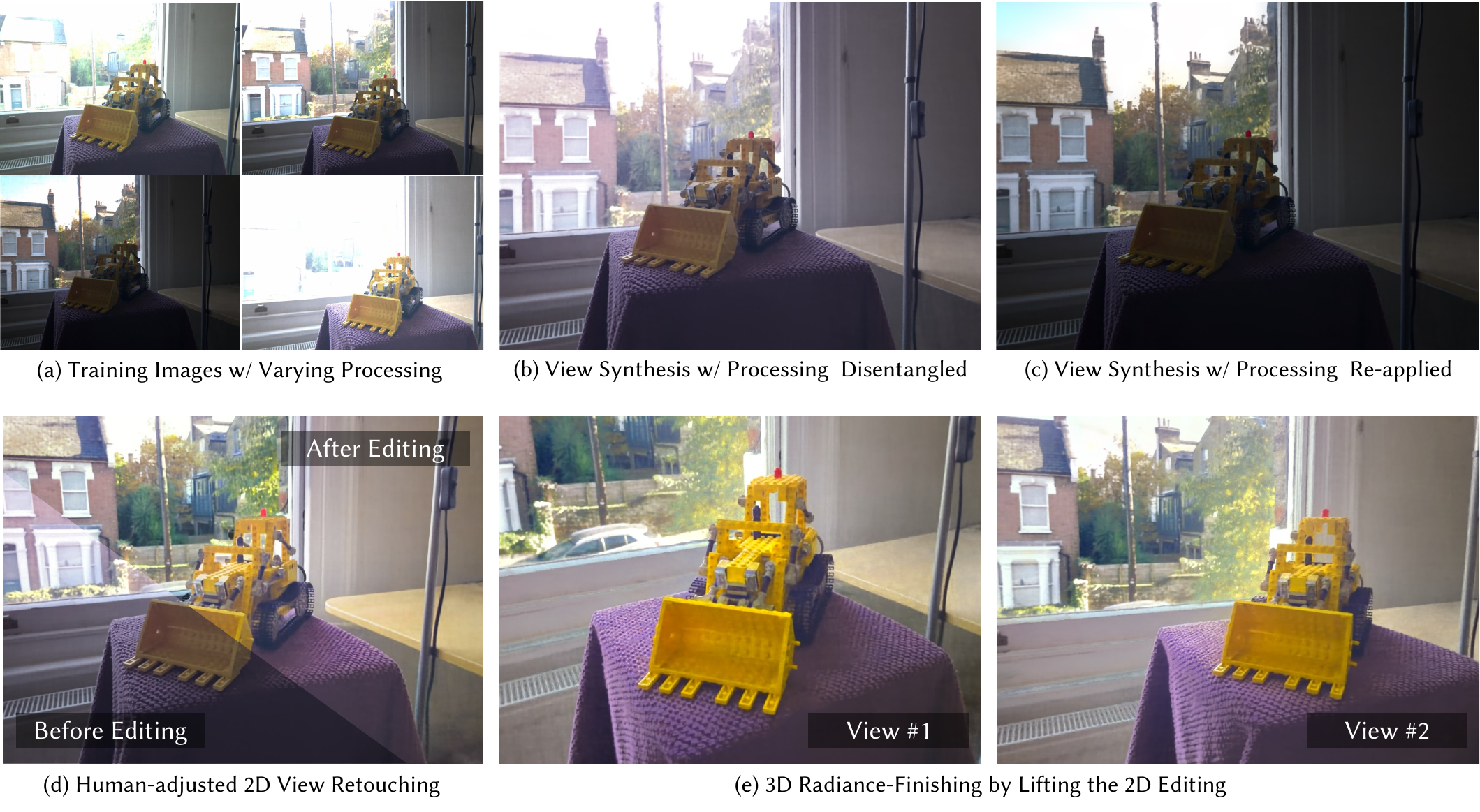}
    % \caption{In the NeRF training stage, our approach improves the novel view synthesis performance by effectively removing the ``floaters'' caused by variation in exposure and light configurations. Subsequently, in the finishing stage, given a 2D view editing, our method enables human-adjusted radiance field retouching at the 3D level, achieving compelling renditions consistently across different views. \tianfan{To make this teaser consistent with our story, I think it might be better to show 3 rendering: 1) Ours without 3D bilateral grid (I know the current one is already one without 3D bilateral grid, but I wonder whether we can a result with no any camera ISP. We can chat that in person.), 2) Ours with default 3D bilateral grid, and 3) Ours with a customized 3D bilateral grid. Also, can we find a better editing? This image does not look very beautiful, particularly the editing output.}}
    % \vspace{-0.3in}
    \caption{Given a set of multi-view images with photometric variation (a), such as varying exposure and local tone mapping, our method reconstructs a high-quality radiance field without ``floaters'' and fuses best-processed areas from training views (b) by disentangling the inconsistent camera pipeline processing for different views. After NeRF training with camera enhancements disentangled, we can re-apply the per-view processing (c). Furthermore, to enable view-consistent enhancements, we propose a radiance-finishing approach that can lift human-adjusted 2D view retouching (d) to 3D, achieving compelling renditions consistently over the entire scene (e).}
    % \tianfan{This looks much better. Still, for figure c, I feel it is not straight forward for a new readers to understand. A more intuitive is just to show image before/after processing (bilateral) in 3D. Figure 1c is actually a good example to for section 3.2. Let us chat in person.}
    \label{fig:teaser}
\end{teaserfigure}

\maketitle

\section{Introduction}
\input{sections/intro}

\section{Related Work}
\input{sections/related_work}

\section{Bilateral Guided Radiance Field Training}
\input{sections/bilanerf_training}

\section{Bilateral Guided Radiance Finishing}
\input{sections/bilanerf_finishing}

\section{Results}
\input{sections/experiments}

% \begin{itemize}
%     \item Evaluation: compares novel view synthesis quality
%     \item Qualitative Comparison: compares editing quality with image-level editing (w/ video consistency), instruct-nerf2nerf, style transfer (Photorealistic Style Transfer)
%     \item Ablation study: 1) Training: bilateral vs. spatial only, Small lr vs. Large lr, TV Loss vs No TV Loss 2) Editing: bilateral vs. spatial only, low-rank vs. direct 4D vs full rank, learn guidance map vs. gray map.
%     \item Applications: 1) Recoloring / Retouching 2) HDR fusion 3) Disentangling Varying Lights
% \end{itemize}

\section{Conclusion}

We present bilateral guided NeRF training and finishing, which improves the quality of novel view synthesis by disentangling photometric variation and lifting human-adjusted 2D retouching to 3D. Specifically, we devise the differentiable 3D bilateral grid to approximate ISP enhancements for each input image and propose the novel low-rank 4D bilateral grid to learn 3D-level manipulation from the given 2D view editing. During NeRF reconstruction, our approach performs best in novel view synthesis on our collected challenging scenes. Furthermore, our finishing pipeline allows users to retouch the entire scene by only editing a single view.

\begin{acks}
    We thank all anonymous reviewers for their constructive feedback and suggestions. We thank Ruikang Li for discussing image signal processing with us. We also thank Wang Wei, Xuejing Huang, and Chengkun Li for their assistance in making the video demo and capturing the dataset. This work is supported by CUHK Direct Grants (RCFUS) No. 4055189.
\end{acks}

\balance
\bibliographystyle{ACM-Reference-Format}
\bibliography{ref}

\appendix
\input{sections/appendix}

\end{document}

%% file: macros.tex
\newcommand{\clink}[1]{{\color{purple}#1}}

\newcommand{\rc}[0]{\cellcolor{red!30}}
\newcommand{\oc}[0]{\cellcolor{orange!30}}
\newcommand{\yc}[0]{\cellcolor{yellow!30}}

\makeatletter
\newcommand{\subalign}[1]{%
  \vcenter{%
    \Let@ \restore@math@cr \default@tag
    \baselineskip\fontdimen10 \scriptfont\tw@
    \advance\baselineskip\fontdimen12 \scriptfont\tw@
    \lineskip\thr@@\fontdimen8 \scriptfont\thr@@
    \lineskiplimit\lineskip
    \ialign{\hfil$\m@th\scriptstyle##$&$\m@th\scriptstyle{}##$\hfil\crcr
      #1\crcr
    }%
  }%
}
\makeatother

%% file: sections/intro.tex
% Neural Radiance Fields (NeRF) \cite{mildenhall2021nerf} is trained on multi-view images processed by camera pipelines, using volumetric ray tracing as the imaging process. This simplistic setting overlooks the complex image signal processing (ISP) pipelines in cameras, which results in the performance degeneration of NeRF in many challenging scenarios. When capturing multi-view images, digital cameras, particularly those in cell phones, will independently process each view with adjustments in exposure time, tone mapping, exposure, etc. This enhances photographs for better visual quality but also introduces color inconsistencies across input views that cannot be encoded by viewing directions, leading to ``floaters'' as mentioned in previous work \cite{martin2021nerf,barron2022mip,sabour2023robustnerf}. Figure \ref{fig:inconsistent-floaters} illustrates this issue in our scene captures at nighttime.
%  cannot be encoded by viewing directions
% This issue is exacerbated in those capture scenarios where light configurations change in different views, e.g., scenes with lamps at nighttime.

\begin{figure}[t]
    \centering
    \begin{subfigure}[a]{\linewidth}
        \centering
        \vspace{0.2in}
        \includegraphics[width=\textwidth]{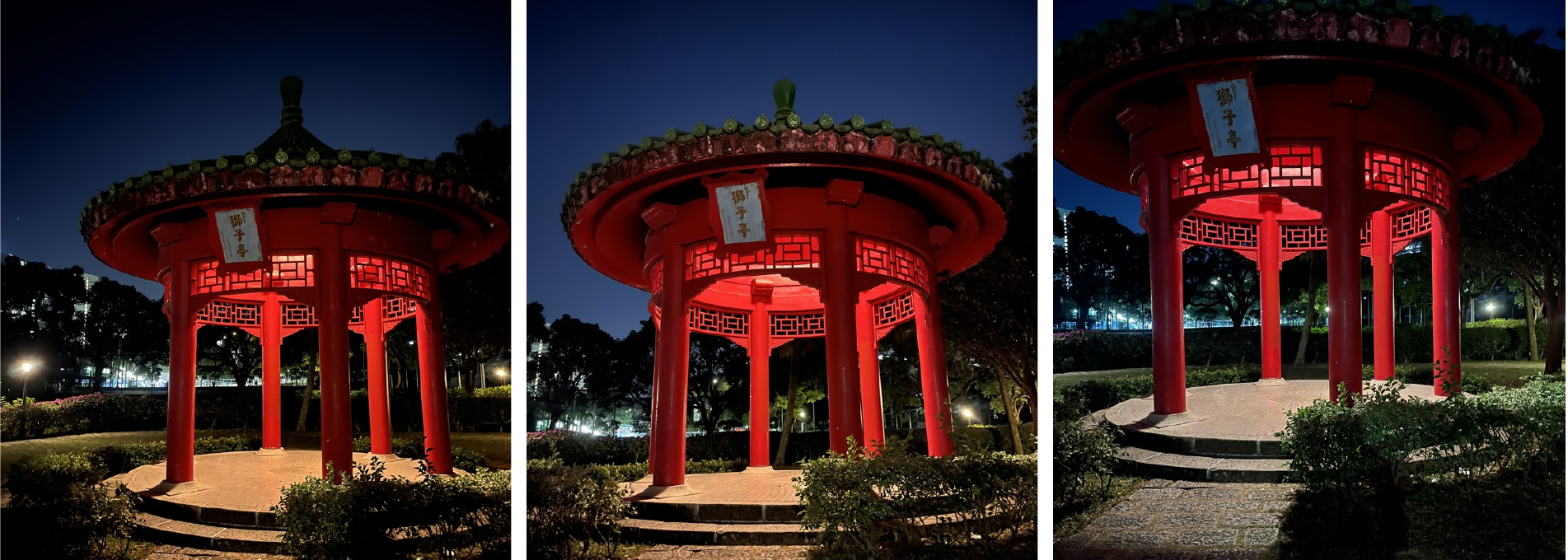}
        \caption{We exhibit three multi-view images captured by cell phone cameras. We do not manually change camera parameters but inconsistencies across different views still appear in the brightness of the sky, the color of the floor, and the hue of the background lights.}
         % \tianfan{I wonder we can better visualize the difference between these three examples due to ISP. Like can we find a line in this image and plot the intensity changes between 3 different views?}
        \label{fig:inconsistent-captures}
    \end{subfigure}
    \begin{subfigure}[a]{\linewidth}
        \centering
        \vspace{0.15in}
        \includegraphics[width=\textwidth]{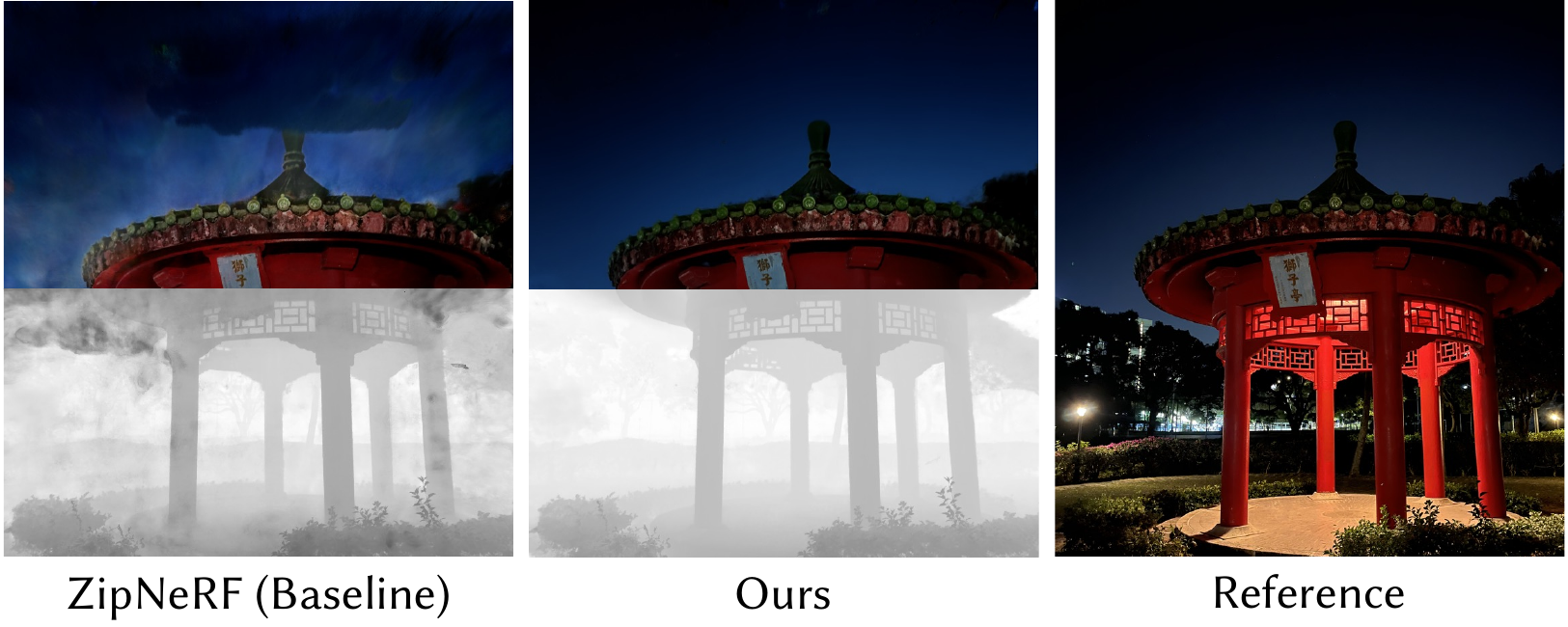}
        \caption{Due to these inconsistencies, the baseline method ZipNeRF \cite{barron2023zipnerf} yields floaters and a strange appearance in the sky, while our method can overcome this issue.}
        % \tianfan{The floater arfiacts are not super obvious in ZipNeRF. Can we find a better patch.}
        \label{fig:floaters}
    \end{subfigure}
    \vspace{-0.1in}
    \caption{Inconsistent camera processing of multi-view images leads to artifacts in novel view synthesis.}
    \label{fig:inconsistent-floaters}
    % \vspace{-0.3in}
\end{figure}

Neural Radiance Fields (NeRF) \cite{mildenhall2021nerf} has demonstrated remarkable performance in novel view synthesis using volumetric ray tracing. These fields are reconstructed from multi-view input images taken by cameras, based on multi-view consistency assumption. However, this basic approach ignores in-camera image signal processing (ISP) that enhances captured images. When capturing multi-view images, modern digital cameras, particularly smartphones, will choose the best processing strategy for each input view, such as exposure, color correction, tone mapping, etc. These enhancements improve visual quality but also introduce brightness or color inconsistency across different input views. Most NeRF models cannot handle them, leading to ``floaters'' as mentioned in previous work \cite{martin2021nerf,barron2022mip,sabour2023robustnerf}. Figure \ref{fig:inconsistent-floaters} demonstrates this problem with nighttime captures. 

Training NeRF on raw images~\cite{mildenhall2022nerf} could eliminate the impacts of ISP and recover radiance fields in linear raw space. However, many cameras do not provide raw sensor data due to either hardware or software limitations, and raw data itself also entails high storage costs. Previous methods adopt GLO vectors~\cite{martin2021nerf} to approximate the multi-view inconsistencies, but it has been shown that this approach degrades the rendering quality~\cite{barron2022mip}. A more principled solution involves simulating the per-view ISP processing after volumetric rendering during the NeRF training. This decouples ISP enhancements from NeRF rendering: per-view enhancements are modeled by simulated ISP, allowing core NeRF training to concentrate on unprocessed images, thereby better preserving multi-view consistency. Nonetheless, ISP pipelines vary significantly among diverse devices and typically involve non-linear and local operations, making accurate simulation challenging. Therefore, to effectively disentangle variations in the input views, we need to design a simple and learnable representation that can closely approximate ISP enhancements for each view.

% In addition to simulating camera pipelines to facilitate NeRF reconstruction.
% In this regard, NeRF-finishing is demanded, which allows for human-adjusted retouching to enhance visual aesthetics.
Also, after disentangling ISP enhancements and core NeRF training, the rendered view by the reconstructed NeRF may be visually less pleasing, as per-view the visual enhancements are removed. Therefore, in addition to removing per-view ISP enhancements during NeRF training, our objective is also to introduce NeRF-finishing, which enables further human-adjusted retouching on recovered NeRFs. For instance, users may need to retouch the color tone of the whole scene, enhance the brightness of the subject, darken the background, and increase the saturation of the sky or vegetation. Such manipulations on photographs have been well developed and integrated into commercial software, e.g., Adobe Lightroom\textregistered. As for 3D scene editing, even those basic steps, like selecting edit areas and point operations, pose challenges. Furthermore, individually processing each view will bring about flickering when changing the viewpoint. To ensure consistent novel view synthesis after editing, 3D-level imaging signal processing (``3D ISP'') is required. 
% To enhance the display of radiance fields, RawNeRF \cite{mildenhall2022nerf} applies photo-finishing algorithms such as HDRNet \cite{gharbi2017deep} and HDR+ \cite{hasinoff2016burst} to the synthesized views.

% Our goal is to seek a unified operator that simulates ISP in the NeRF training stage and performs 3D-level manipulation of the reconstructed scenes in the finishing stage. With respect to this, we resort to the bilateral grid \cite{chen2007real}, an efficient data structure storing coefficients of local affine models in bilateral space. It has been adopted to approximate non-linear, local, and edge-aware photographic transformations in previous work \cite{chen2016bilateral,gharbi2017deep}. Inspired by this, we employ differentiable bilateral grids for radiance field processing. First, we jointly optimize per-view 3D bilateral grids with NeRF to represent the ISP enhancements on input images. Second, considering plenty of photo editing and enhancement tools are available, we propose to learn a low-rank 4D bilateral grid from a single view edit. After a short-time optimization, the 4D bilateral grid can point-wisely transform the appearance in 3D space to achieve user-desired retouching given in the edited view. We also study the tailored training strategies and regularizations for optimizing bilateral grids with NeRF.

Our goal is to seek a unified operator that simulates ISP in the NeRF training stage and performs ISP-like enhancement of the reconstructed scenes in the finishing stage.
With respect to this, we resort to the bilateral grid \cite{chen2007real}, an efficient image operator that can approximate non-linear, local, and edge-aware photographic transformations~\cite{chen2016bilateral,gharbi2017deep}. As demonstrated in~\cite{gharbi2017deep}, bilateral grid can approximate the majority of ISP operations. Inspired by this, we employ differentiable bilateral grids for radiance field processing. First, to reduce multi-view inconsistency introduced by ISP processing, we jointly optimize NeRF with per-view 3D bilateral grids, which represent the ISP enhancements. Second, to model the ISP-like radiance finishing, we devise 4D bilateral grid and use it to manipulate radiance fields. To obtain the coefficients in the 4D bilateral grid, we first ask users to edit a selected input view, using existing photo editing software or enhancement tools, and lift this 2D editing to 3D space using a novel low-rank 4D bilateral grid. After a short-time optimization, the 4D bilateral grid can point-wisely transform the appearance in 3D space to achieve user-desired retouching specified by the edited view. We also study the tailored training strategies and regularizations for optimizing bilateral grids with NeRF. The overview pipeline of our method is present in Figure \ref{fig:pipeline-figure}.

We demonstrate the effectiveness of our method on multiple scenes captured at nighttime, RawNeRF dataset \cite{mildenhall2022nerf}, and the mip-NeRF 360 dataset \cite{barron2022mip}. Our bilateral guided camera pipeline approximation in the training stage is shown to effectively disentangle ISP enhancements and achieve state-of-the-art performance in novel view synthesis. During the bilateral guided radiance-finishing stage, our method allows users to retouch the whole scene by editing a single view image, unlocking 3D-level brightness adjustments, local curve, recoloring, etc.

% Prior work by \citet{fan2022unified} and \citet{zhang2022arf} transfers 2D image styles to 3D scenes by optimizing radiance fields to match the reference styles.

% the imaging process in NeRF lacks 3D-level editable controls in brightness, local curve, tone mapping, etc. This hinders users from producing their desired finishing effects. To this end, our goal is to append a radiance processing pipeline to disentangle inconsistent effects of ISP in NeRF reconstruction and empower users with the flexibility to enhance and retouch the scenes.

% Training NeRF on raw images \cite{mildenhall2022nerf} covers high dynamic range (HDR) but is costly in storage. Resembling the classical HDR approach \cite{debevec2023recovering} for photographs, a possible solution is to reconstruct radiance fields in HDR from multi-view LDR images under different exposures. This requires a simulation of ISP to inverse sRGB to linear raw space. \citet{huang2022hdr} simplify ISP as per-channel tone mappers parameterized by MLPs. 
% Prior work \citet{huang2022hdr} uses MLPs to parameterize ISP as per-channel tone mappers.

% \citep{mildenhall2022nerf} enable NeRF training on raw images. \citep{huang2022hdr} further adopts MLPs to simulate per-channel tone mapper.

%% file: sections/related_work.tex
We first briefly review the literature on ISP and image enhancements. Next, we will discuss the recent advances in NeRF appearance editing. Even these two topics are quite different, our work bridges them through bilateral guided processing.

% \tianfan{Related work is too long. Let us try to shrink it later. Also, add 1-2 references to work that can invert camera pipeline, like "Learning srgb-to-raw-rgb de-rendering with content-aware metadata" or "Unprocessing Images for Learned Raw Denoising".}

\subsection{Image Signal Processing and Bilateral Grid}
\label{sec:related_work_isp}

We first introduce Image Signal Processing (ISP) pipeline. With an input raw linear image, ISP first transforms it to the linear sRGB space through normalization, denoising, white balancing, demosaicking, and color transformation. After that, the sRGB gamma correction curve is applied to enhance the darker areas. Advanced ISP pipelines usually process images with more complex steps. Exposure fusion \cite{mertens2007exposure} blends the underexposed and overexposed frames to achieve higher dynamic range, HDR+ \cite{hasinoff2016burst} uses local tone mapping to squeeze a high-dynamic capture into a low-dynamic-range image, and Laplacian pyramids \citet{paris2011local} achieve edge-preserving contrast enhancement. \citet{bychkovsky2011learning} pioneer in predicting adjustments from a large training set to automate image enhancements.

% In practice, ISP pipelines usually present nonlinear local processing and exhibit remarkable variability across different cameras. This brings difficulties in developing a unified model to represent ISP. 
In the meanwhile, there are several attempts to model existing commercial ISP pipelines. One line of work is to invert sRGB images back to raw linear images, either by a simplified ISP pipeline~\cite{brooks2019unprocessing} or a trained network~\cite{nam2022learning}. Another line is to tune ISP parameters or the entire ISP. \citet{nishimura2018automatic} explore data-driven parameter tuning of a conventional ISP, \citet{tseng2019hyperparameter} attempt to optimize ISP parameters via differentiable proxies, and \citet{yu2021reconfigisp} try to search for optimal ISP configurations. Recently, \citet{conde2022model} learn a bidirectional mapping between raw and RGB space via dictionary representations, and \citet{tseng2022neural} propose a neural camera pipeline by using a set of proxy networks for individual processing steps. \citet{wang2023implicit} adopt implicit neural fields to reconstruct an all-in-focus HDR image from a sparse Time-Aperture-Focus stack, providing post-processing control over focus, aperture, and exposure. Despite all these efforts, ISP pipelines usually present nonlinear local processing and exhibit remarkable variability across different cameras, making it difficult to develop a unified model to represent ISP. 
% These methods require network training on extensive datasets and do not possess high performance for our scenario.

For efficient ISP processing, bilateral grid \cite{chen2007real} is first introduced to accelerate bilateral filtering \cite{tomasi1998bilateral,durand2002fast}. Due to its efficiency and edge-preserving properties, the bilateral grid has been widely used in low-level vision tasks. \citet{barron2015fast} recover stereo depth maps in bilateral space. \citet{barron2016fast} further extends this approach to encompass colorization, depth super-resolution, and semantic segmentation. Inspired by joint bilateral upsampling \cite{kopf2007joint} and guided filter \cite{he2012guided}, \citet{chen2016bilateral} fit a local affine model between downsampled input and guided images in a bilateral grid. \citet{gharbi2017deep} then incorporate this idea into deep learning fashion to accelerate different image enhancements. These works show the capability of the bilateral grid in approximating a wide range image processing operations.
% They design a neural network architecture to predict coefficients of bilateral grids from input images, enabling content-based automatic photo enhancements.
% Building on the success illustrated by prior work in utilizing the bilateral grid for image processing, we adopt it as our proxy to approximate the camera finishing pipeline.

\subsection{NeRF Appearance Editing}

Neural scene representations, like NeRF, have emerged as versatile mediums for representing 3D scenes and objects. To modify the appearance of 3D objects reconstructed from images, \cite{srinivasan2021nerv,zhang2021nerfactor,verbin2022ref,boss2022samurai,munkberg2022extracting,jin2023tensoir} enables material editing and relighting via joint optimization of PBR materials, 3D shape, and lighting from multi-view captures.

Another line of work emulates image/video manipulation in NeRF. INS~\cite{fan2022unified}, Arf \cite{zhang2022arf}, and StyleRF \cite{liu2023stylerf} transfer 2D image styles to 3D scenes by optimizing radiance fields to match the reference styles. PaletteNeRF \cite{kuang2023palettenerf} and RecolorNeRF \cite{gong2023recolornerf} introduce palette-based NeRF editing, empowering users to edit the scene via color pickers.
% NeuMesh \cite{yang2022neumesh} introduces a solution to appearance editing of neural implicit fields via texture swapping and painting. Seal-3D \cite{wang2023seal} further presents an interactive NeRF editing system with support for pixel-level manipulation. 
\citet{gong2023seamlessnerf} perform seamless appearance blending of part NeRFs, analogous to Poisson seamless editing on images \cite{perez2023poisson}.
% resembling ``Color Mixer'' in Adobe Lightroom\textregistered.

Many previous methods \cite{martin2021nerf,jang2021codenerf,liu2021editing,tancik2022block,park2021nerfies,barron2022mip} use GLO vectors \cite{bojanowski2017optimizing}, a.k.a. appearance embeddings, a.k.a. latent codes, to modulate variation in appearance. This technique is first incorporated into NeRF by \cite{martin2021nerf}, aiming to train NeRF on in-the-wild photographs. 
% The GLO vectors can be trained to encode the inconsistent elements in the training images. Additionally, there are some variants of GLO. ZipNeRF \cite{barron2023zipnerf} proposes affine generative latent optimization, where GLO is mapped to affine transformations of the NeRF bottleneck vectors, akin to AdaIn \cite{huang2017arbitrary}. \citet{cheng2023uc} predict a per-view color-correction affine transformation using GLO.
Previous work also utilizes pre-trained models to facilitate NeRF editing. CLIP-NeRF \cite{wang2022clip} explores text-and-image driven manipulations by mapping CLIP \cite{radford2021learning} features into NeRF appearance and shape code. NeRF-Art \cite{wang2023nerf} proposes text-driven NeRF stylization, which is achieved by optimizing losses in the CLIP embedding space. \citet{kobayashi2022decomposing} distill LSeg \cite{li2022languagedriven} and DINO \cite{caron2021emerging} features into 3D fields, allowing for selective editing through a text or image patch. Instruct-NeRF2NeRF \cite{instructnerf2023} conducts NeRF editing with text instructions, which leverages an image-conditioned diffusion model~\cite{brooks2023instructpix2pix} to iteratively edit images in the training set while optimizing the underlying scene.

%% file: sections/bilanerf_training.tex
\begin{figure*}[t]
	\centering
	\includegraphics[width=0.95\textwidth]{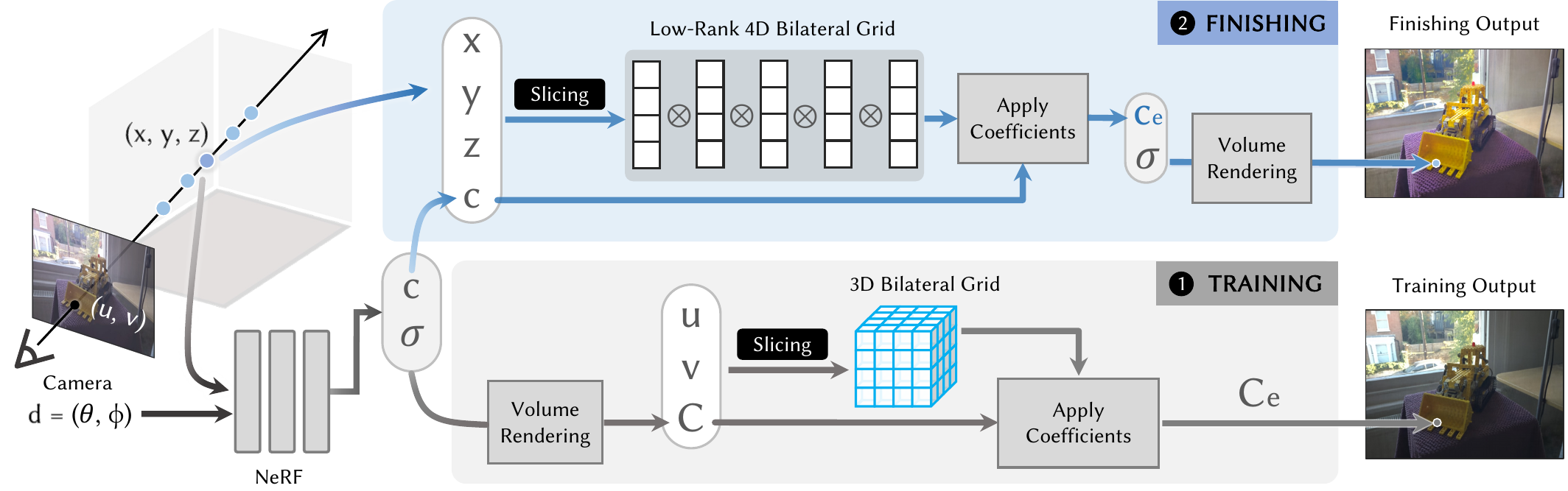}
	%\vspace{-10px}
	\caption{Pipeline of our proposed method. Our approach consists of two stages: 1) In the training stage, we use the 3D bilateral grid to approximate view-dependent camera enhancements on the rendering results; 2) In the finishing stage, we slice the low-rank 4D bilateral grid to apply 3D-level enhancements.}
    % \tianfan{1) This diagram is mostly consists of symbols and blocks, which is less interesting and take some efforts to follow. Maybe to replace a few blocks by actual images/3D. For example, we can use a real 3D point cloud to replace the 3D volume, a 2D rendering image to replace both ce in training and ce in finishing. 2) Our pipeline actually consists two steps, the first step is training and the second one is finishing in rendering. This is not super clear in this diagram right now. Let us chat in person.}
	\label{fig:pipeline-figure}
	% \vspace{-5px}
\end{figure*}

\vspace{10px}
\subsection{Preliminary: Neural Radiance Fields}

We first overview the NeRF pipeline and introduce essential notations. Given a set of multi-view images, NeRF optimizes a volumetric scene representation by minimizing rendering loss. Specifically, NeRF establishes a mapping $F_\Theta(\boldsymbol{x}, \boldsymbol{d}) \rightarrow (c, \sigma)$, where $\boldsymbol{x}$ is a spatial coordinate in 3D space, $\boldsymbol{d}$ is the viewing direction $(\theta, \phi)$, $c$ is the predicted RGB color, and $\sigma$ is the predicted density. Through the scene representation $F_\Theta$, we can perform volumetric ray tracing to synthesize views. For each pixel at $(u, v)$, we cast a ray towards the pixel and the color $\widehat{C}(u, v)$ is evaluated via quadrature (Equation \ref{eqn:quadrature}):
\begin{equation}
    \widehat{C}(u, v) = \sum^N_{j=1} T_j \big(1-\exp(-\sigma_j \Delta_j ) \big) c_j,
    ~~\quad~~T_j = \exp\bigg(-\sum_{i=1}^{j-1}\sigma_i \Delta_i \bigg),
\label{eqn:quadrature}
\end{equation}
where density $\sigma_j$, color $c_j$, and transmittance $T_j$ are sampled
along the ray with intervals $\Delta_j$. In the training stage, the rendering loss for optimizing the parameters $\Theta$ is given in Equation \ref{eqn:render_loss}:
\begin{equation}
    \mathcal{L} = \sum_{(u,v) \in \mathcal{S}} \left \| ~~\widehat{C}(u,v) - C(u,v)~~ \right \|_2^2,
    \label{eqn:render_loss}
\end{equation}
where $\mathcal{S}$ is a training batch with randomly sampled pixels from different views and $C(u,v)$ is the ground truth color in training views. As the entire process is differentiable, the gradients from this rendering loss can be back-propagated to NeRF parameters $\Theta$.
% ~\tianfan{I call it ``the radiance field parameter set''. Any better name here?}
% \tianfan{Question for a notation: why we use capital $C(u, v)$, instead of $c$?}

There are many ways to model the scene representation $F_\Theta(\boldsymbol{x}, \boldsymbol{d})$. The na\"ive NeRF \cite{mildenhall2021nerf} parameterizes $F_\Theta$ as an MLP with positional encoding \cite{vaswani2017attention,tancik2020fourier}. Another seminal work Instant-NGP \cite{muller2022instant} adopts multi-resolution hash encoding to achieve much faster training speed. Both representations fit our bilateral guided processing.
% \tianfan{Say something like our solution can adapt to any NeRF representation (if that is true).}

% To tackle the input images with distractors, e.g., varying exposure, NeRF-W \cite{martin2021nerf} augments the scene representation as $F_\Theta(\boldsymbol{x}, \boldsymbol{d}, \boldsymbol{\psi}_k)$, where $\boldsymbol{\psi}_k$ is the GLO vector for the $k$th view.

\subsection{Differentiable 3D Bilateral Grid}

% The imaging process of NeRF is volume rendering without considering camera pipelines. Our goal is to append a post-processing $g(\cdot)$ after the volume rendering to simulate ISP enhancements on the rendering images. This post-processing should be differentiable, efficient, and generalizable for diverse image processing. As demonstrated by HDRNet \cite{gharbi2017deep}, the bilateral grid \cite{chen2007real} can approximate various image enhancements and is compatible with neural networks due to its differentiability. Motivated by this, we choose it as a learnable proxy of ISP in NeRF.

Default NeRF training does not consider the imaging process and simply assumes that any ISP processing can be baked into the 3D radiance field. While this is true if a simple, global, and identical processing is applied to all input views, more complex and view-dependent processing cannot be modeled by a 3D radiance field, resulting in floater artifacts shown in Figure~\ref{fig:floaters}.

Therefore, we propose to append a post-processing after the volume rendering to simulate ISP enhancements on the rendered images. This post-processing should be differentiable, efficient, and generalizable for diverse image processing tasks. As demonstrated by HDRNet \cite{gharbi2017deep}, the bilateral grid \cite{chen2007real} can approximate various image enhancements and is compatible with neural networks due to its differentiability. Motivated by this, we choose it as a learnable proxy for ISP in NeRF.

The bilateral grid is defined as a four-dimensional tensor $\boldsymbol{A} \in \mathbb{R}^{W \times H \times M \times 12}$, where $W$, $H$ and $M$ determine the grid resolution, and $12$ corresponds to the length of a flattened $3 \times 4$ affine color transformations. To apply the bilateral grid $\boldsymbol{A}$ to a rendered image, a \textit{slicing} operation \cite{chen2007real} is performed per pixel to retrieve color transformations for the given image. Specifically, for a pixel at $(u, v) \in [0, 1]^2$ (normalized into a unit square) with RGB value $C = [C_r, C_g, C_b] \in [0, 1]^3$, slicing can be written as:
\begin{equation}
\begin{split}
    &\boldsymbol{A}\left[u, v, C \right] = \sum_{i,j,k} \kappa_{i,j,k}\bigg(u, v, g(C)\bigg) \boldsymbol{A}_{i,j,k}~~, \\
    &\kappa_{i,j,k}(u, v, w) = \Lambda\big(W \cdot u - i\big) \Lambda\big(H \cdot v - j\big) \Lambda\big(M \cdot w - k\big)~~,
\end{split}
\label{eqn:bilagrid_slice}
\end{equation}
where $\Lambda(t) = \max \left(1 - |t|, 0 \right)$ is the ``hat'' function used as the linear interpolation kernel, $g(\cdot)$ is a guidance function mapping RGB colors to scalars in $[0, 1]$, $\boldsymbol{A}_{i,j,k}$ denotes the tensor element indexed at $(i,j,k)$. After slicing, we can reshape $\boldsymbol{A}[u, v, C] \in \mathbb{R}^{12}$ to an affine transformation matrix $\boldsymbol{\mathcal{A}} \in \mathbb{R}^{3\times 4}$ and multiply it with the pixel color to yield the processed color $C_{e} = \boldsymbol{\mathcal{A}}[C~|~1]^\top$. For the guidance function $g(\cdot)$, we simply formulate it as luminance, similar to \cite{chen2016bilateral}, which is shown to be effective for most of our scenarios. The resolution of the bilateral grid is much smaller than the one of the input image (in the experiment, we use grid resolutions varying from $8\times 8$ to $32 \times 32$). This not only reduces the computational costs but also prevents the bilateral grid from encoding high-frequency content of the input image in the translation components.

% It is noteworthy that the slicing operation is sub-differentiable. Technically, the derivative of $\Lambda(t)$ is indeterminate when $t = -1$, $t = 1$ or $t=0$. To propagate gradients everywhere to $\boldsymbol{A}_{i,j,k}$ and $C$, we specify $\nabla_t\Lambda(0) = 1$ and $\nabla_t\Lambda(1) = \nabla_t\Lambda(-1) = 0$. \tianfan{if running out of space, this paragraph can be removed, as I think many people should know that slicing is differentiable.}

% \tianfan{Add 1-2 sentences (or a short paragraph) here to highlight that 3D bilateral grid can be used to approximate ISP enhancements, as illustrated in HDRnet paper. This also motivate our bilateral solution.}

\begin{figure}[t]
    \centering
    \centering
    \includegraphics[width=\linewidth]{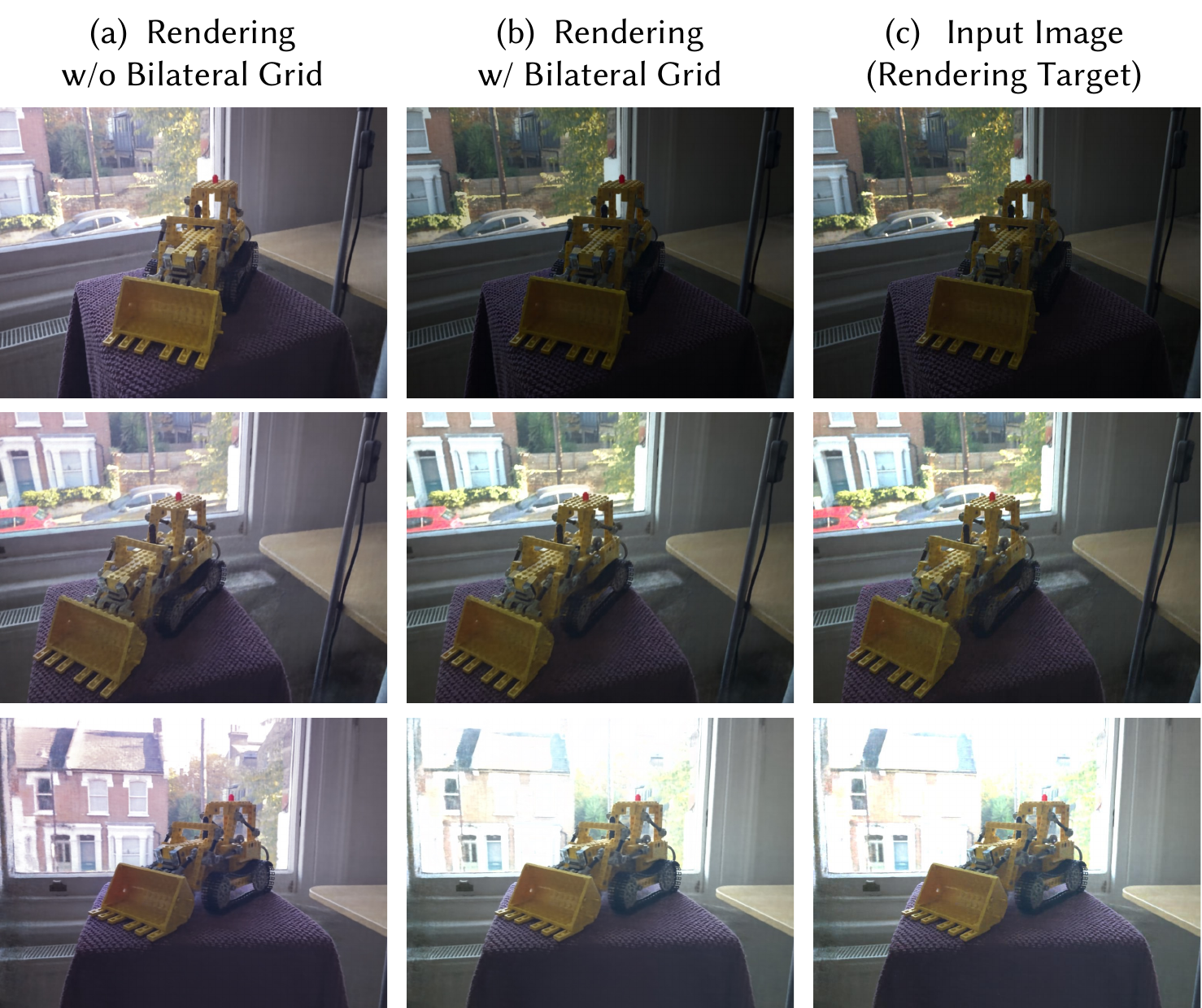}
    \caption{
    Illustration of how 3D bilateral grids guide NeRF training. Directly rendered images (a) from the radiance field do not incorporate per-view camera enhancement and do not match the input images (rendering target). With a per-view bilateral grid applied, the rendered images (b) can reproduce camera enhancement and are almost identical to input images (c).
    % \tianfan{Updated. Please check.}
    % Results of our bilateral guided NeRF training. After enhancements and adjustments of input images are represented in per-view 3D bilateral grids during training, we can synthesize views with consistent appearance (a). Applying the optimized 3D bilateral grid back to the rendering (b) can reproduce the input image (c).
    }
    % \vspace{-10px}
    % \tianfan{1) Move (a) to the right, as in this figure, it actually serves as a ground truth. 2) Can we name these three columns like: "Rendering
    % w/o bilateral grid", "Rendering with Bilateral Grid", "Rendering target" (or some similar name)?}
    \label{fig:bilanerf_training}
\end{figure}

\subsection{Optimize 3D Bilateral Grids with Radiance Fields.}
% \tianfan{Make this paragraph into a subsection.}
In the NeRF training stage, a bilateral grid will be assigned to each training view, used for disentangling variation caused by camera pipelines. After training, we can directly render NeRF without those bilateral grids. Suppose there are $L$ training images, we initialize $L$ bilateral grids $\{\boldsymbol{A}^1, \boldsymbol{A}^2, \dots, \boldsymbol{A}^L \}$, with each cell in the grid storing coefficients of identity affine transformations. For each training pixel at $(u, v)$ in the $l$-th training view, we first evaluate the color $\widehat{C}(u,v)$ via volume rendering, then slice and apply $\boldsymbol{A}^l[u,v,\widehat{C}(u,v)]$ to obtain the processed color $\widehat{C}_{e}(u,v)$. During training, we add a \textit{total variation} (TV) term to regularize the smoothness of bilateral grids. The final objective is:
\begin{equation}
\begin{split}
    \mathcal{L} &= \sum_{(u,v) \in \mathcal{S}} \left \| ~~\widehat{C}_{e}(u,v) - C(u,v)~~ \right \|_2^2 + \lambda_{TV} \mathcal{L}_{TV}~~, \\
    \mathcal{L}_{TV} &= \sum_{i,j,k,l} \frac{1}{|\boldsymbol{A}^l|} \left\|\Delta_x\boldsymbol{A}^l_{i,j,k}\right \|_2^2 + \left\|\Delta_y\boldsymbol{A}^l_{i,j,k}\right \|_2^2 + \left\|\Delta_z\boldsymbol{A}^l_{i,j,k}\right \|_2^2~~,
\end{split}
\label{eqn:bilagrid_nerf_loss}
\end{equation}
where $\Delta$ is the finite difference operator. We find large $\lambda_{TV}$ is the key to the success of bilateral grid optimization. In all of our experiments, we generally set $\lambda_{TV}=10$.
% To avoid bilateral grids from distorting the rendered pixels to undesired colors, the learning rate of bilateral grids is set to $\times 10$ smaller than the learning rate of radiance fields.
% Although the differentiable 3D bilateral grid can be optimized, inversing the whole ISP pipeline remains challenging. Our goal is not to map the processed images back to the linear space. 

% \tianfan{I made a lot of change to next two paragraphs. Please check.}
As discussed in Section~\ref{sec:related_work_isp}, inversing the whole ISP pipeline is challenging, but our goal is not to map the processed images back to the linear space. 
Instead, like previous NeRF training, we assume that radiance field can bake an average processing, and the proposed 3D bilateral grid only need to model additional per-view processing on top of the average one, making the grid optimization feasible.
% Therefore, the proposed optimization target in Equation~\ref{eqn:bilagrid_nerf_loss}.

Figure \ref{fig:bilanerf_training} illustrates that bilateral grid is capable of modeling view-dependent camera processing. In this example, the input views (a) have large brightness variations caused by inconsistent camera processing. Rendered images by the radiance field without a bilateral grid cannot model these large differences caused by cameras, and thus may result in large training errors when matching the input images (c). The learned per-view bilateral grids can perfectly close this gap, generating images (b) that are very close to the input images, and thus reduce the impacts of inconsistent camera processing over the NeRF training.
% Instead, we only aim to find a shared intermediate with basic processing across different views, which is sufficient for our task. We present the rendering results with and without bilateral grids as well as the reference input images in Figure \ref{fig:bilanerf_training}.

% \tianfan{I think it worths to show one figure, to illustrate how bilateral grid close the gap between different views. Let us chat in person.}

% Although the differentiable 3D bilateral grid can be optimized by gradient-based optimizers, the problem itself is underdetermined. We can consider solving $\boldsymbol{A}X = C$ in each cell of the bilateral grid, where both $\boldsymbol{A}$ (transformation) and $X$ (color) are unknown. Suppose $\boldsymbol{A}=\boldsymbol{A}^\ast, X=X^\ast$ is a solution, then $\boldsymbol{A}=\boldsymbol{A}^\ast/s, X=sX^\ast$ is also a solution. When $s$ becomes excessively large/small, $X$ is subjected to over-distortion, leading to unrealistic color. In our solution to this issue, we down-scale the learning rate of the bilateral grid to $\times 0.1$ of the NeRF's learning rate. By doing so, the coefficients in the bilateral grid undergo a slow optimization, starting from identity. Since the processing differences across training views will not be significant in most cases, the downscaled learning rate can effectively avoid reconstructing NeRF with distorted colors.

%% file: sections/bilanerf_finishing.tex
\begin{figure}[t]
    \centering
    \centering
    \includegraphics[width=\linewidth]{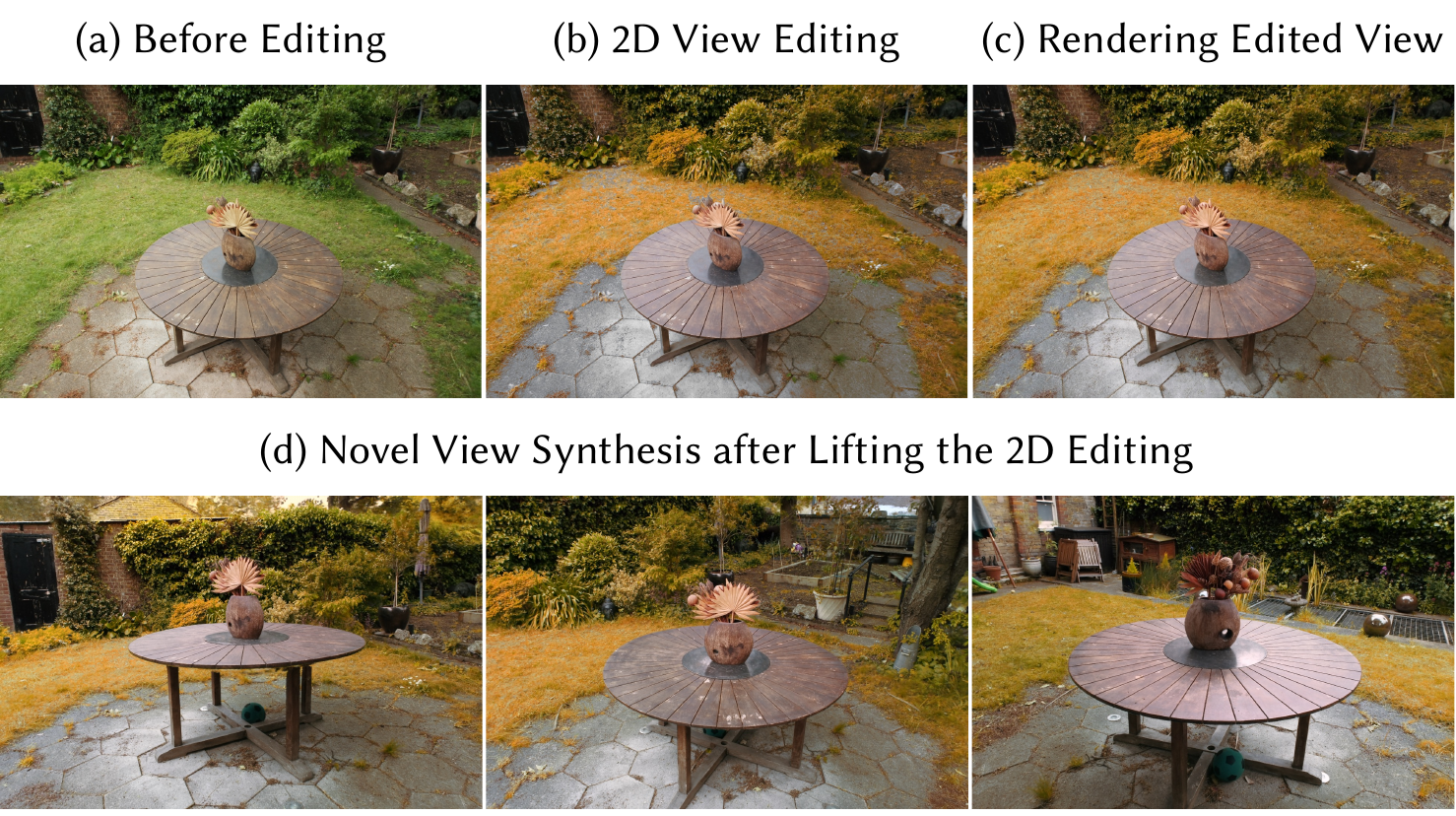}
    \caption{Results of our bilateral guided NeRF finishing. During 3D-level finishing, users are asked to select a view (a) and retouch it with image tools (b). Then, our method trains a 4D bilateral grid to close the gap between the NeRF rendering and the user editing. Upon optimizing the 4D bilateral grid, the user-adjusted retouching is not only mapped to the edited areas (c), but is also transferred to other areas (d) that are unseen in the edited view.}
    \label{fig:bilanerf_finishing}
\end{figure}

After the training stage, the per-view image enhancements by camera processing are removed. For better rendering quality, we need to enable human-adjusted finishing to enhance and retouch the reconstructed scene. In contrast to those approaches working on style transfer, our proposed workflow for retouching NeRF involves: 1) select a synthesized view; 2) edit the view in a photo editor (e.g., Adobe Lightroom\textregistered) or process the view with an automatic photo retoucher (e.g., HDRNet \cite{gharbi2017deep}); 3) Lift the 2D view edit to the 3D scene. Figure \ref{fig:bilanerf_finishing} illustrates this concept with our results.

There are several advantages of lifting 2D editing to 3D, instead of direct 3D editing. First, image enhancement is a well-studied problem, and we can utilize the existing 2D enhancement algorithms to ensure the best visual quality. Second, 2D image editing is more intuitive and familiar to users, and there are many powerful 2D editing tools. Thus, this interface is also more user-friendly.
To lift 2D editing to 3D space, we need to design a 3D point operation on the reference image, which transforms radiance per point in the 3D space. Drawing inspiration from photo editing in bilateral space, we devise the 4D bilateral grid as the data structure to accommodate the 3D point operation. Specifically, a 4D bilateral grid is formulated as a tensor $\boldsymbol{H} \in \mathbb{R}^{D \times W \times H \times M \times 12}$. Based on the slicing of 3D bilateral grid, we can define the slicing operation for 4D bilateral grid:
\begin{equation}
\begin{split}
    & \kappa'_{h,i,j,k}\bigg(x, y, z, g(c)\bigg) = \kappa_{i,j,k}\bigg(x, y, g(c)\bigg)~ \Lambda\bigg(D \cdot z - h\bigg) ~~, \\
    & \boldsymbol{H}\left[x, y, z, c \right] = \sum_{h,i,j,k} \kappa'_{h,i,j,k}\bigg(x, y, z, g(c)\bigg) \boldsymbol{H}_{h,i,j,k}~~,
\end{split}
\label{eqn:bilagrid3d_slice}
\end{equation}
% \tianfan{Define what $\vcentcolon=$ is}
where $(x,y,z)$ are normalized coordinates in $[0,1]^3$. Taking a point $\boldsymbol{x} = (x,y,z)$ with viewing direction $\boldsymbol{d}$, the color affine transformation at this point can be obtained by reshaping the slice $\boldsymbol{H}\left[x, y, z, c \right]$ to a $3\times 4$ matrix $\boldsymbol{\mathcal{H}}$, where $c$ is predicted through querying the scene representation $F_\Theta(\boldsymbol{x}, \boldsymbol{d})$. We can then apply the 4D bilateral grid $\boldsymbol{\mathcal{H}}$ to generate the processed radiance color $c_{e} = \boldsymbol{\mathcal{H}}[c~|~1]^\top$, as illustrated in Figure~\ref{fig:pipeline-figure}.

\paragraph{Low-Rank 4D Bilateral Grid.} Next, we need to learn the coefficients of the affine models in $\boldsymbol{H}$ from a given edited view $\mathcal{I}_{e}$. As shown in \cite{chen2016bilateral}, we can first approximate the transformation from the original image to the processed image approximated in a 3D bilateral grid. To find $\boldsymbol{H}$, a straightforward approach is to back-project the 3D bilateral grid to the 4D bilateral space. However, since we only have a single view editing, only a small proportion of the high-dimensional tensor $\boldsymbol{H}$ receives values by the back-projection, and we cannot propagate them to the rest of bilateral grid, without a strong prior to fill missing points. 

In this work, we model it as a matrix completion problem \cite{candes2010matrix,candes2012exact}. Under the low-rank assumption, missing points in an incomplete matrix can be recovered by solving a rank minimization. 
% Thus, we attempt to adopt the low-rank constraint to handle our problem.
But for a high-dimensional tensor, determining or approximating tensor rank is intractable. Thus, to solve tensor completion, rather than incorporating rank into the optimization objective, we alternatively construct a low-rank approximation of 4D bilateral grid $\boldsymbol{H}$.

Specifically, we use CP decomposition \cite{carroll1970analysis} to factorize the 4D bilateral grid $\boldsymbol{H}$ (a 5D-tensor) as a sum of outer products of vectors:
\begin{equation}
    \boldsymbol{H} = \sum_{r=1}^R \boldsymbol{\xi}^{1,r} \otimes \boldsymbol{\xi}^{2,r} \otimes \boldsymbol{\xi}^{3,r} \otimes \boldsymbol{\xi}^{4,r} \otimes \boldsymbol{\xi}^{5,r},
\label{eqn:bilagrid3dcp}
\end{equation}
where $\boldsymbol{\xi}^{1,r}\in \mathbb{R}^D$, $\boldsymbol{\xi}^{2,r}\in \mathbb{R}^W$, $\boldsymbol{\xi}^{3,r}\in \mathbb{R}^H$, $\boldsymbol{\xi}^{4,r}\in \mathbb{R}^M$, and $\boldsymbol{\xi}^{5,r}\in \mathbb{R}^{12}$ are factors for each dimension of $\boldsymbol{H}$. Their outer product yields a rank-1 tensor and the sum of $R$ rank-1 tensor results in a tensor with maximal rank of $R$. Specifying $R$ to relatively small number can constrain the rank of $\boldsymbol{H}$. This low-rank representation of high-dimensional tensors is also leveraged and extended by \citet{chen2022tensorf} to model 3D radiance fields. With the decomposed factors, the slicing operation can be rewritten as:
\begin{equation}
\begin{split}
    &\mathcal{V}_{h,i,j,k}^r\big(x, y, z, g(c)\big) = \kappa'_{h,i,j,k}\big(x, y, z, g(c)\big) \boldsymbol{\xi}^{1,r}_{h} \boldsymbol{\xi}^{2,r}_i \boldsymbol{\xi}^{3,r}_j \boldsymbol{\xi}^{4,r}_k, \\
    &\boldsymbol{H}\left[x, y, z, c \right] = \boldsymbol{U}\left[\bigoplus_{r=1,\dots, R} \left[ \sum_{h,i,j,k} \mathcal{V}_{h,i,j,k}^r\big(x, y, z, g(c)\big) \right]\right],
\end{split}
\label{eqn:bilagrid3dcp_slice}
\end{equation}
where $\mathcal{V}_{h,i,j,k}^r\big(x, y, z, g(c)\big)$ evaluates the linear interpolation along the first four dimensions, $\boldsymbol{U}\in \mathbb{R}^{12\times R}$ is a matrix constructed as $\left[\boldsymbol{\xi}^{5,1}, \boldsymbol{\xi}^{5,2}, \dots, \boldsymbol{\xi}^{5,R}\right]$, $\oplus$ is the concatenation operator that stacks $R$ scalars to a $R$-dimensional column vector. 

To initialize the factors, we begin by constructing a 4D bilateral grid (a 5D tensor) where all cells are set to identity affine transformations. Subsequently, we add small noises to the coefficients to make the 5D tensor full rank. Finally, we run PARAFAC decomposition \cite{kolda2009tensor} to obtain the initialized factors.

As we only require 2D editing of a single view, the low-rank 4D bilateral grid optimization is conducted over the edited view. Similar to the bilateral guided NeRF training stage, we also use rendering loss and TV loss (Equation \ref{eqn:bilagrid_nerf_loss}). In particular, due to the linearity of the factorization, the TV loss can be computed on the factors. We set the multiplier of TV loss $\lambda_{TV}$ to $1$ for all of our experiments. Over the optimization, we freeze the parameters of the reconstructed radiance fields and only propagate gradients to the factors.

% \begin{figure}[t]
%     \centering
%     \includegraphics[width=\linewidth]{figs/ablation_bilavsspatial.pdf}
%     \caption{Ablation study on bilateral space processing. The spatial only affine model fails to preserve the brightness of areas with high local contrast and will result in artifacts in synthesized novel views. While our bilateral space affine model can solve this problem by separately processing darker areas and brighter areas. \tianfan{The floater is less obvious in a single image rendering. Highlight that the floater is clearer in supplementary video. If the paper is too long, I think we can also just move this to supp.}}
%     \label{fig:ablation-bila-vs-spatial}
% \end{figure}

%% file: sections/experiments.tex
% \subsection{Implementation Details}

% % \tianfan{Should be 3D/4D bilateral grid? Similar for the next one}

% We implement our proposed method in Python, using the PyTorch framework. We choose ZipNeRF \cite{barron2023zipnerf} (re-implemented in PyTorch \cite{zipnerf-pytorch}) as the backbone, which is shown to achieve the state-of-the-art performance in novel view synthesis. As our method does not rely on a specific NeRF model, other backbones can also work with our implementation of the 3D/4D bilateral grid. We train the 3D/4D bilateral grid using Adam optimizer \cite{KingBa15}. Note that \cite{chen2016bilateral,gharbi2017deep} fit a bilateral grid on down-sampled images. In our scenario, since NeRF is trained on full-resolution images, we directly optimize bilateral grids on the same input, which can also produce considerable results. All of our experiments are conducted on a single RTX 3090 GPU.

% \tianfan{Very low priority: in the supplementary document, maybe show one sequence with reconstructed by either 8x8x4 or 16x16x8, to illustrate our method is robust to this parameter.}

% \subsection{Datasets}

\begin{table}[t]
    \centering
    \caption{We compare our method's performance in novel view synthesis to the baseline method (ZipNeRF), GLO-based approaches (w/GLO and w/AGLO), and NeRF variants with camera response function (HDRNeRF) and dark scene enhancement (LLNeRF).}
    % \vspace{-10px}
    % { >{\raggedright\arraybackslash}m {6.8em} | >{\centering\arraybackslash}m {8em} | >{\centering\arraybackslash}m {8em} |  >{\centering\arraybackslash}m {8em} |  >{\centering\arraybackslash}m {3.5em} }
    \begin{tabular}{l|c|c|c|c}
    % \begin{tabular}{l|c|c|c|c}
        % \toprule[0.5pt]
        \textbf{} & sRGB & \multicolumn{3}{c}{Affine-aligned sRGB} \\
        \textbf{} & PSNR $\uparrow$ & PSNR $\uparrow$ & SSIM $\uparrow$ & LPIPS $\downarrow$ \\ \hline
        ZipNeRF & \cellcolor{yellow!30}23.07 & 25.17 & \cellcolor{yellow!30}0.8224 & \cellcolor{yellow!30}0.1823 \\
        ZipNeRF w/GLO & \cellcolor{orange!30}23.15 & \cellcolor{orange!30}26.54 & \cellcolor{orange!30}0.8521 & \cellcolor{orange!30}0.1439 \\
        ZipNeRF w/AGLO & 22.16 & \cellcolor{yellow!30}25.39 & 0.8156 & 0.2003 \\  \hline
        HDRNeRF & 22.56 & 24.03 & 0.6451 & 0.3951 \\ 
        LLNeRF & 18.37 & 21.71	& 0.6602 & 0.4466 \\ \hline
        Ours & \cellcolor{red!30}23.83 & \cellcolor{red!30}27.53 & \cellcolor{red!30}0.8696 & \cellcolor{red!30} 0.1336 \\ \hline
    \end{tabular}
    \label{tab:quanti_eval}
    % \vspace{-10px}
\end{table}

\subsection{Evaluation}

We first quantitatively assess the effectiveness of our method in handling photometric variation for novel view synthesis.

% \begin{figure}[t]
%     \centering
%     \includegraphics[width=\linewidth]{figs/cmp_glo.pdf}
%     \vspace{-15px}
%     \caption{Comparison of GLO-based methods and our method.}
%     \label{fig:cmp-glo}
%     % \vspace{-0.2in}
% \end{figure}

\vspace{5px}
\paragraph{Datasets.} Existing real-world NeRF datasets, such as mip-NeRF 360 \cite{barron2022mip}, are captured carefully under fixed camera parameters and consistent lighting. To evaluate the quality of novel view synthesis for scenes captured with photometric variation, we use cell phone cameras to collect 7 real-world nighttime scenes for numerical evaluation. Two of them are captured with iPhone 13 and the rest are captured with OnePlus 9. During capture, in addition to the automatic adjustments by camera programs, we accordingly change local exposure and ISO for each view when the environment is too dark. In certain views, we intentionally introduce wild exposure variation to stress test our method, which also mimics HDR capture for improving image quality in low-light environments. Our dataset only contains sRGB images. As in most NeRF datasets, we use COLMAP \cite{schonberger2016structure} to estimate camera poses.

% TODO: This practice aligns with the routine of cell phone users, ensuring the capture quality

\begin{figure}[t]
    \centering
    \includegraphics[width=0.95\linewidth]{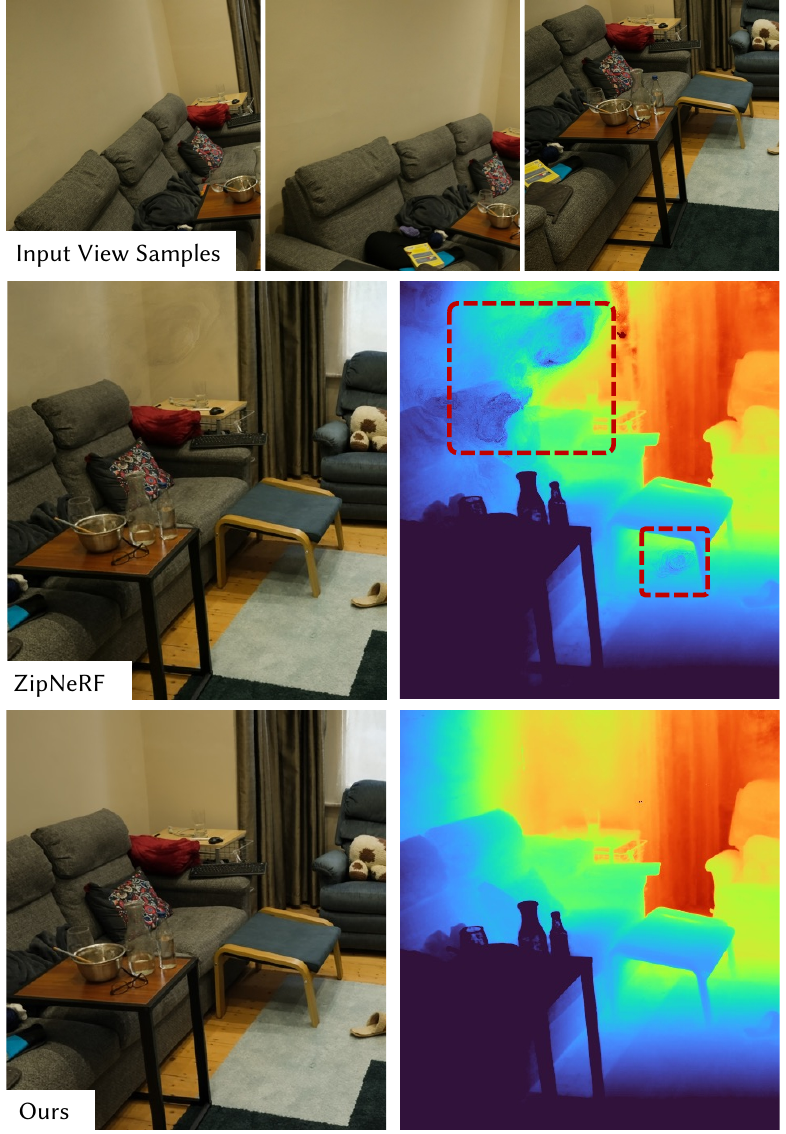}
    % \vspace{-15px}
    \caption{Comparison of our method and ZipNeRF baseline \cite{barron2023zipnerf} on an indoor scene with relatively minor photometric variation across the input views. Even though, the minor variation results in floaters in the ZipNeRF results. Our bilateral guided training can effectively disentangle the variation and overcome this issue.
    % from the mip-NeRF 360 dataset
    % \tianfan{1) Put ours at the bottom and ZipNeRF on the top. 2) Do you think we need to also show some input views here (to illustrate that there is only minor photometric variation?
    }
    \label{fig:cmp-slightproc}
\end{figure}

\begin{figure}[t]
    \centering
    \includegraphics[width=\linewidth]{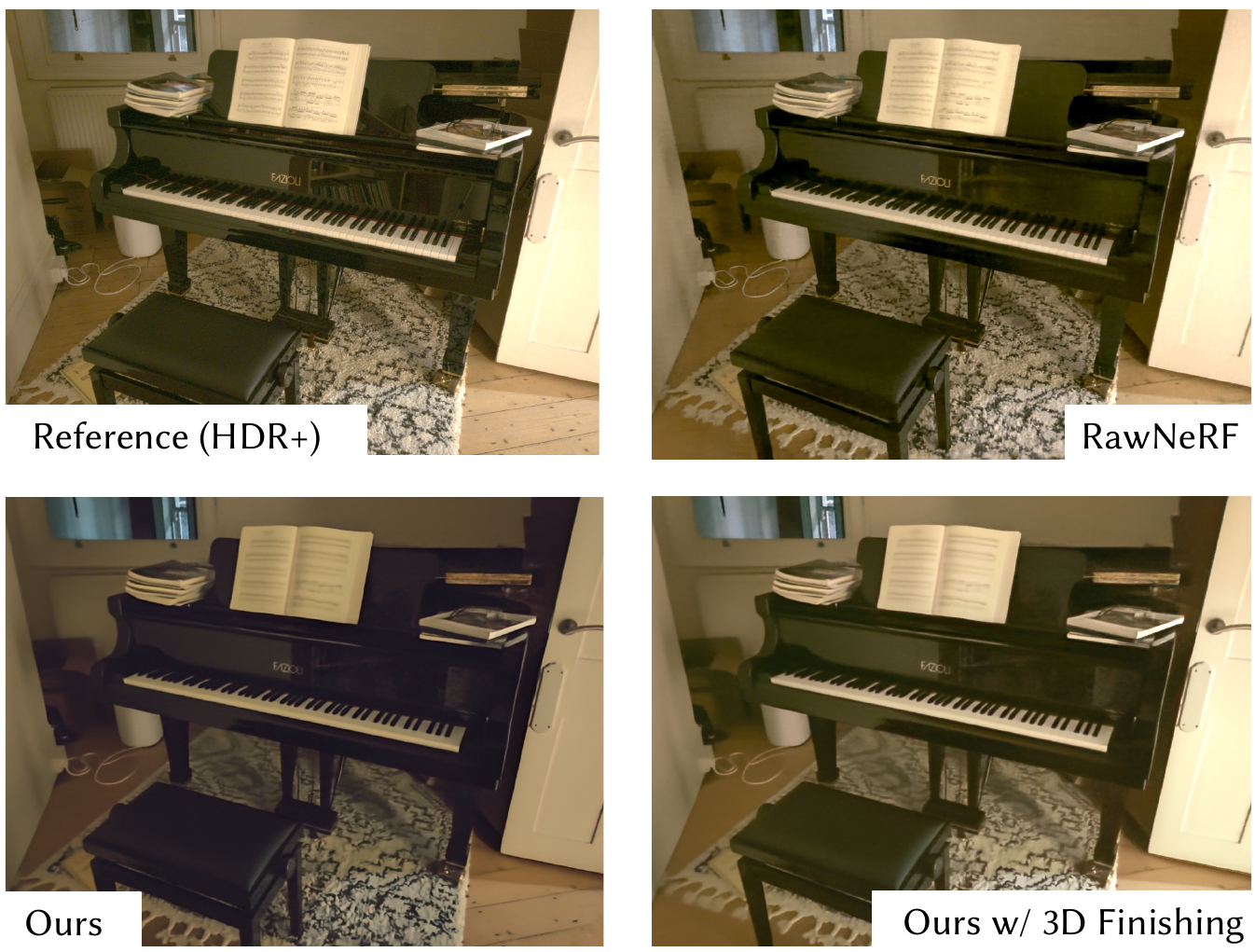}
    % \vspace{-15px}
    \caption{Comparison of RawNeRF \cite{mildenhall2022nerf} and our method. The reference image is merged using HDR+. RawNeRF is directly trained on raw images, while our method only requires processed sRGB images. The color bias observed between our result and the reference image is caused by the denoiser, which distorts noise, leading to a purple shift. To address this, we apply our bilateral guided radiance finishing to extend the correct color from the reference image across the entire scene.}
    \label{fig:cmp-rawnerf}
    % \vspace{-0.2in}
\end{figure}

\vspace{5px}
\paragraph{Compared Methods.} Our comparison focuses on methods designed for scenes with photometric variation or low-light conditions. GLO \cite{martin2021nerf} is a widely adopted technique for encoding the variation in exposure and lighting. We compare both the original GLO (ZipNeRF w/GLO) and the affine GLO  (ZipNeRF w/AGLO) proposed in \cite{barron2023zipnerf}. To handle multi-view captures with varying exposure, HDRNeRF \cite{huang2022hdr} approximates the camera response function with a per-channel MLP. \citet {wang2023lighting} (LLNeRF) propose a method to enhance the lightness, reduce noise, and correct color distortion for training NeRF in dark scenes. A baseline method ZipNeRF \cite{barron2023zipnerf} will be included in our comparison as a contrast that does not handle photometric variation.

\begin{figure*}[t]
    \centering
    \includegraphics[width=\textwidth]{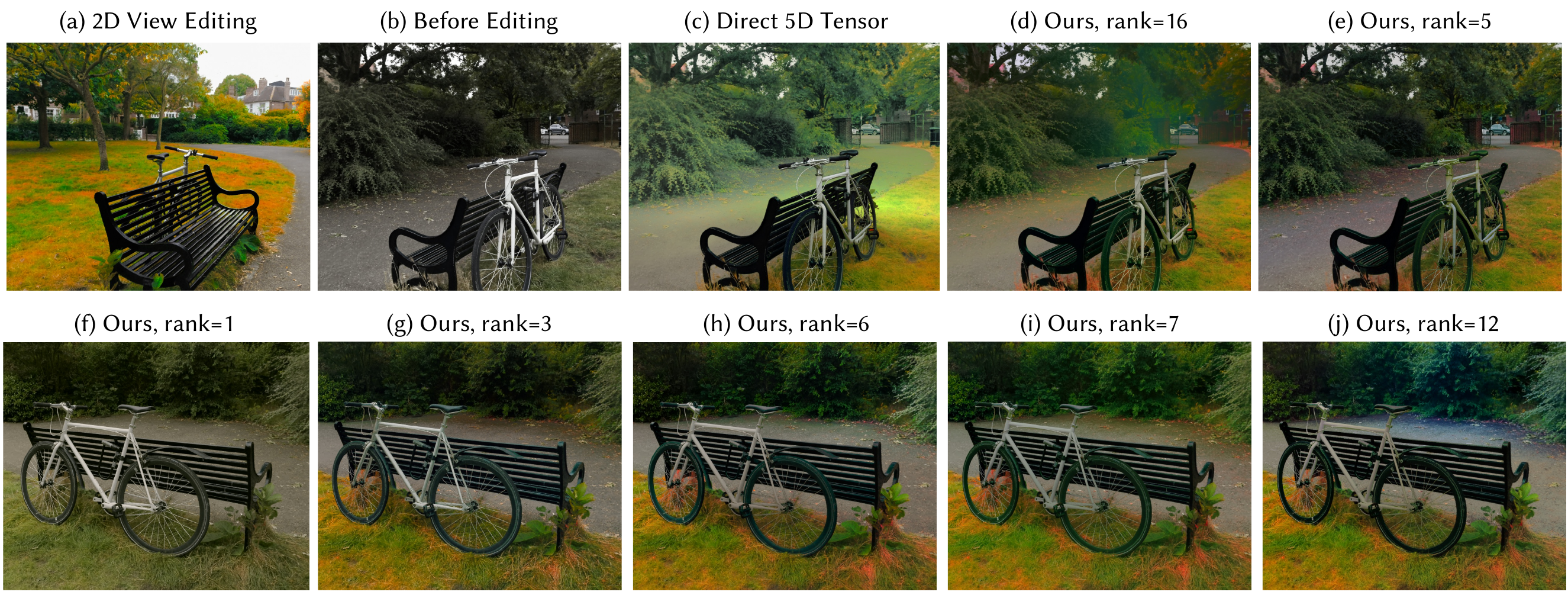}
    \caption{Ablation study on the low-rank approximation of 4D bilateral grid for radiance-finishing. Given that there is only one edited view, the direct 5D tensor (c) fails to recover the affine model coefficients for the unseen areas, leading to color distortion in the synthesized view. For high-rank approximation (d), unnatural color also appears. Our proposed low-rank approximation (e) with rank=5 can produce reasonable appearance in unseen areas. From (f) to (j), we further illustrate the impacts of ranks on the finishing results.}
    % \tianfan{1) It is better to move this figure to Section 4, as an illustration of the Low-Rank approximation. 2) Shall we also show 2D view w.o. editing?}
    % \tianfan{Low priority: result (d) is slightly too dark and grass land is a bit over saturated. Maybe we can tune 2D view a bit to make it better, or change a different view.}
    \label{fig:ablation-direct5d-vs-lowrank}
    % \vspace{-10px}
\end{figure*}

\begin{figure}[t]
    \centering
    \includegraphics[width=\linewidth]{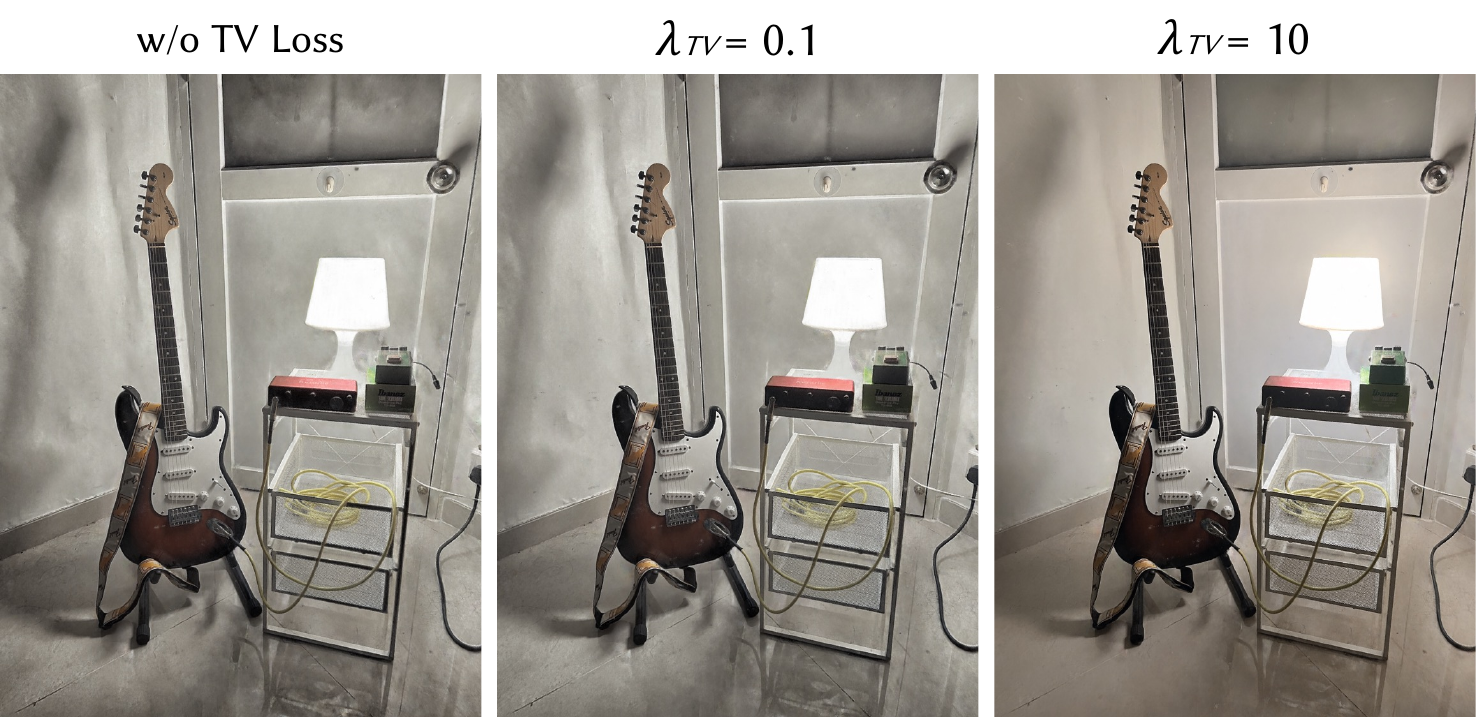}
    \caption{Ablation study on the weight of TV loss. Without TV loss, the 3D bilateral over-processes the appearance during NeRF training, resulting in ghost artifacts (left). Applying a weak TV regularization cannot mitigate this issue (middle). When raising the weight of TV loss to 10, most ghosting artifacts disappear (right).}
    % \vspace{-10px}
    \label{fig:ablation-tvloss}
    % \vspace{-10px}
\end{figure}

\begin{figure}[t]
    \centering
    \includegraphics[width=0.95\linewidth]{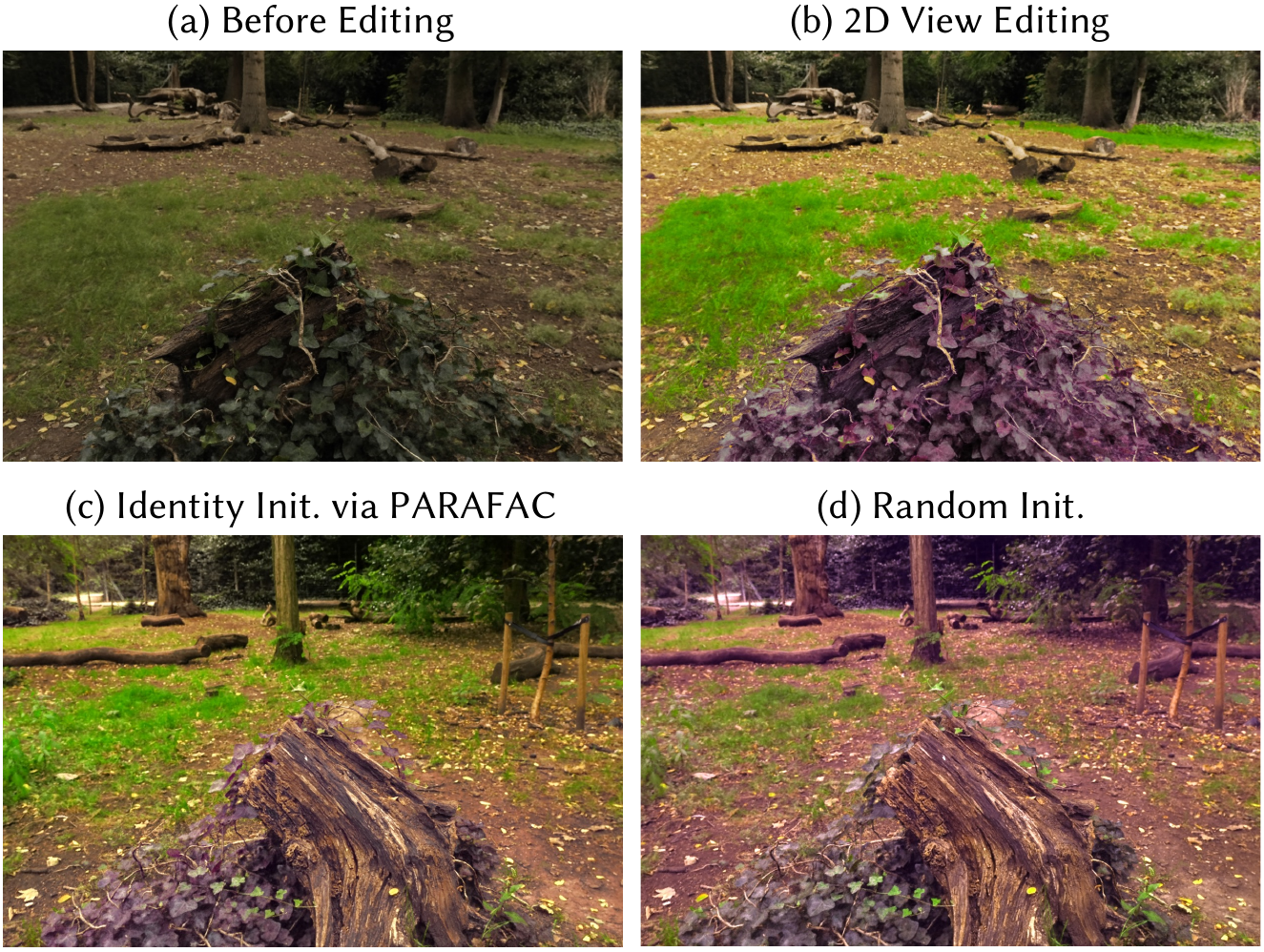}
    \caption{Ablation study on the initialization schemes of the 4D bilateral grid. Using identity initialization via PARAFAC (c), the optimized low-rank bilateral grid achieves consistent editing propagation to the whole scene. Random initialization (d) will cause the propagation to degrade.}
    \label{fig:ablation-parafacinit}
    % \vspace{-10px}
\end{figure}

\begin{figure*}[t]
    \centering
    \includegraphics[width=0.91\textwidth]{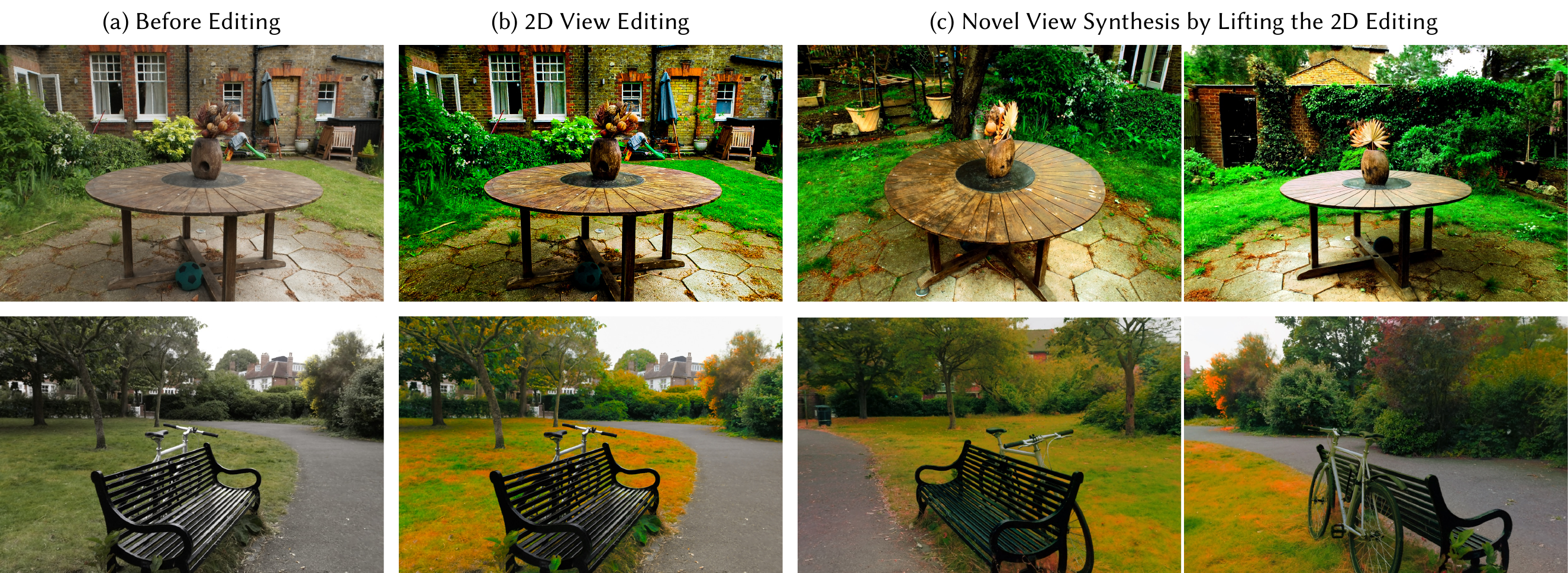}
    % \vspace{-5px}
    \caption{The bilateral guided NeRF finishing results on 360 scenes. Our method can consistently lift the 2D view editing to the entire 3D scene.}
    \label{fig:res_360_retouching}
\end{figure*}

\begin{figure*}[t]
    \centering
    % \vspace{-4px}
    \includegraphics[width=0.91\textwidth]{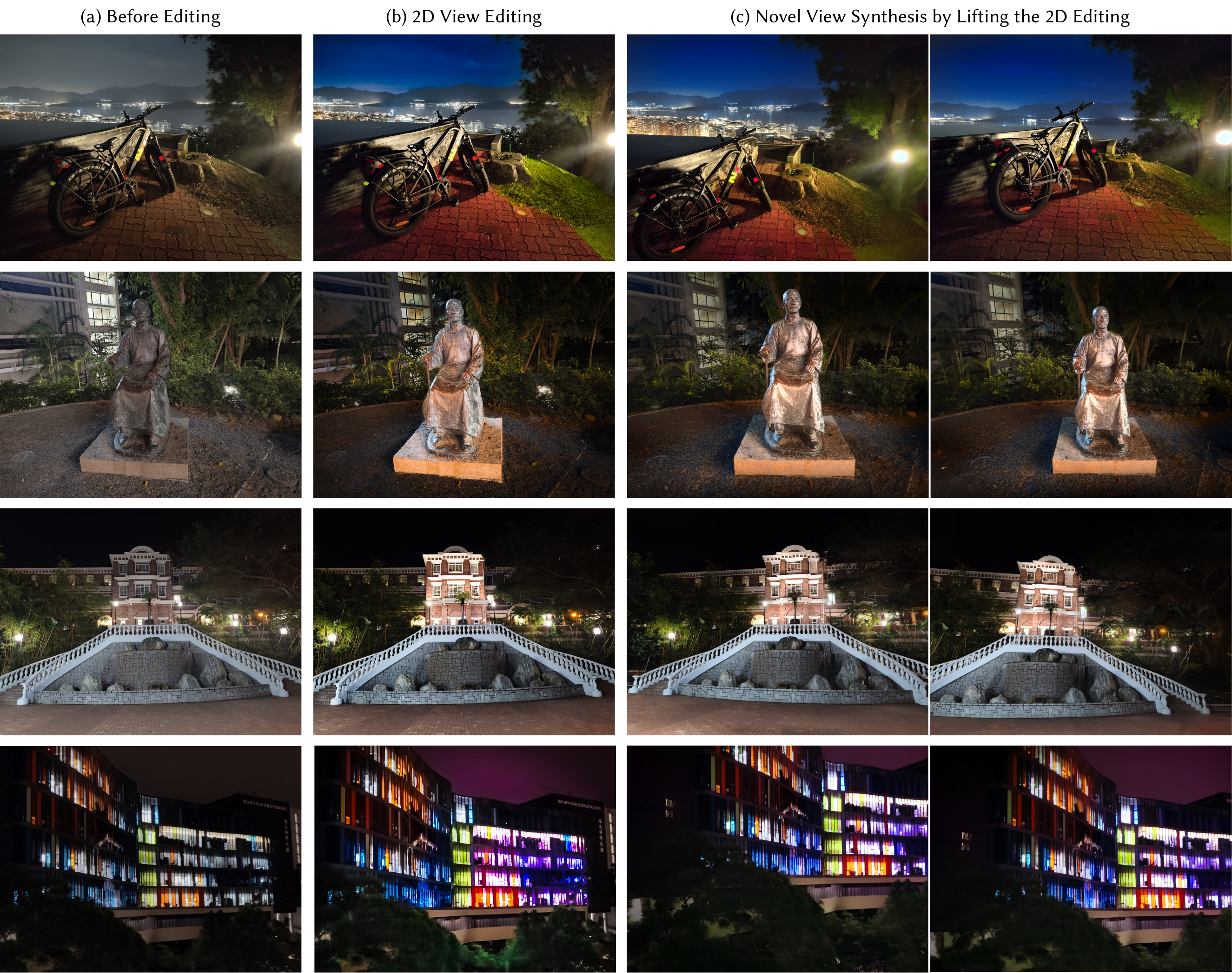}
    % \vspace{-5px}
    \caption{The bilateral guided NeRF finishing results on our captured nighttime scenes. Light enhancement, color adjustment, etc. are supported.}
    % \tianfan{Low priority: In the first row, the purple sky looks slightly fake to me. Maybe the dark blue tone might be more natural. https://stitchpalettes.com/palette/night-sky-spa0196/}
    \label{fig:res_ourdata_retouching}
\end{figure*}

\begin{figure*}[t]
    \centering
    \includegraphics[width=0.97\textwidth]{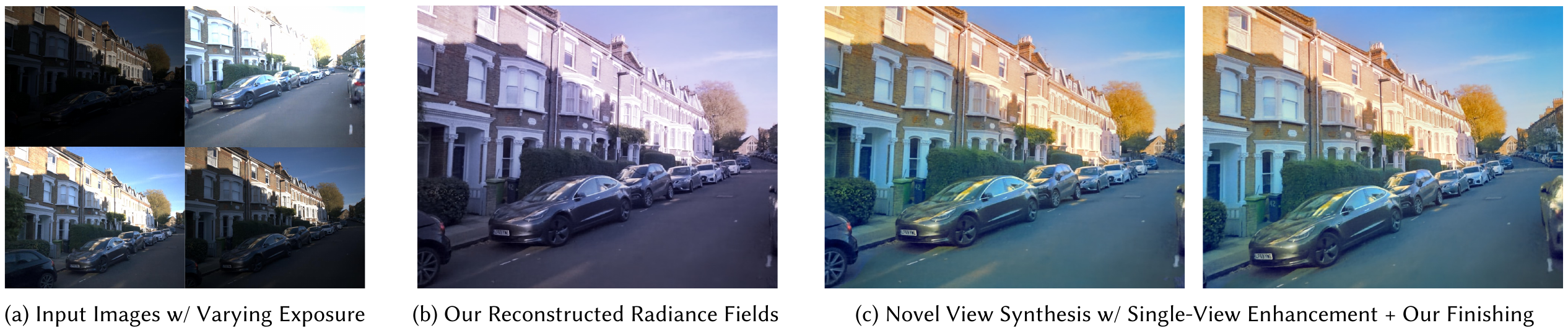}
    \vspace{-5px}
    \caption{% Our bilateral guided NeRF training can fuse best-exposure parts from the input images (a) into the reconstructed radiance fields (b). Our radiance-finishing can further adjust the color tone of the fused radiance fields by lifting a single view enhancement (c).
    Our bilateral guided NeRF training can fuse multiple low-dynamic-range input images (a) with different exposures into a high-dynamic-range radiance field (b). Note that the reconstructed radiance field preserves both details in highlights (e.g. the bright parts of the building) and deep shadows (e.g. the bushes), while none of the input images can preserve both. Our radiance-finishing can further adjust the color tone of the fused radiance fields by lifting a single view enhancement (c).
    }
    % \tianfan{Revise this caption to highlight HDR effect. feel free to revert if you are not comfortable.}
    % \tianfan{I think we can highlight that our method can recover \textit{HDR} radiance fields, as from (a) to (b), our method clearly does HDR fusion.}
    \label{fig:hdr-fusion}
\end{figure*}

\begin{figure*}[t]
    \centering
    \includegraphics[width=0.97\textwidth]{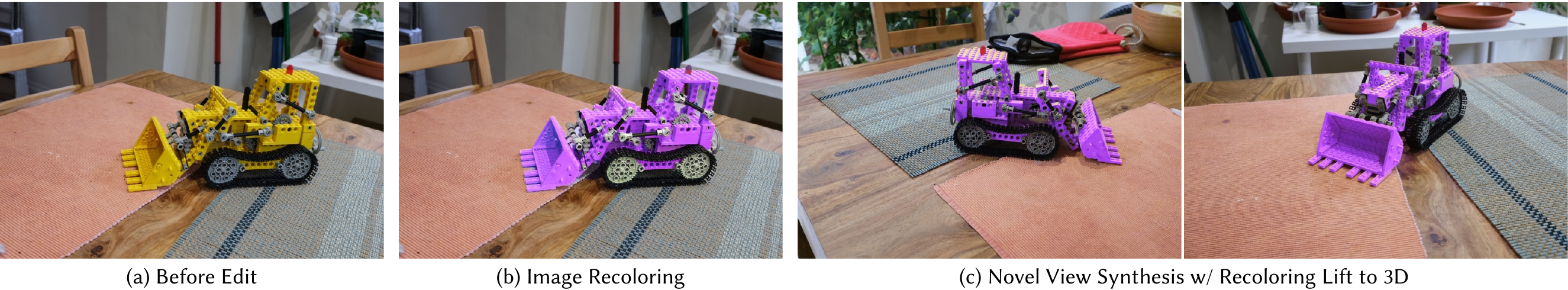}
    % \vspace{-10px}
    \vspace{-5px}
    \caption{Since our bilateral guided radiance-finishing is a local operator, we can use it for object-level color editing. In this case, we first edit the color of the bulldozer in a single view, then train the low-rank 4D bilateral grid on the 2D editing to perform 3D-level recoloring.}
	\label{fig:recoloring}
\end{figure*}

\begin{figure*}[t]
    \centering
    % \vspace{10px}  
    \includegraphics[width=0.97\textwidth]{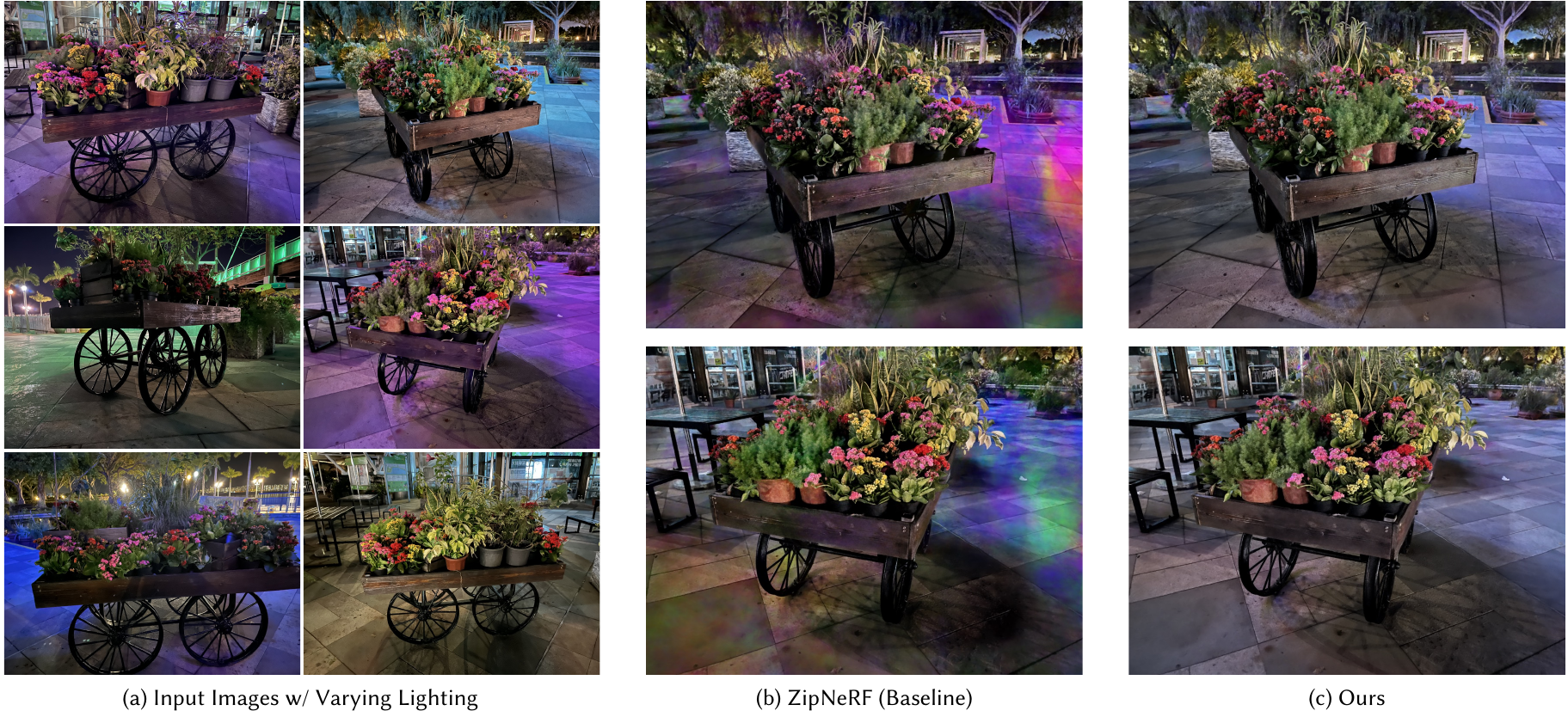}
    % \vspace{-10px}
    \vspace{-5px}
    \caption{Our method can disentangle the varying environmental lighting. In this scene, the color of the light source is changing over time. The baseline method will incur artifacts in synthesized novel views. Although our method does not explicitly model the light source, the bilateral grid is shown to be capable of encompassing the variation in the light color.}
	\label{fig:varying-light}
    \vspace{-5px}
\end{figure*}

\vspace{5px}
\paragraph{Comparative Evaluation.} We present our quantitative results in Table \ref{tab:quanti_eval}. Due to the per-view ISP enhancements in our dataset, there is color tint bias among ground truth images and the output. Following the evaluation protocol in \cite{mildenhall2022nerf,barron2022mip}, we calculate a per-channel affine transformation to align the output color tints with the ground truth tints (Affine-aligned sRGB). Table \ref{tab:quanti_eval} shows that our bilateral guided radiance field achieves the best performance in all metrics. Compared to GLO-based approaches, our method disentangles photometric variation without compromising rendering quality. Compared to the per-channel MLP tone mapper used in HDRNeRF, which can only model the global tone curve, our method can also model local operations in modern cameras. Moreover, the light enhancement module in LLNeRF fails to overcome the varying exposure and lighting conditions in dark scenes, while our method can handle it.

To validate the performance of our method on multi-view captures with slight photometric variation (a more common scenario for NeRF input), we conduct a comparison of our method and the baseline ZipNeRF on an indoor scene from the mip-NeRF 360 dataset, as shown in Figure \ref{fig:cmp-slightproc}. This scene is captured in a well-lit condition without manually adjusting camera parameters. The small inconsistencies across input views still induce floaters in the rendered views of ZipNeRF, which are removed by our bilateral guided training.

% \TODO{Figure \ref{fig:cmp-rawnerf}: comparison of our method and RawNeRF.}

% training NeRF in the linear raw space can avoid the impact of ISP processing. However, raw data accessibility is limited due to hardware restrictions or storage overhead. At last, we compare with RawNeRF \cite{mildenhall2022nerf}, we utilize a scene from the RawNeRF dataset, where raw images and meta information are available.
In the last, we compare with RawNeRF \cite{mildenhall2022nerf} utilizing a scene from the RawNeRF dataset, where raw images and meta information are available. For our model input, we implement a simple ISP with the Restormer denoiser \cite{zamir2022restormer} and HDR+ finishing\footnote{We use this open-source implementation: \clink{\url{https://github.com/amonod/hdrplus-python}}.}. Figure \ref{fig:cmp-rawnerf} demonstrates that both our method and RawNeRF can produce floater-free view synthesis. RawNeRF achieves this by directly training on linear raw input, while our method achieves similar results without raw image access. Note that many cameras do not support raw image access due to hardware limitations or storage overhead. Nevertheless, the short exposure time results in significant noise, which degrades the denoiser and leads to a color bias (purple shift) in the sRGB input of our method. Thereby the color bias will be baked onto the optimized radiance fields after training. To restore the correct color, we first take the reference image as the retouching target, which exhibits the correct color after HDR+ merging. Then, our proposed bilateral guided radiance finishing can be employed to lift the correct color in the reference image to the whole 3D radiance field.
% To mitigate this issue, we use the reference image as the target editing to restore the correct color of the radiance field using our bilateral guided radiance finishing.
% \tianfan{Change this paragraph. Please verify}

\subsection{Ablation Study}

We conduct ablations for two main components. For more ablations, please see our supplementary.

\vspace{10px}
\paragraph{TV Loss.} We demonstrate the impacts of TV loss in Figure \ref{fig:ablation-tvloss}. Removing the TV loss, our 3D bilateral grid leaves ghost artifacts on the reconstructed radiance fields. This is because the translation component of the local affine model can partially encode low-frequency appearance, e.g., the shadow on the wall. Thus, imposing TV loss can penalize the differences among the translation components in adjacent cells. However, we observe that setting a small weight of TV loss ($\lambda_{TV} = 0.1$) fails to address this issue successfully. It is only with a large weight of TV loss ($\lambda_{TV} = 10$) that results without the ghost artifacts are obtained.

% \vspace{-5px}
% \paragraph{Bilateral Processing.} 
% To highlight the advantages of bilateral processing, we juxtapose our method with spatial-only processing, where each pixel will be processed only depending on its spatial position. As illustrated in Figure \ref{fig:ablation-bila-vs-spatial}, spatial-only processing exhibits degradation in highly contrasted areas, leading to artifacts in novel view synthesis. This is attributed to the conflicting processing within a local patch for brighter and darker pixels. In contrast, our method treats pixels based on both their spatial positions and values. Therefore, even two neighboring pixels can be processed by different affine models if there is a disparity in their values.

% Since our method treats pixels based on their spatial positions and values, even two neighboring pixels will be processed by different affine models if there is gap between their values.

\paragraph{Low-Rank Approximation.} 
To affirm that low-rank approximation is pivotal for achieving considerable finishing results, we model the 4D bilateral grid as a direct 5D tensor and train it on a 2D view editing. Additionally, we set the rank $R$ to 16 to investigate whether high-rank approximation can attain comparable results. Figure \ref{fig:ablation-direct5d-vs-lowrank} presents their comparisons with our low-rank approximation in a 360$^\circ$ scene. For the direct 5D tensor and high-rank approximation, color distortion occurs in those areas that are invisible to the editing view. Whereas our low-rank modeling can fill the unseen areas with appropriate coefficients, maintaining the harmonized appearance. This validates that the low-rank constraint can effectively regularize the unseen space in the 4D bilateral grid.

In our editing results, we typically use a rank of 5 for the 4D bilateral grid. Deficient rank, e.g. rank=1, will limit the capacity of the bilateral grid and fails to lift the 2D view editing, as shown in Figure \ref{fig:ablation-direct5d-vs-lowrank}(f). Furthermore, we find using ranks of 3, 5, 6, and 7 all allows for consistent lifting of the 2D retouching to 3D with similar outcomes, indicating our approach is not highly sensitive to the choice of rank within this range. Setting the rank near or higher than 10 is generally not recommended due to potential color distortion issues, as demonstrated in Figure \ref{fig:ablation-direct5d-vs-lowrank}(j), where a rank-12 4D bilateral grid is used.

\paragraph{Initialization of 4D Bilateral Grids.} Figure \ref{fig:ablation-parafacinit} illustrates the importance of the identity initialization for optimizing a 4D bilateral grid. Recall that the low-rank 4D bilateral grid is represented in CP-decomposed vectors. With random initializations of those vectors, color transformations applied to the stump (changing from dark green to violet) erroneously affect the background. In contrast, our identity initialization scheme using PARAFAC can maintain considerable retouching results for unseen areas. Since the regions outside the edited view are not directly supervised during the finishing stage, bilateral grid cells covering those regions are optimized using only the low-rank prior and TV regularization. This demonstrates that employing identity transformations as a starting point is a practical approach for this ``semi-supervised'' procedure.
% As those regions outside the edited view are not directly supervised during the finishing stage, starting optimization with identity transformations can give a good initialization for preserving the original appearance patterns.

\subsection{Applications}

We showcase applications empowered by our proposed bilateral guided training and finishing method. For more results, please refer to the supplementary materials.

\begin{figure}[t]
    \centering
    \begin{subfigure}[a]{\linewidth}
        \centering
        \includegraphics[width=0.95\linewidth]{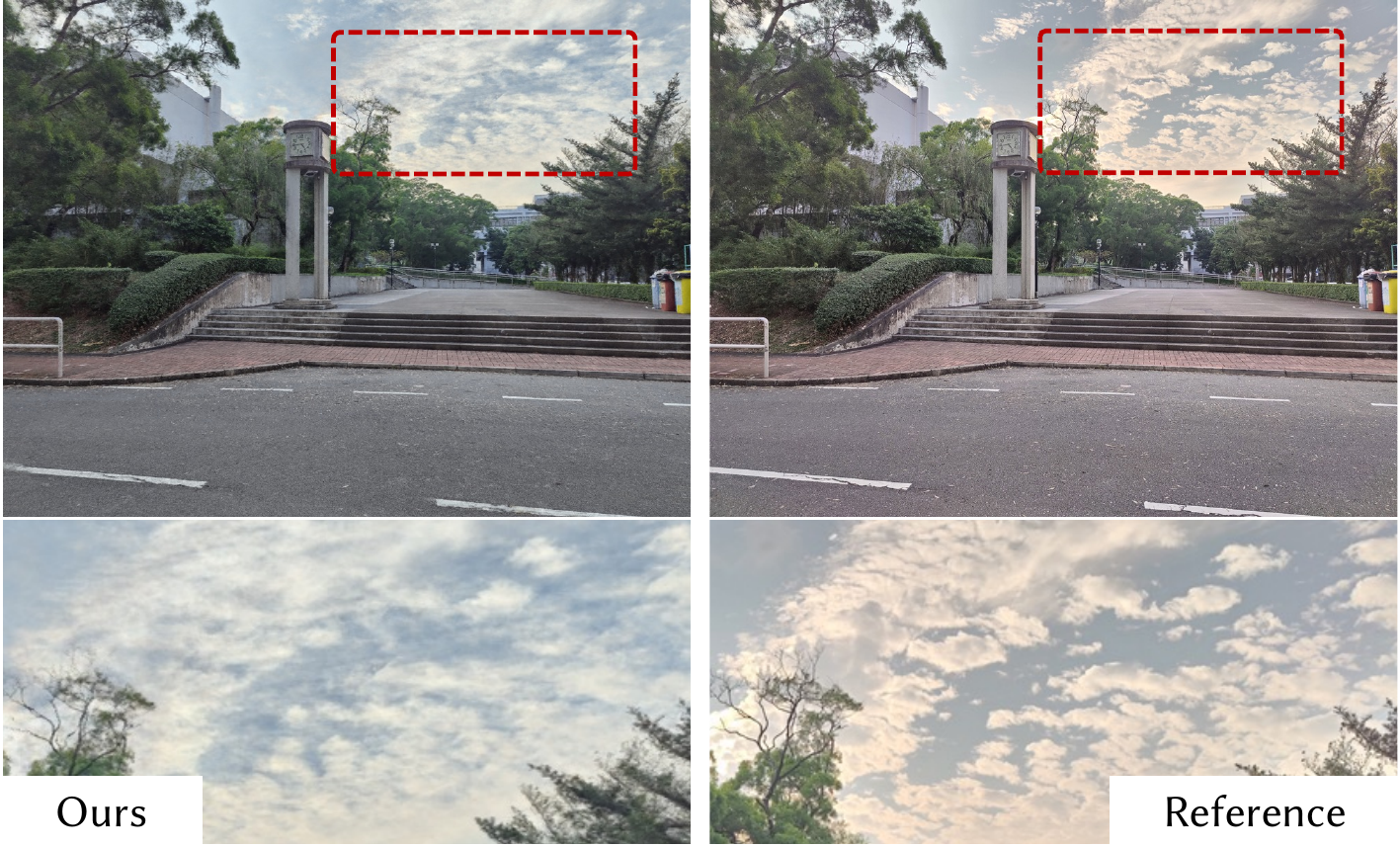}
        \caption{Floaters caused by moving and transient objects, e.g., clouds, cannot be handled by our approach.}
        \label{fig:lim-transient}
    \end{subfigure}
    \begin{subfigure}[a]{\linewidth}
        \centering
        \vspace{0.1in}
        \includegraphics[width=0.95\linewidth]{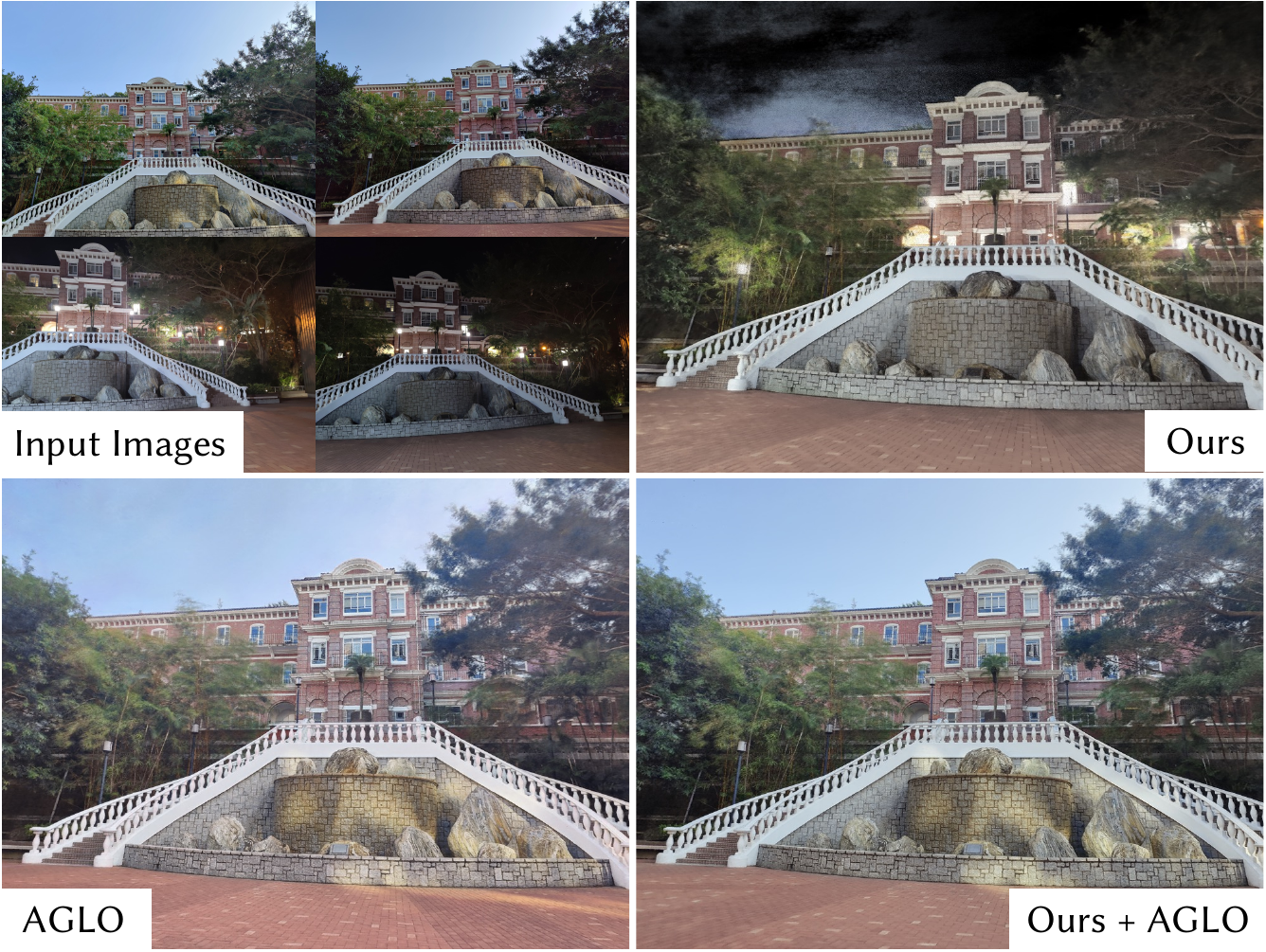}
        \caption{Results of our method on the ``in-the-wild'' setting. Compared with GLO-based methods, our approach is only capable of handling photometric variation by ISP. We also present a result obtained by combining our approach and AGLO.}
        \label{fig:lim-nerfw}
    \end{subfigure}
    \caption{Typical failure cases of our method in decoupling multi-view inconsistencies and distractors for floater-free NeRF reconstruction.}
    % \vspace{-10px}
    \label{fig:failturecases-training}
\end{figure}

\paragraph{3D Enhancements and Retouching.} Our radiance-finishing provides a user-friendly way to enhance and retouch 3D radiance fields. Figure \ref{fig:res_360_retouching} shows its performance on 360$^\circ$ scene editing. 
% For our collected data, our method can allow 
With the proposed approach, users can apply desired enhancements to stylize the night view captures, as exhibited in Figure \ref{fig:res_ourdata_retouching}. We also develop a simple interactive 3D editor with our method, built on Instant-NGP \cite{muller2022instant}. See our supplementary demo video and project page for the live demo and more results.
% See our supplementary video for the live demo.

\paragraph{HDR Fusion in Radiance Fields.} Our 3D bilateral grid can merge input images with varying exposure into a single radiance field. In Figure \ref{fig:hdr-fusion}, we present an example of HDR fusion in NeRF. The input multi-view images are generated from RawNeRF \cite{mildenhall2022nerf} with underexposure, normal exposure, slight overexposure, and overexposure. Our method manages to fuse the best-exposed parts from different images, resulting in a radiance field that preserves both details in shadows and highlights (the bright sky and buildings). After that, our finishing pipeline can perform color adjustment and tone mapping for the whole scene at the 3D level.

% \vspace{10px}
\paragraph{Object Recoloring} Figure \ref{fig:recoloring} presents an object recoloring result accomplished by our proposed method. Since our radiance-finishing is conducted via local affine models in bilateral space, the image-level color transformation will be mostly mapped to those cells covering both the bulldozer and yellow pixels.

\paragraph{Disentanglement of Varying Lighting.} In addition to modeling the variation in camera pipeline processing, our method shows capability for scenes with varying light colors. In Figure \ref{fig:varying-light}, we test our method on an outdoor scene, where the color of the light source significantly changes during the capture. Without handling the light source color, ``disco'' artifacts appear in the results of the baseline method. In contrast, the bilateral grid in our method can disentangle the impacts of varying light colors, facilitating novel view synthesis with unified lighting.

\subsection{Limitations}

Our method shows considerable performance in facilitating floater-free NeRF reconstruction and 3D finishing. But we also explore its deficiencies. First, unlike GLO, the bilateral grid struggles to model transient objects, e.g., the moving clouds in the sky (Figure \ref{fig:lim-transient}). It also fails to support in-the-wild NeRF reconstruction (Figure \ref{fig:lim-nerfw}), while GLO is originally designed for this setting. We also show that our method and GLO can jointly work for this setting, resulting in slightly improved visual quality compared with the GLO-only method, e.g., the original staircase color in the input images is preserved. A future direction is to ensure bilateral grids disentangle photometric variation and GLO vectors solely explain away other distractors.

\begin{figure}[t]
    \centering
    \includegraphics[width=\linewidth]{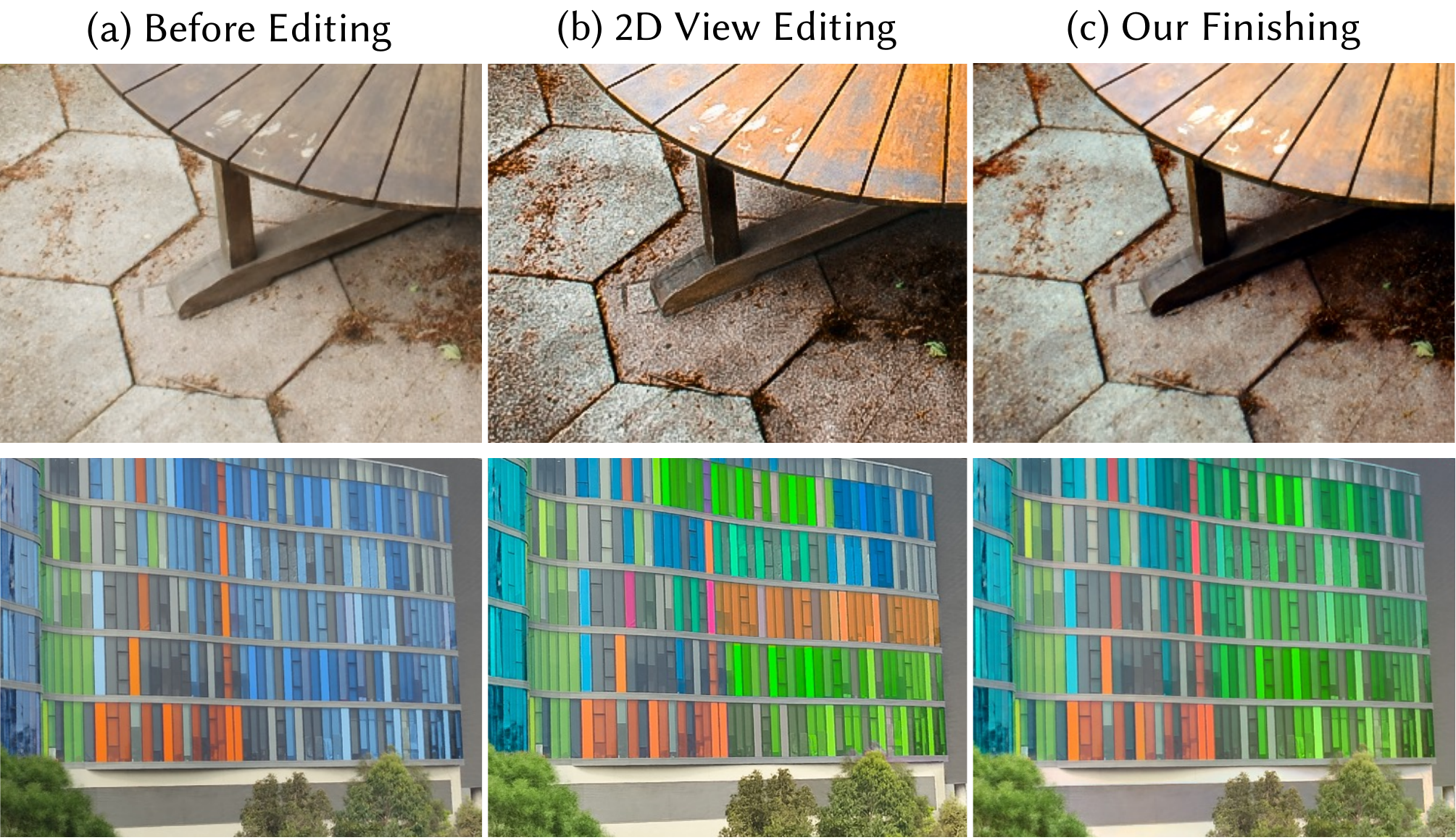}
    \caption{Limitaions of our proposed finishing method. The first row shows that the 4D bilateral grid fails to lift sharpening manipulations. The second row illustrates a challenging scenario where local recoloring is degenerated.}
    \label{fig:failurecases-finishing}
\end{figure}

For the finishing stage, we show two failure cases of our method in Figure \ref{fig:failurecases-finishing}. In the first case, since the 4D bilateral grid is a point operator, 2D image manipulations, including blurring, sharpening, etc., cannot be lifted to 3D. For the second case where a challenging local edit is applied to the building, using a single low-resolution 4D bilateral grid fails to express the local changes perfectly. Larger bilateral grid resolution and higher rank cannot effectively improve the performance. One potential solution to the failures is to devise a multi-scale bilateral grid as mentioned in HDRNet \cite{gharbi2017deep}, which processes the scene in multiple levels of detail.
% One potential solution to the failure of this local recoloring is to devise a non-uniform bilateral grid, which adaptively processes the scene by the distributions of edited areas~\tianfan{Maybe you can instead mentioned the multi-scale bilateral,  which mentioned in HDRnet paper.}.
Another starting point is to explore more ``expressive'' priors other than low-rank for filling the unseen cells of the 4D bilateral grid.

%% file: sections/appendix.tex
\section{Implementation Details}
% \tianfan{Should be 3D/4D bilateral grid? Similar for the next one}

We implement our proposed method in Python, using the PyTorch framework. We choose ZipNeRF \cite{barron2023zipnerf} (re-implemented in PyTorch \cite{zipnerf-pytorch}) as the backbone, which is shown to achieve the state-of-the-art performance in novel view synthesis. As our method does not rely on a specific NeRF model, other backbones can also work with our implementation of the 3D/4D bilateral grid. We train the 3D/4D bilateral grid using Adam optimizer \cite{KingBa15}. Note that \cite{chen2016bilateral,gharbi2017deep} fit a bilateral grid on down-sampled images. In our scenario, since NeRF is trained on full-resolution images, we directly optimize bilateral grids on the full-resolution synthesized views. All of our experiments are conducted on a single RTX 3090 GPU.

In our experiments for NeRF training with 3D bilateral grids, we set the resolution $(W, H, M)$ to $(8, 8, 4)$ for scenes with moderate photometric variation and $(16, 16, 8)$ for scenes with relatively large variation. For the experiments on bilateral guided finishing, the 4D bilateral grid resolution $(D, W, H, M)$ is set to $(16, 16, 16, 8)$ or $(32, 32, 32, 16)$, depending on the level of fineness in the view editing. The rank $R$ is set to 5 for coarser editing and set to 8 for finer editing.

Due to the efficiency of the bilateral grid, the training time primarily depends on the specific NeRF backbone. For our backbone, the training time is $\sim$2.25 hours, thereby our training stage consumes the same time. We train the low-rank 4D bilateral grid for 2,500 iterations during the 3D lifting stage, which costs $\sim$15 minutes at convergence. For a faster finishing stage, training for 500-1000 iterations can also produce considerable results.

\section{Data Preparation}

We use three datasets in our experiments: 1) our own dataset with wild photometric variation; 2) mip-NeRF 360 dataset \cite{barron2022mip}; 3) RawNeRF dataset \cite{mildenhall2022nerf}.

We use hand-held cellphones (iPhone 13 and OnePlus 9) to capture our own data. For each scene, we capture 20-93 images. The duration of the capture process for each scene spans between 10 and 60 minutes, as long exposure is enabled for low-light photography.

To test our method on RawNeRF scenes, we process its provided noisy raw images to sRGB images. Since the raw data in the RawNeRF dataset is excessively noisy for most image denoisers, we opt to directly train RawNeRF models as a denoiser and append a simple finishing ISP after RawNeRF output to generate the input for our method. To stress-test our method, the finishing ISP will scale the RawNeRF output by random digital gains and tone-map the linear space via the sRGB gamma curve or HDR+ local tone mapping \cite{hasinoff2016burst}. Note that for a fair comparison between our method and RawNeRF (Figure 7 in the main text), Restormer \cite{zamir2022restormer} is adopted as the sRGB denoiser in lieu of RawNeRF to generate the input images.

Adobe Lightroom\textregistered is the primary software used in our experiments to adjust and enhance the selected view. We use the ``Masking'' tool to conduct local manipulation. We also use the ``Presets'' to retouch the image. Other than Lightroom\textregistered, HDRNet \cite{gharbi2017deep}, Google Photos\textregistered, etc. are also suitable for our setting.

\begin{figure}[t]
    \centering
    \includegraphics[width=\linewidth]{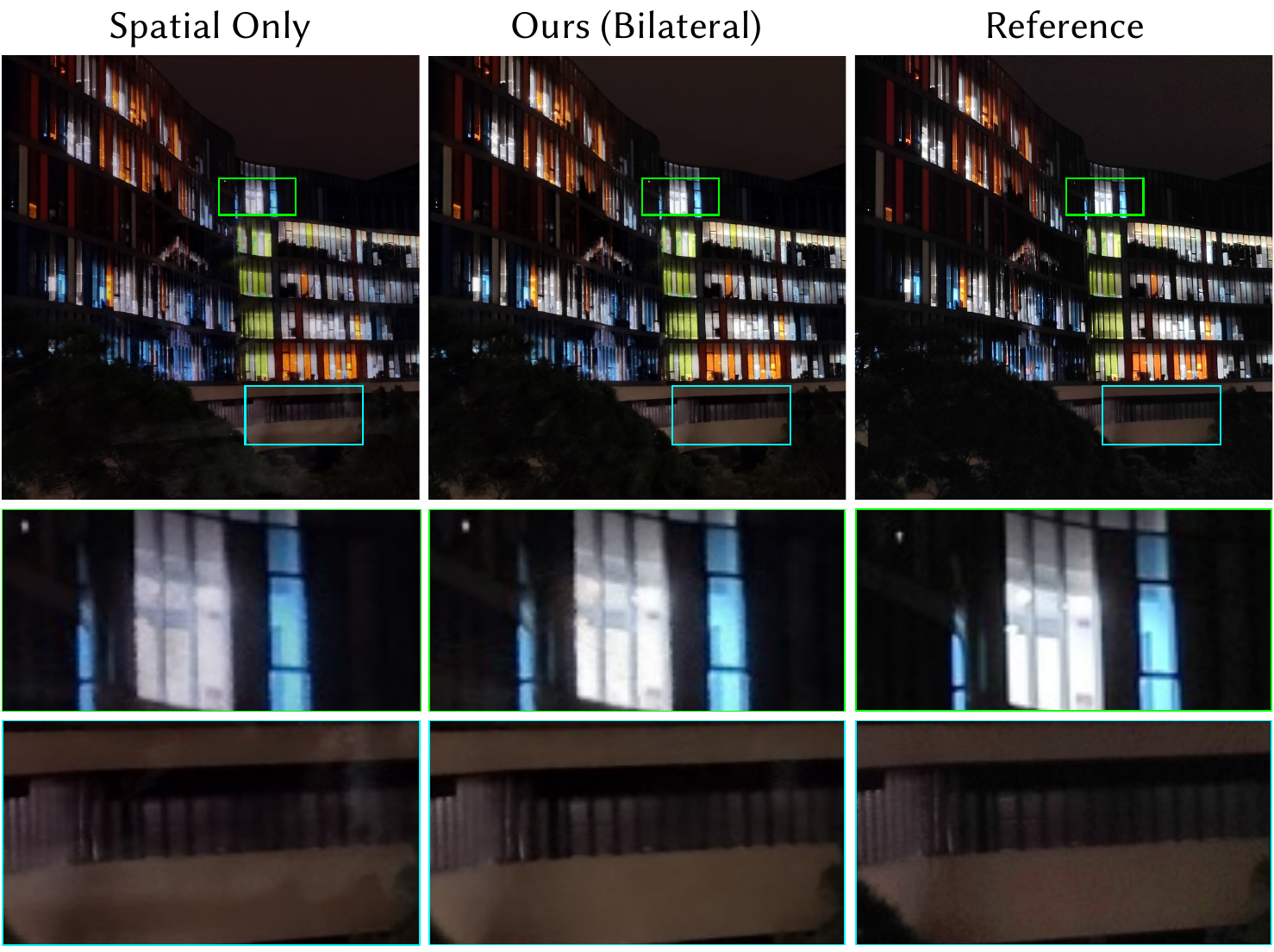}
    \caption{Ablation study on bilateral space processing. The spatial-only affine model fails to preserve the brightness of areas with high local contrast and will result in artifacts in synthesized novel views. While our bilateral space affine model can solve this problem by separately processing darker areas and brighter areas.}
    % \tianfan{The floater is less obvious in a single image rendering. Highlight that the floater is clearer in supplementary video. If the paper is too long, I think we can also just move this to supp.}
    \label{fig:ablation-bila-vs-spatial}
\end{figure}

\section{Additional Ablation Study}

\paragraph{Bilateral Processing.} 
To highlight the advantages of bilateral processing in the training stage, we juxtapose our method with spatial-only processing, where each pixel will be processed only depending on its spatial position. As illustrated in Figure \ref{fig:ablation-bila-vs-spatial}, spatial-only processing exhibits degradation in highly contrasted areas, leading to artifacts in novel view synthesis. This is attributed to the conflicting processing within a local patch for brighter and darker pixels. In contrast, our method treats pixels based on both their spatial positions and values. Therefore, even two neighboring pixels can be processed by different affine models if there is a disparity in their values. In Figure \ref{fig:ablation-bila-finsihing}, we show the advantage of bilateral guided processing over spatial-only processing in the finishing stage. For the first case, without the value dimension, spatial-only manipulation fails to distinguish the editing subject and the neighboring undesired areas. The value dimension in bilateral processing will inform that only the local areas in yellow will be transformed to purple. For the second case, lifting through bilateral space is shown to be more effective for transferring the color schemes from the edited view to the whole scene.

\begin{figure}[H]
    \centering
    \includegraphics[width=\linewidth]{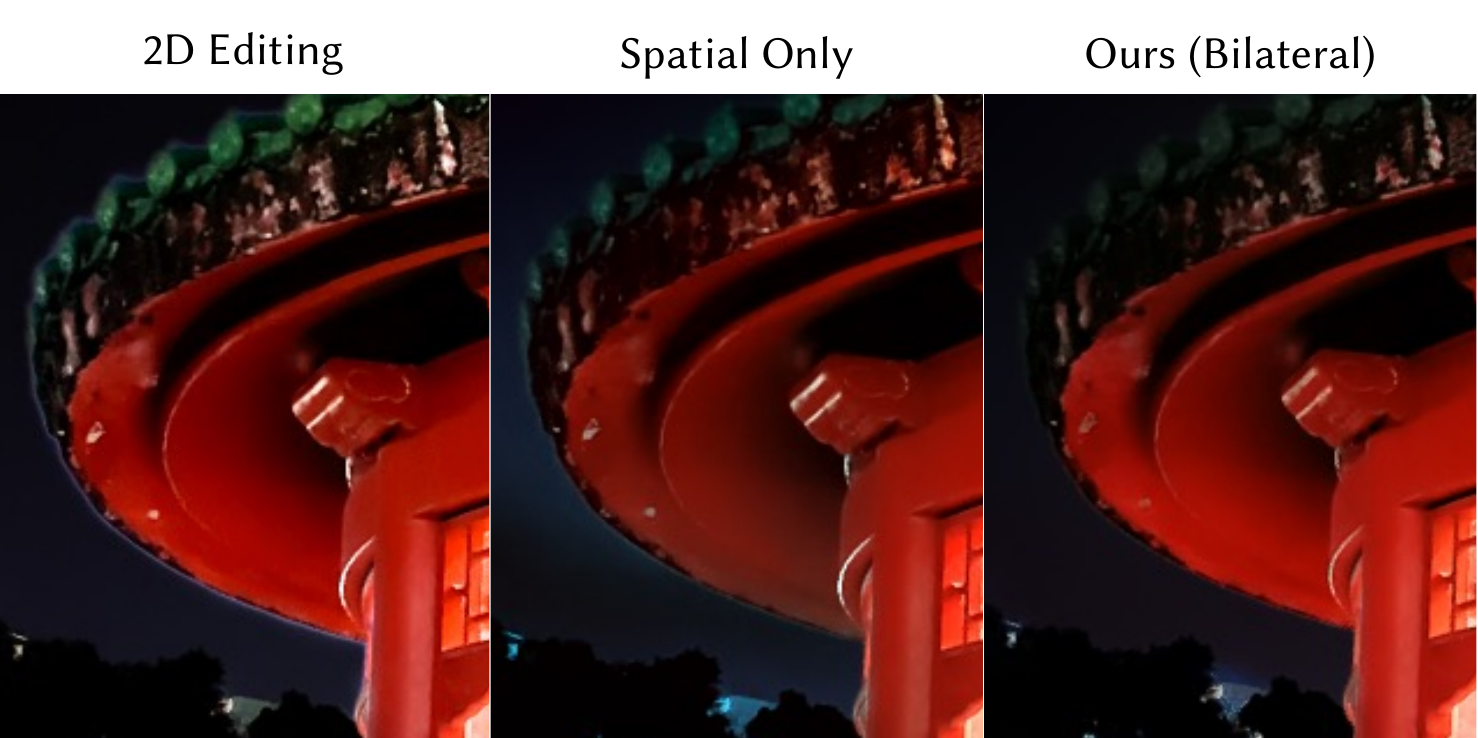}
    \caption{Edge-aware editing results of our bilateral guided 3D finishing. With processing in bilateral space, the ``halo'' on the edge of the foreground and background can be mitigated on the finishing results.}
    \label{fig:halo}
\end{figure}

\paragraph{Edge-aware Editing.}
The bilateral grid is recognized for its edge-aware property. Although this advantage is not very obvious in 3D editing (as the depth information can separate the foreground and background), we find a case where bilateral guided 3D finishing excels in edge-aware editing. In Figure \ref{fig:halo}, brightened foreground and darkened background introduce ``halo'' in the 2D editing. Using spatial-only finishing will leave the ``halo'' on the renderings. In contrast, our bilateral guided approach can alleviate this issue.

\begin{figure}[H]
    \centering
    \includegraphics[width=\linewidth]{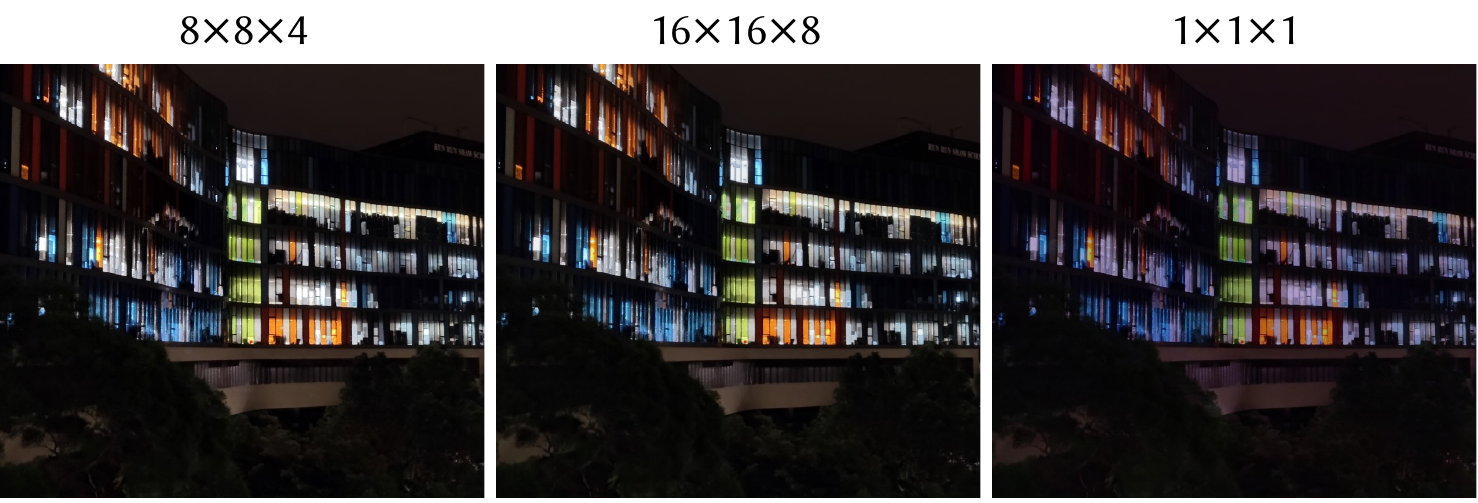}
    \caption{Ablation study on bilateral grid resolutions during the training stage. In this scene, 8$\times$8$\times$4 and 16$\times$16$\times$8 bilateral grids are adequate for approximating ISP. While 1$\times$1$\times$1 cannot yield a correct color.}
    \label{fig:ablation-bila-res}
\end{figure}

\begin{figure*}[t]
    \centering
    \includegraphics[width=\textwidth]{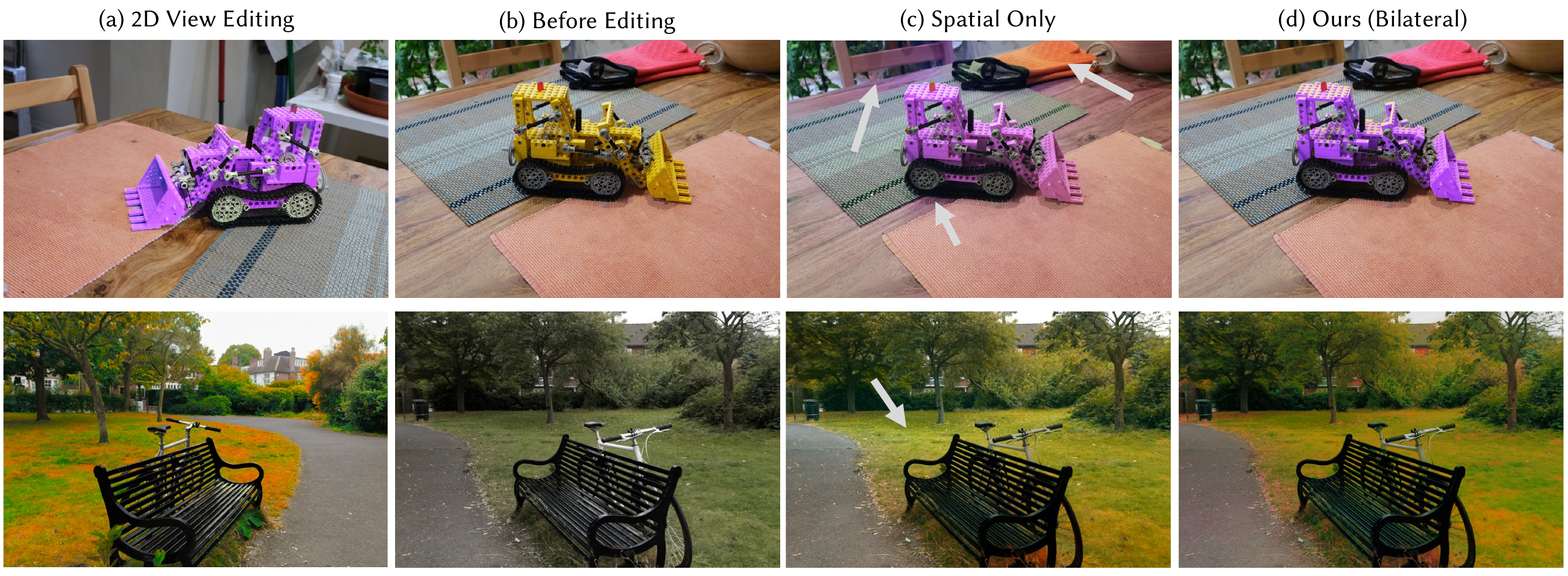}
    \caption{Ablation study on bilateral space processing for 3D radiance-finishing. Filling spatial-only operation for unseen areas using the low-rank prior fails to effectively lift the color scheme in the edited view and may even cause ``color pollution''. With processing in bilateral space, the operation will be filled based on spatial and color dimensions, yielding more desirable results.}
    \label{fig:ablation-bila-finsihing}
\end{figure*}

\begin{figure*}[t]
    \centering
    \includegraphics[width=\textwidth]{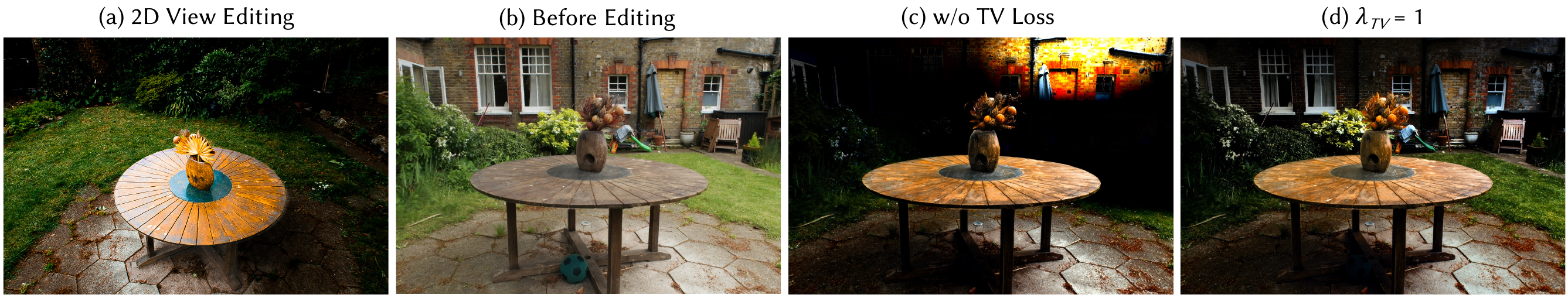}
    \caption{Ablation study on TV loss for bilateral guided 3D radiance-finishing. Omitting TV loss leads to unnatural color schemes in unseen areas. The TV loss can facilitate a smoother lifting of the view editing to novel views.}
    \label{fig:ablation-tvloss-finsihing}
\end{figure*}

\begin{figure}[t]
    \centering
    \includegraphics[width=\linewidth]{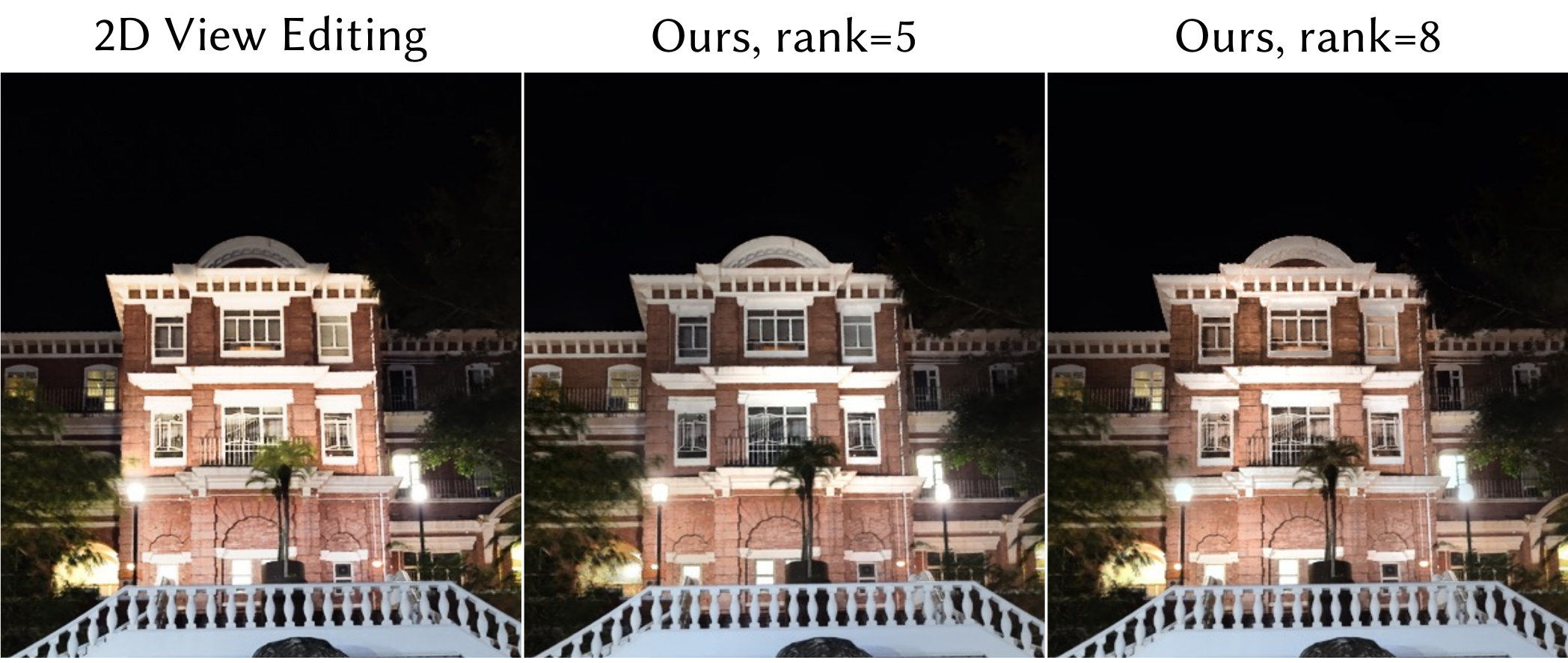}
    \caption{The impact of increasing rank of the 4D bilateral grid. For this editing case with large local contrast changes, a rank-5 4D bilateral grid degenerates. By raising the rank to 8, the editing can be better expressed.}
    \vspace{10px}
    \label{fig:impact-inc-rank}
\end{figure}

\paragraph{Bilateral Grid Resolutions.}
In Figure \ref{fig:ablation-bila-res}, we compare the novel view synthesis results under different resolutions of the bilateral grid. For the first two columns, there is no obvious difference after raising the resolution from $8\times 8 \times 4$ to $16 \times 16 \times 8$ as the local operation is not significant in this case. Once the resolution is reduced $1\times 1 \times 1$ (the third column), the bilateral grid degenerates to a global affine transformation. The color distortion in the result of the third column highlights the importance of approximating local processing using the bilateral grid.

\paragraph{Rank Choice.}
Figure \ref{fig:impact-inc-rank} presents a case where increasing rank can enhance the expressiveness of the 4D bilateral grid for 3D finishing. In this example of target view editing, we brighten the second layer of the building while simultaneously darkening the neighboring third layer, creating a high-contrast effect in the local area. Using a relatively low-rank approximation (rank=5) tends to average the brightness changes across these areas. Increasing the rank to 8 results in a closer 3D retouching to the view editing, preserving finer local changes. In practice, we start by choosing rank=5 and adjust the rank as needed; for instance, if lifting with rank=5 fails to achieve the desired editing, we will choose rank=8.

\paragraph{TV Loss on 4D Bilateral Grid.}
When lifting 2D view editing, we also impose total variation (TV) regularization on the low-rank 4D bilateral grid. Figure \ref{fig:ablation-tvloss-finsihing} illustrates the significance of TV loss in ensuring more natural color schemes in novel views. The TV loss can be regarded as a local prior of smoothness for the low-rank tensor completion of the 4D bilateral grid. Generally, for the finishing stage, we set $\lambda_{TV} = 1$ for all the cases.

\begin{figure}[t]
    \centering
    \begin{subfigure}[a]{\linewidth}
        \centering
        \vspace{10px}
        \includegraphics[width=0.95\linewidth]{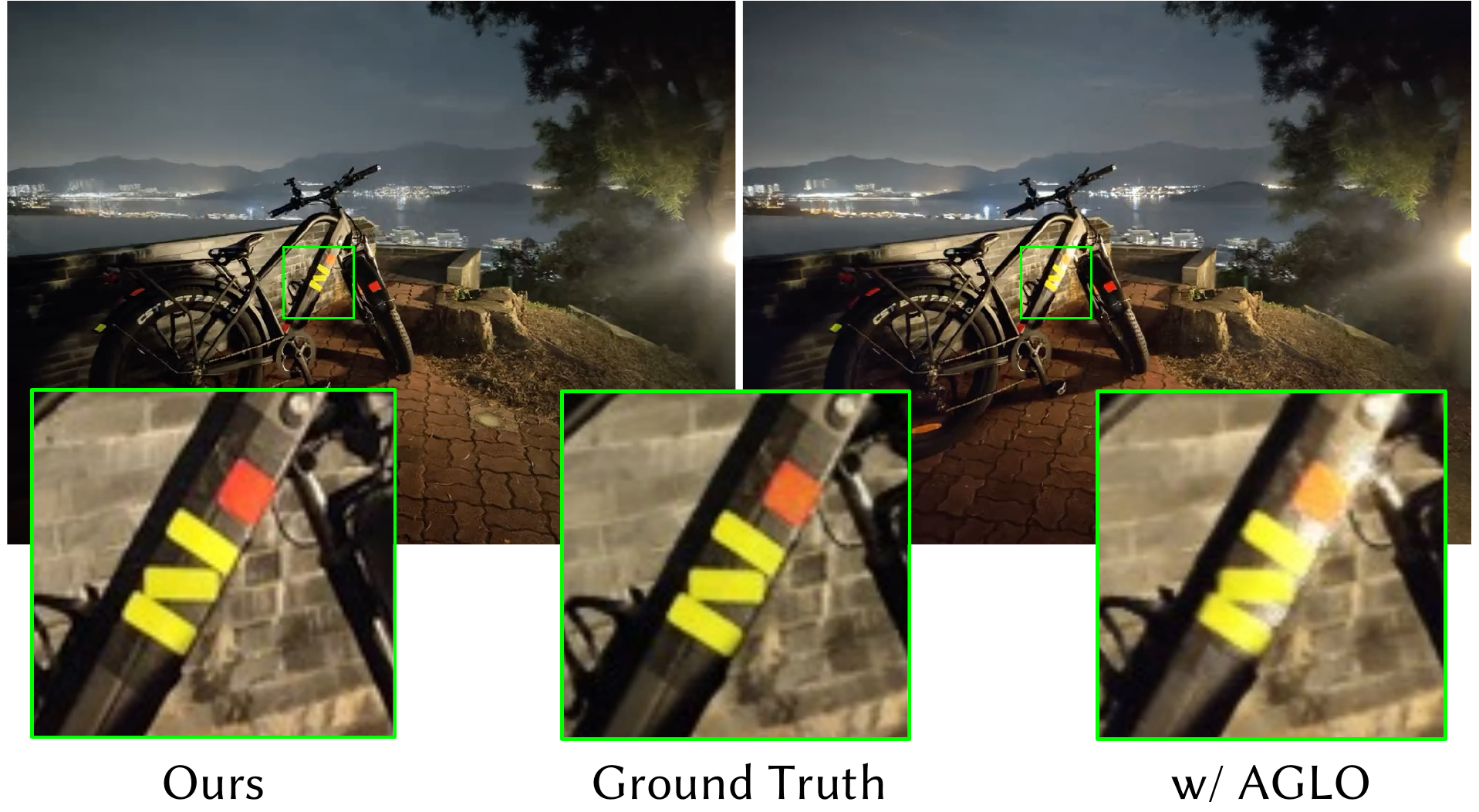}
        \caption{GLO will explain away view-dependent effects. In this case, the specular highlights are baked into the scene.}
        \label{fig:glo-viewdepend}
    \end{subfigure}
    
    \begin{subfigure}[a]{\linewidth}
        \centering
        \vspace{10px}
        \includegraphics[width=0.95\linewidth]{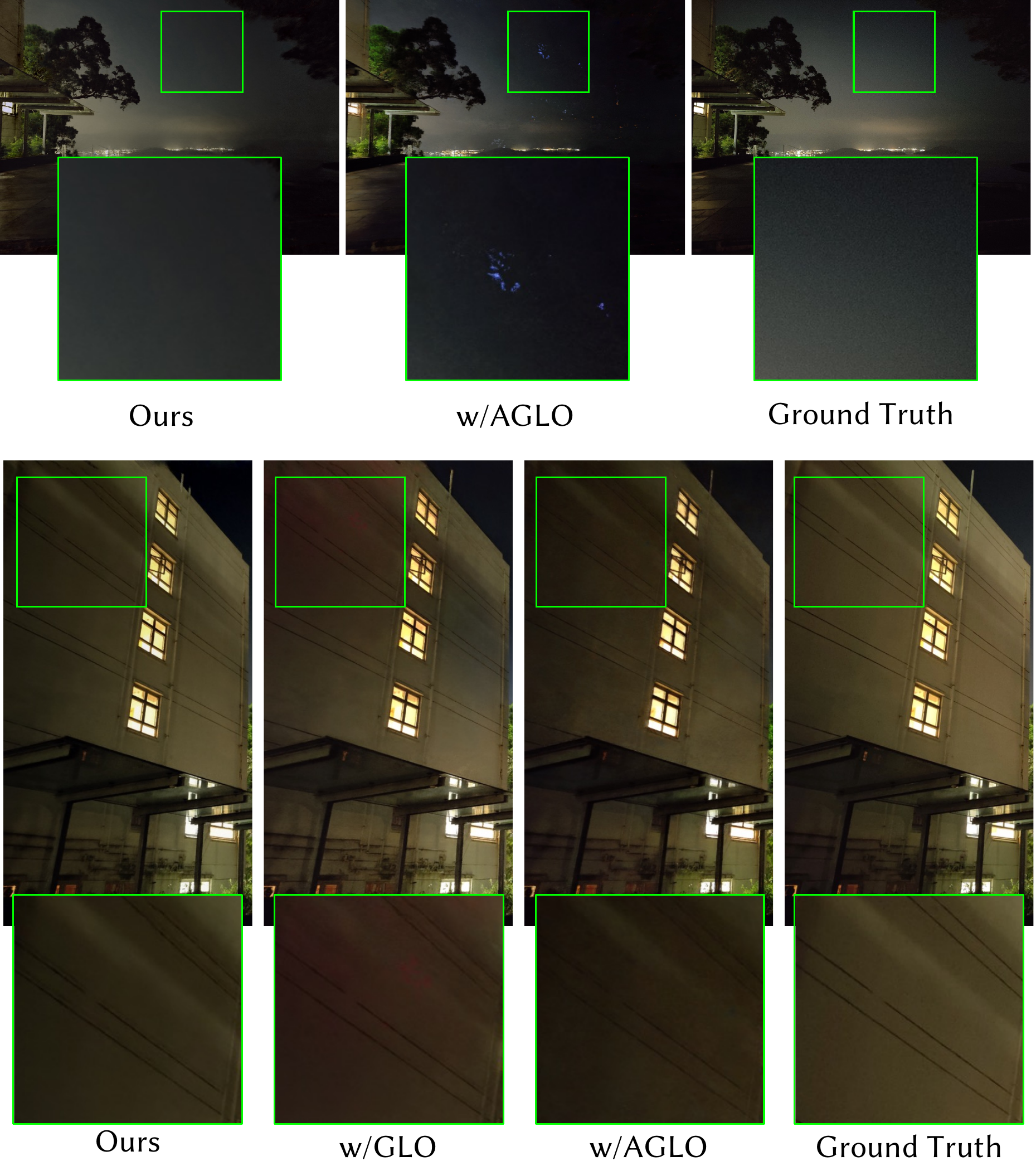}
        \caption{In the low-light environment, GLO will cause artifacts in the dark regions.}
        \label{fig:glo-noisy}
    \end{subfigure}
    \caption{Comparison of GLO-based methods and our method.}
    \label{fig:cmp-glo}
\end{figure}

\section{Additional Comparisons}

% We captured 7 real-world nighttime scenes with photometric variation using cell phones. We compared different methods to handle scenes with photometric variation or low-light conditions, including GLO, Affine GLO, HDRNeRF, LLNeRF, and ZipNeRF. Our bilateral guided radiance field training outperformed all other methods in all metrics, showing superiority over GLO-based approaches. However, HDRNeRF and LLNeRF showed degraded performance. We calculated a per-channel affine transformation to align the output color tints with the ground truth tints. All experiments were evaluated against both output color and color-corrected versions.
We have listed the comprehensive per-scene evaluation results of our method, ZipNeRF (baseline) \cite{barron2023zipnerf}, GLO-based NeRFs \cite{martin2021nerf}, HDRNeRF \cite{huang2022hdr}, and LLNeRF \cite{wang2023lighting} in Tables \ref{table:expanded_numerical}, where PSNR, SSIM, LPIPS scores, and their corresponding affine-aligned metrics are enumerated for each individual scene. Note that the baseline method ZipNeRF outperforms over other compared methods in certain scenes. ZipNeRF tends to model the inconsistent brightness and color changes introduced by camera processing as view-dependent lighting effects. This overfitting may cause the global brightness and color tint to match the ground truth, resulting in high PSNR but visual artifacts. Our disentanglement fixes this camera-introduced inconsistency to improve visual quality, but not necessarily improve PSNR due to the gaps in global brightness and color tint. With affine alignment, the overall tint and brightness are scaled to the same level, making artifacts the major factor affecting PSNR. In this way, CC PSNR reflects the visual quality more accurately, wherein our method achieves the best performance in most cases.

% Those metrics starting with ``CC'' refer to metrics computed on affine-aligned sRGB images, following the conventions in RawNeRF \cite{mildenhall2022nerf} and mip-NeRF 360 \cite{barron2022mip}.

We present a qualitative comparison with GLO-based methods \cite{martin2021nerf,barron2023zipnerf} in Figure \ref{fig:glo-viewdepend} and \ref{fig:glo-noisy}. We find GLO vectors will explain away some view-dependent effects (e.g., specular highlights), which should be primarily encoded by the view directions. Furthermore, in the low-light capture conditions, images will contain significant random noise. GLO vectors will modulate the high-frequency noise, which may introduce artifacts in the dark areas. The bilateral grid is designed to represent ISP, a relatively low-frequency manipulation. Consequently, our bilateral guided training circumvents the aforementioned problems associated with GLO-based approaches.

\section{Learnable Guidance Function}

Recall that we use gray-scale values as the guidance function:
\begin{equation}
\begin{split}
    g(C)=[0.299, 0.587, 0.114] \cdot C.
\end{split}
\label{eqn:bilagrid_nerf_loss}
\end{equation}
In our experiments, we find this function can generally work for the training stage. While in the finishing stage, this function may degenerate. For instance, suppose the green component in the edited view is transferred to red, other color components with the same luminance as the green might be affected by this manipulation due to the low-rank prior. In this scenario, another guidance function that more effectively differentiates between various color components in the edited view is demanded.

\begin{figure*}[t]
    \centering
    \includegraphics[width=\textwidth]{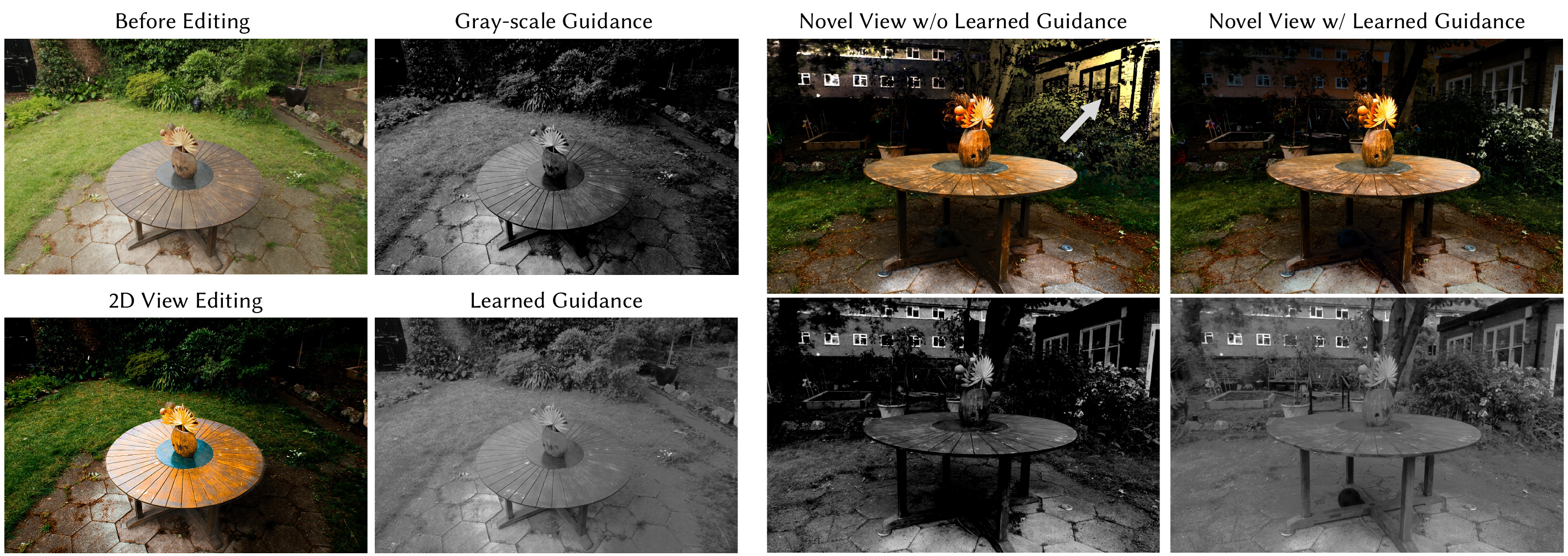}
    \caption{Ablation study on learnable guidance for bilateral guided 3D radiance-finishing. We visualize the guidance map for each rendered view. In this case, using gray-scale guidance will lead to unexpected colors in the novel view. The guidance learned on the edited view can preserve the original hue while adjusting the local brightness.}
    \label{fig:ablation-learngray-finsihing}
\end{figure*}

\begin{figure*}[t]
    \centering
    \includegraphics[width=\textwidth]{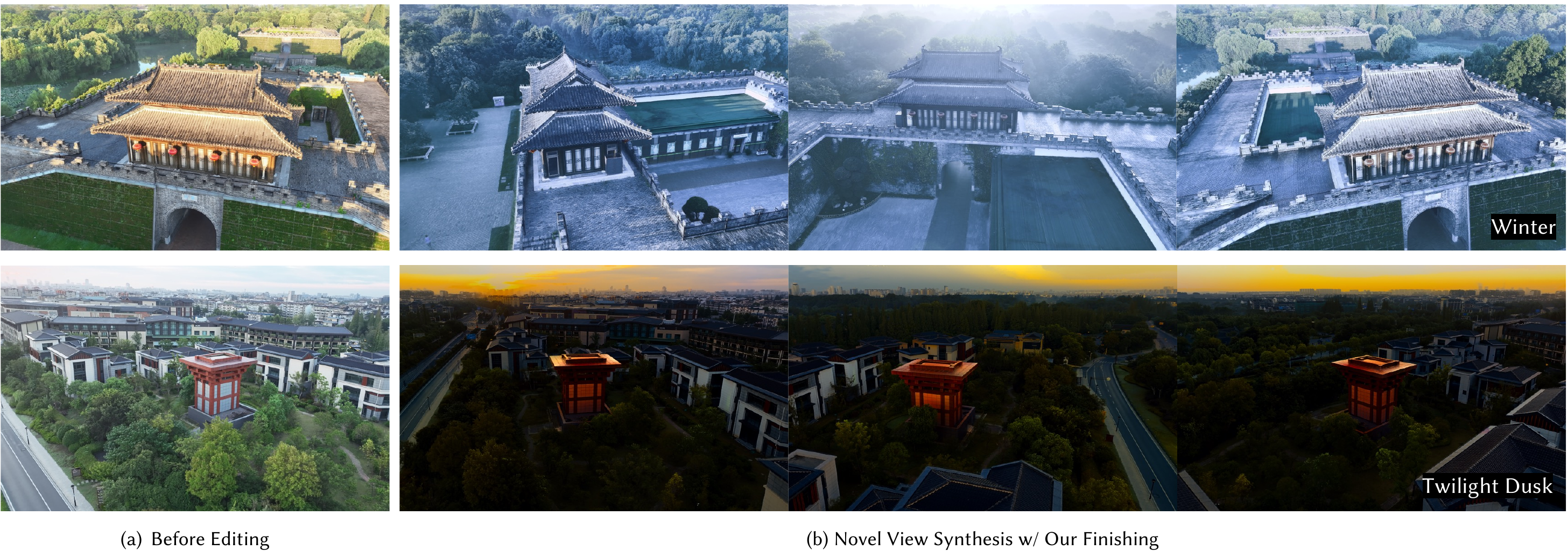}
    \vspace{-10px}
    \caption{Our 3D finishing results of scenes captured with drones. We perform single-view editing and lift the editing to 3D scenes using low-rank 4D bilateral grids. Our approach can stylize radiance fields to varying weather and daytime.}
    \label{fig:drone-results}
\end{figure*}

Taking inspiration from HDRNet \cite{gharbi2017deep}, which learns a pointwise nonlinear guidance function on an image dataset, we propose a similar approach tailored to our context. However, instead of utilizing external data, we aim to fit a simple neural network solely on the edited view. Specifically, the network is a 2-layer MLP. The width of the hidden layer is 8. Then, the guidance map is formulated as:
\begin{equation}
\begin{split}
    g(C)=\operatorname{tanh}\left(2 \cdot \left [ \left (\operatorname{ReLU}(C W_1^T + b_1)\right ) W_2^T + b_2 \right ] \right) / 2 + 0.5,
\end{split}
\label{eqn:bilagrid_nerf_loss}
\end{equation}
where $W_1 \in \mathbb{R}^{8\times 3}$, $b_1\in \mathbb{R}^{8}$, $W_2\in \mathbb{R}^{1\times 8}$, and $b_2\in \mathbb{R}$ are learnable network weights. We empirically find that the learnable guidance function can respond to the color changes in the edited view. Figure~\ref{fig:ablation-learngray-finsihing} presents a comparison between the naive gray-scale guidance and learned guidance. In the view editing, a uniform darkening manipulation is applied to the entire background (including the lawn, floor, etc.). This is reflected on the learned guidance where the contrast of values on the background objects appears lower, compared to the naive gray-scale guidance values. This suggests these areas will be guided to undergo similar transformations, contributing to a more seamless color scheme propagation from the edited 2D view to the whole 3D scene.

Although we find this trick robust for most of our editing cases, overfitting guidance function only on the edited view probably causes potential issues. This simple design of the learnable guidance may not generalize well when faced with quite different color distributions in unseen areas. Thus, we reserve learnable guidance as a direction for further exploration.

\section{Results on Drone-Captured Scenes}

In Figure \ref{fig:drone-results}, we present our finishing results for scenes larger than those previously showcased ones. We use two drone videos from the DL3DV-10K dataset \cite{ling2023dl3dv}. After optimizing a radiance field for each scene, we retouch selected views to change their weather and daytime settings. Then we optimize low-rank bilateral grids to lift the view editing. The results show that our finishing method can work consistently on the drone-captured scenes.

The second case in Figure \ref{fig:drone-results} is challenging as the processing applied to the red temple differs significantly from the processing applied to the sky (sunset) and other areas (dimmed). To effectively address these variations, we set up a separate 4D bilateral grid for the red temple and another bilateral grid for the remaining areas. To support more challenging editing, a more advanced design for bilateral grids is demanded.

\begin{table*}[t]
\caption{The expansion of Table 1 from the main paper. We report PSNR, SSIM, and LPIPS metrics for each scene separately. The ``CC'' prefix suggests the metrics are computed on affine-aligned sRGB images.}
\begin{tabular}{l|l|l|l|l|l|l|l||l}
\multicolumn{9}{c}{PSNR} \\ \hline
           & PondBike & ChineseArch & NighttimePond & Statue & Building & Strat  & LionPavilion & \textbf{average}     \\ \hline
ZipNeRF        & \rc26.5908  & 18.0626     & \rc25.8513   & 21.3064   & 23.3193   & \rc21.4395 & \oc24.9181      & \yc23.0697 \\
ZipNeRF w/GLO  & \yc25.5925  & \yc20.7079     & 22.9177   & \yc23.1146   & \oc26.1209   & 19.1329 & \yc24.4943      & \oc23.1544 \\
ZipNeRF w/AGLO & 25.5710  & \oc21.2042     & 21.9639   & 21.1405   & \yc23.8586   & 18.4416 & 22.7012      & 22.1259 \\ \hline
HDRNeRF        & 25.5923  & 18.1798     & \oc24.1734   & \rc23.9950   & 22.9460   & \oc20.3188 & 22.6953      & 22.5572 \\
LLNeRF         & 24.0634  & 18.4498     & 21.3729   & 18.3786   & 17.7224   & 7.7494  & 20.8351      & 18.3674 \\ \hline
Ours           & \oc26.1644  & \rc21.5463     & \yc23.8502   & \oc23.3337   & \rc26.4077   & \yc19.6839 & \rc25.7971      & \rc23.8262 \\ \hline
\end{tabular} \\ \hfill \break

\begin{tabular}{l|l|l|l|l|l|l|l||l}
\multicolumn{9}{c}{CC PSNR} \\ \hline
        & PondBike & ChineseArch & NighttimePond & Statue & Building & Strat  & LionPavilion & \textbf{average}     \\ \hline
ZipNeRF        & \cellcolor{yellow!30}28.0486  & 21.4935     & \cellcolor{red!30}26.5362   & 23.7384   & 26.8481   & 23.5512 & \cellcolor{orange!30}25.9849      & 25.1716 \\ 
ZipNeRF w/GLO  & \cellcolor{orange!30}28.4127  & \cellcolor{orange!30}25.4512     & \cellcolor{yellow!30}24.7949   & \cellcolor{orange!30}25.9525   & \cellcolor{orange!30}29.8511   & \cellcolor{orange!30}25.4714 & \cellcolor{yellow!30}25.8575      & \cellcolor{orange!30}26.5416 \\ 
ZipNeRF w/AGLO & 27.6385  & \cellcolor{yellow!30}25.1003     & 24.1903   & \cellcolor{yellow!30}23.9220   & \cellcolor{yellow!30}27.9123   & \cellcolor{yellow!30}24.7473 & 24.1891      & \cellcolor{yellow!30}25.3857 \\ \hline
HDRNeRF        & 26.6960  & 22.0012     & 25.4842   & 24.0930   & 24.8259   & 22.1104 & 23.0263      & 24.0339 \\ 
LLNeRF         & 25.9435  & 21.1180     & 25.9706   & 21.0700   & 22.1995   & 13.7086 & 21.9384      & 21.7069 \\ \hline
Ours           & \cellcolor{red!30}29.6209  & \cellcolor{red!30}26.6002     & \cellcolor{orange!30}26.4227   & \cellcolor{red!30}26.5037   & \cellcolor{red!30}30.8510   & \cellcolor{red!30}25.8521 & \cellcolor{red!30}26.8258      & \cellcolor{red!30}27.5252 \\ \hline
\end{tabular} \\ \hfill \break

\begin{tabular}{l|l|l|l|l|l|l|l||l}
\multicolumn{9}{c}{SSIM} \\ \hline
        & PondBike & ChineseArch & NighttimePond & Statue & Building & Strat  & LionPavilion & \textbf{average}    \\ \hline
ZipNeRF        & \yc0.8559   & 0.6789      & \rc0.8524    & \yc0.7488    & \yc0.8420    & \oc0.8504 & \oc0.8088       & \yc0.8053 \\
ZipNeRF w/GLO  & \rc0.8655   & \oc0.7629      & \oc0.8288    & \rc0.8283    & \rc0.8948    & \yc0.8406 & \yc0.7955       & \rc0.8309 \\
ZipNeRF w/AGLO & 0.8398   & \yc0.7397      & 0.7784    & 0.7468    & 0.8378    & 0.8029 & 0.7420       & 0.7839 \\ \hline
HDRNeRF        & 0.7728   & 0.4797      & \yc0.8120    & 0.5590    & 0.7068    & 0.4692 & 0.6381       & 0.6339 \\
LLNeRF         & 0.7330   & 0.5662      & 0.8040    & 0.5288    & 0.6852    & 0.4766 & 0.6393       & 0.6333 \\ \hline
Ours           & \oc0.8588   & \rc0.7862      & 0.7618    & \oc0.8264    & \oc0.8921    & \rc0.8531 & \rc0.8363       & \oc0.8307 \\ \hline
\end{tabular} \\ \hfill \break

\begin{tabular}{l|l|l|l|l|l|l|l||l}
\multicolumn{9}{c}{CC SSIM} \\ \hline
       & PondBike & ChineseArch & NighttimePond & Statue & Building & Strat  & LionPavilion & \textbf{average}    \\ \hline
ZipNeRF        & \yc0.8682   & 0.7233      & \oc0.8544    & 0.7735    & \yc0.8756    & \yc0.8552 & \oc0.8070       & \yc0.8225 \\
ZipNeRF w/GLO  & \oc0.8806   & \oc0.8153      & \yc0.8280    & \oc0.8475    & \oc0.9116    & \cellcolor{red!30}0.8814 & \yc0.8004       & \oc0.8521 \\
ZipNeRF w/AGLO & 0.8566   & \yc0.7800      & 0.8095    & \yc0.7784    & 0.8638    & 0.8532 & 0.7675       & 0.8156 \\ \hline
HDRNeRF        & 0.7837   & 0.5320      & 0.8182    & 0.5610    & 0.7204    & 0.4856 & 0.6151       & 0.6451 \\
LLNeRF         & 0.7410   & 0.6055      & 0.8080    & 0.5748    & 0.7393    & 0.5604 & 0.5926       & 0.6602 \\ \hline
Ours           & \cellcolor{red!30}0.8851   & \cellcolor{red!30}0.8291      & \cellcolor{red!30}0.8593    & \cellcolor{red!30}0.8550    & \cellcolor{red!30}0.9246    & \oc0.8800 & \cellcolor{red!30}0.8544       & \cellcolor{red!30}0.8696 \\ \hline
\end{tabular} \\ \hfill \break

\begin{tabular}{l|l|l|l|l|l|l|l||l}
\multicolumn{9}{c}{LPIPS} \\ \hline
        & PondBike & ChineseArch & NighttimePond & Statue & Building & Strat  & LionPavilion & \textbf{average}    \\ \hline
ZipNeRF                    & \yc0.2067   & 0.2267      & \oc0.2185    & \yc0.1933    & \yc0.1221    & \yc0.2064 & \yc0.1274       & \yc0.1859 \\
ZipNeRF w/GLO              & \rc0.1855   & \rc0.1462      & \yc0.2342    & \oc0.1374    & \oc0.0893    & \rc0.1559 & \oc0.1186       & \oc0.1525 \\
ZipNeRF w/AGLO             & 0.2136   & \yc0.2069      & 0.3060    & 0.2284    & 0.1693    & 0.2103 & 0.1768       & 0.2159 \\ \hline
HDRNeRF                    & 0.3900     & 0.5888      & 0.3899    & 0.5432    & 0.3682    & 0.2749 & 0.3157       & 0.4101 \\
LLNeRF                     & 0.3695   & 0.5015      & 0.3711    & 0.5373    & 0.3514    & 0.6027 & 0.3607       & 0.4420  \\ \hline
Ours & \oc0.1879   & \oc0.1507      & \rc0.2091    & \rc0.1368    & \rc0.0863    & \oc0.1634 & \rc0.1029       & \rc0.1482 \\ \hline
\end{tabular} \\ \hfill \break

\begin{tabular}{l|l|l|l|l|l|l|l||l}
\multicolumn{9}{c}{CC LPIPS} \\ \hline
       & PondBike & ChineseArch & NighttimePond & Statue & Building & Strat  & LionPavilion & \textbf{average}    \\ \hline
ZipNeRF        & \yc0.2008   & 0.2211      & \oc0.2236    & \yc0.1959    & \yc0.1216    & 0.1899 & \yc0.1229       & \yc0.1823 \\
ZipNeRF w/GLO  & \oc0.1794   & \rc0.1266      & \yc0.2551    & \oc0.1294    & \oc0.0852    & \rc0.1185 & \oc0.1128       & \oc0.1439 \\
ZipNeRF w/AGLO & 0.2042   & \yc0.1912      & 0.3050    & 0.2153    & 0.1505    & \yc0.1709 & 0.1649       & 0.2003 \\ \hline
HDRNeRF        & 0.3679   & 0.5603      & 0.3633    & 0.5275    & 0.3644    & 0.2672 & 0.3151       & 0.3951 \\
LLNeRF         & 0.3651   & 0.4999      & 0.3733    & 0.5452    & 0.3509    & 0.6388 & 0.3530       & 0.4466 \\ \hline
Ours           & \rc0.1737   & \oc0.1314      & \rc0.2054    & \rc0.1270    & \rc0.0811    & \oc0.1211 & \rc0.0954       & \rc0.1336 \\ \hline
\end{tabular}
\label{table:expanded_numerical}
\end{table*}

%% file: main.bbl

\begin{thebibliography}{68}

%%% the Web address.

\ifx \showCODEN    \undefined \def \showCODEN     #1{\unskip}     \fi
\ifx \showDOI      \undefined \def \showDOI       #1{#1}\fi
\ifx \showISBNx    \undefined \def \showISBNx     #1{\unskip}     \fi
\ifx \showISBNxiii \undefined \def \showISBNxiii  #1{\unskip}     \fi
\ifx \showISSN     \undefined \def \showISSN      #1{\unskip}     \fi
\ifx \showLCCN     \undefined \def \showLCCN      #1{\unskip}     \fi
\ifx \shownote     \undefined \def \shownote      #1{#1}          \fi
\ifx \showarticletitle \undefined \def \showarticletitle #1{#1}   \fi
\ifx \showURL      \undefined \def \showURL       {\relax}        \fi
% invisible to TeX
\providecommand\bibfield[2]{#2}
\providecommand\bibinfo[2]{#2}
\providecommand\natexlab[1]{#1}
\providecommand\showeprint[2][]{arXiv:#2}

\bibitem[Barron et~al\mbox{.}(2015)]%
        {barron2015fast}
\bibfield{author}{\bibinfo{person}{Jonathan~T Barron}, \bibinfo{person}{Andrew Adams}, \bibinfo{person}{YiChang Shih}, {and} \bibinfo{person}{Carlos Hern{\'a}ndez}.} \bibinfo{year}{2015}\natexlab{}.
\newblock \showarticletitle{Fast bilateral-space stereo for synthetic defocus}. In \bibinfo{booktitle}{\emph{Proceedings of the IEEE Conference on Computer Vision and Pattern Recognition}}. \bibinfo{pages}{4466--4474}.
\newblock


\bibitem[Barron et~al\mbox{.}(2022)]%
        {barron2022mip}
\bibfield{author}{\bibinfo{person}{Jonathan~T Barron}, \bibinfo{person}{Ben Mildenhall}, \bibinfo{person}{Dor Verbin}, \bibinfo{person}{Pratul~P Srinivasan}, {and} \bibinfo{person}{Peter Hedman}.} \bibinfo{year}{2022}\natexlab{}.
\newblock \showarticletitle{Mip-nerf 360: Unbounded anti-aliased neural radiance fields}. In \bibinfo{booktitle}{\emph{Proceedings of the IEEE/CVF Conference on Computer Vision and Pattern Recognition}}. \bibinfo{pages}{5470--5479}.
\newblock


\bibitem[Barron et~al\mbox{.}(2023)]%
        {barron2023zipnerf}
\bibfield{author}{\bibinfo{person}{Jonathan~T. Barron}, \bibinfo{person}{Ben Mildenhall}, \bibinfo{person}{Dor Verbin}, \bibinfo{person}{Pratul~P. Srinivasan}, {and} \bibinfo{person}{Peter Hedman}.} \bibinfo{year}{2023}\natexlab{}.
\newblock \showarticletitle{Zip-NeRF: Anti-Aliased Grid-Based Neural Radiance Fields}.
\newblock \bibinfo{journal}{\emph{ICCV}} (\bibinfo{year}{2023}).
\newblock


\bibitem[Barron and Poole(2016)]%
        {barron2016fast}
\bibfield{author}{\bibinfo{person}{Jonathan~T Barron} {and} \bibinfo{person}{Ben Poole}.} \bibinfo{year}{2016}\natexlab{}.
\newblock \showarticletitle{The fast bilateral solver}. In \bibinfo{booktitle}{\emph{European conference on computer vision}}. Springer, \bibinfo{pages}{617--632}.
\newblock


\bibitem[Bojanowski et~al\mbox{.}(2017)]%
        {bojanowski2017optimizing}
\bibfield{author}{\bibinfo{person}{Piotr Bojanowski}, \bibinfo{person}{Armand Joulin}, \bibinfo{person}{David Lopez-Paz}, {and} \bibinfo{person}{Arthur Szlam}.} \bibinfo{year}{2017}\natexlab{}.
\newblock \showarticletitle{Optimizing the latent space of generative networks}.
\newblock \bibinfo{journal}{\emph{arXiv preprint arXiv:1707.05776}} (\bibinfo{year}{2017}).
\newblock


\bibitem[Boss et~al\mbox{.}(2022)]%
        {boss2022samurai}
\bibfield{author}{\bibinfo{person}{Mark Boss}, \bibinfo{person}{Andreas Engelhardt}, \bibinfo{person}{Abhishek Kar}, \bibinfo{person}{Yuanzhen Li}, \bibinfo{person}{Deqing Sun}, \bibinfo{person}{Jonathan Barron}, \bibinfo{person}{Hendrik Lensch}, {and} \bibinfo{person}{Varun Jampani}.} \bibinfo{year}{2022}\natexlab{}.
\newblock \showarticletitle{Samurai: Shape and material from unconstrained real-world arbitrary image collections}.
\newblock \bibinfo{journal}{\emph{Advances in Neural Information Processing Systems}}  \bibinfo{volume}{35} (\bibinfo{year}{2022}), \bibinfo{pages}{26389--26403}.
\newblock


\bibitem[Brooks et~al\mbox{.}(2023)]%
        {brooks2023instructpix2pix}
\bibfield{author}{\bibinfo{person}{Tim Brooks}, \bibinfo{person}{Aleksander Holynski}, {and} \bibinfo{person}{Alexei~A Efros}.} \bibinfo{year}{2023}\natexlab{}.
\newblock \showarticletitle{Instructpix2pix: Learning to follow image editing instructions}. In \bibinfo{booktitle}{\emph{Proceedings of the IEEE/CVF Conference on Computer Vision and Pattern Recognition}}. \bibinfo{pages}{18392--18402}.
\newblock


\bibitem[Brooks et~al\mbox{.}(2019)]%
        {brooks2019unprocessing}
\bibfield{author}{\bibinfo{person}{Tim Brooks}, \bibinfo{person}{Ben Mildenhall}, \bibinfo{person}{Tianfan Xue}, \bibinfo{person}{Jiawen Chen}, \bibinfo{person}{Dillon Sharlet}, {and} \bibinfo{person}{Jonathan~T Barron}.} \bibinfo{year}{2019}\natexlab{}.
\newblock \showarticletitle{Unprocessing images for learned raw denoising}. In \bibinfo{booktitle}{\emph{Proceedings of the IEEE/CVF Conference on Computer Vision and Pattern Recognition}}. \bibinfo{pages}{11036--11045}.
\newblock


\bibitem[Bychkovsky et~al\mbox{.}(2011)]%
        {bychkovsky2011learning}
\bibfield{author}{\bibinfo{person}{Vladimir Bychkovsky}, \bibinfo{person}{Sylvain Paris}, \bibinfo{person}{Eric Chan}, {and} \bibinfo{person}{Fr{\'e}do Durand}.} \bibinfo{year}{2011}\natexlab{}.
\newblock \showarticletitle{Learning photographic global tonal adjustment with a database of input/output image pairs}. In \bibinfo{booktitle}{\emph{CVPR 2011}}. IEEE, \bibinfo{pages}{97--104}.
\newblock


\bibitem[Candes and Recht(2012)]%
        {candes2012exact}
\bibfield{author}{\bibinfo{person}{Emmanuel Candes} {and} \bibinfo{person}{Benjamin Recht}.} \bibinfo{year}{2012}\natexlab{}.
\newblock \showarticletitle{Exact matrix completion via convex optimization}.
\newblock \bibinfo{journal}{\emph{Commun. ACM}} \bibinfo{volume}{55}, \bibinfo{number}{6} (\bibinfo{year}{2012}), \bibinfo{pages}{111--119}.
\newblock


\bibitem[Candes and Plan(2010)]%
        {candes2010matrix}
\bibfield{author}{\bibinfo{person}{Emmanuel~J Candes} {and} \bibinfo{person}{Yaniv Plan}.} \bibinfo{year}{2010}\natexlab{}.
\newblock \showarticletitle{Matrix completion with noise}.
\newblock \bibinfo{journal}{\emph{Proc. IEEE}} \bibinfo{volume}{98}, \bibinfo{number}{6} (\bibinfo{year}{2010}), \bibinfo{pages}{925--936}.
\newblock


\bibitem[Caron et~al\mbox{.}(2021)]%
        {caron2021emerging}
\bibfield{author}{\bibinfo{person}{Mathilde Caron}, \bibinfo{person}{Hugo Touvron}, \bibinfo{person}{Ishan Misra}, \bibinfo{person}{Herv{\'e} J{\'e}gou}, \bibinfo{person}{Julien Mairal}, \bibinfo{person}{Piotr Bojanowski}, {and} \bibinfo{person}{Armand Joulin}.} \bibinfo{year}{2021}\natexlab{}.
\newblock \showarticletitle{Emerging properties in self-supervised vision transformers}. In \bibinfo{booktitle}{\emph{Proceedings of the IEEE/CVF international conference on computer vision}}. \bibinfo{pages}{9650--9660}.
\newblock


\bibitem[Carroll and Chang(1970)]%
        {carroll1970analysis}
\bibfield{author}{\bibinfo{person}{J~Douglas Carroll} {and} \bibinfo{person}{Jih-Jie Chang}.} \bibinfo{year}{1970}\natexlab{}.
\newblock \showarticletitle{Analysis of individual differences in multidimensional scaling via an N-way generalization of “Eckart-Young” decomposition}.
\newblock \bibinfo{journal}{\emph{Psychometrika}} \bibinfo{volume}{35}, \bibinfo{number}{3} (\bibinfo{year}{1970}), \bibinfo{pages}{283--319}.
\newblock


\bibitem[Chen et~al\mbox{.}(2022)]%
        {chen2022tensorf}
\bibfield{author}{\bibinfo{person}{Anpei Chen}, \bibinfo{person}{Zexiang Xu}, \bibinfo{person}{Andreas Geiger}, \bibinfo{person}{Jingyi Yu}, {and} \bibinfo{person}{Hao Su}.} \bibinfo{year}{2022}\natexlab{}.
\newblock \showarticletitle{Tensorf: Tensorial radiance fields}. In \bibinfo{booktitle}{\emph{European Conference on Computer Vision}}. Springer, \bibinfo{pages}{333--350}.
\newblock


\bibitem[Chen et~al\mbox{.}(2016)]%
        {chen2016bilateral}
\bibfield{author}{\bibinfo{person}{Jiawen Chen}, \bibinfo{person}{Andrew Adams}, \bibinfo{person}{Neal Wadhwa}, {and} \bibinfo{person}{Samuel~W Hasinoff}.} \bibinfo{year}{2016}\natexlab{}.
\newblock \showarticletitle{Bilateral guided upsampling}.
\newblock \bibinfo{journal}{\emph{ACM Transactions on Graphics (TOG)}} \bibinfo{volume}{35}, \bibinfo{number}{6} (\bibinfo{year}{2016}), \bibinfo{pages}{1--8}.
\newblock


\bibitem[Chen et~al\mbox{.}(2007)]%
        {chen2007real}
\bibfield{author}{\bibinfo{person}{Jiawen Chen}, \bibinfo{person}{Sylvain Paris}, {and} \bibinfo{person}{Fr{\'e}do Durand}.} \bibinfo{year}{2007}\natexlab{}.
\newblock \showarticletitle{Real-time edge-aware image processing with the bilateral grid}.
\newblock \bibinfo{journal}{\emph{ACM Transactions on Graphics (TOG)}} \bibinfo{volume}{26}, \bibinfo{number}{3} (\bibinfo{year}{2007}), \bibinfo{pages}{103--es}.
\newblock


\bibitem[Conde et~al\mbox{.}(2022)]%
        {conde2022model}
\bibfield{author}{\bibinfo{person}{Marcos~V Conde}, \bibinfo{person}{Steven McDonagh}, \bibinfo{person}{Matteo Maggioni}, \bibinfo{person}{Ales Leonardis}, {and} \bibinfo{person}{Eduardo P{\'e}rez-Pellitero}.} \bibinfo{year}{2022}\natexlab{}.
\newblock \showarticletitle{Model-based image signal processors via learnable dictionaries}. In \bibinfo{booktitle}{\emph{Proceedings of the AAAI Conference on Artificial Intelligence}}, Vol.~\bibinfo{volume}{36}. \bibinfo{pages}{481--489}.
\newblock


\bibitem[Durand and Dorsey(2002)]%
        {durand2002fast}
\bibfield{author}{\bibinfo{person}{Fr{\'e}do Durand} {and} \bibinfo{person}{Julie Dorsey}.} \bibinfo{year}{2002}\natexlab{}.
\newblock \showarticletitle{Fast bilateral filtering for the display of high-dynamic-range images}. In \bibinfo{booktitle}{\emph{Proceedings of the 29th annual conference on Computer graphics and interactive techniques}}. \bibinfo{pages}{257--266}.
\newblock


\bibitem[Fan et~al\mbox{.}(2022)]%
        {fan2022unified}
\bibfield{author}{\bibinfo{person}{Zhiwen Fan}, \bibinfo{person}{Yifan Jiang}, \bibinfo{person}{Peihao Wang}, \bibinfo{person}{Xinyu Gong}, \bibinfo{person}{Dejia Xu}, {and} \bibinfo{person}{Zhangyang Wang}.} \bibinfo{year}{2022}\natexlab{}.
\newblock \showarticletitle{Unified implicit neural stylization}. In \bibinfo{booktitle}{\emph{European Conference on Computer Vision}}. Springer, \bibinfo{pages}{636--654}.
\newblock


\bibitem[Gharbi et~al\mbox{.}(2017)]%
        {gharbi2017deep}
\bibfield{author}{\bibinfo{person}{Micha{\"e}l Gharbi}, \bibinfo{person}{Jiawen Chen}, \bibinfo{person}{Jonathan~T Barron}, \bibinfo{person}{Samuel~W Hasinoff}, {and} \bibinfo{person}{Fr{\'e}do Durand}.} \bibinfo{year}{2017}\natexlab{}.
\newblock \showarticletitle{Deep bilateral learning for real-time image enhancement}.
\newblock \bibinfo{journal}{\emph{ACM Transactions on Graphics (TOG)}} \bibinfo{volume}{36}, \bibinfo{number}{4} (\bibinfo{year}{2017}), \bibinfo{pages}{1--12}.
\newblock


\bibitem[Gong et~al\mbox{.}(2023a)]%
        {gong2023recolornerf}
\bibfield{author}{\bibinfo{person}{Bingchen Gong}, \bibinfo{person}{Yuehao Wang}, \bibinfo{person}{Xiaoguang Han}, {and} \bibinfo{person}{Qi Dou}.} \bibinfo{year}{2023}\natexlab{a}.
\newblock \showarticletitle{RecolorNeRF: Layer Decomposed Radiance Fields for Efficient Color Editing of 3D Scenes}. In \bibinfo{booktitle}{\emph{Proceedings of the 31st ACM International Conference on Multimedia}} (, Ottawa ON, Canada,) \emph{(\bibinfo{series}{MM '23})}. \bibinfo{publisher}{Association for Computing Machinery}, \bibinfo{address}{New York, NY, USA}, \bibinfo{pages}{8004–8015}.
\newblock
\showISBNx{9798400701085}
\urldef\tempurl%
\url{https://doi.org/10.1145/3581783.3611957}
\showDOI{\tempurl}


\bibitem[Gong et~al\mbox{.}(2023b)]%
        {gong2023seamlessnerf}
\bibfield{author}{\bibinfo{person}{Bingchen Gong}, \bibinfo{person}{Yuehao Wang}, \bibinfo{person}{Xiaoguang Han}, {and} \bibinfo{person}{Qi Dou}.} \bibinfo{year}{2023}\natexlab{b}.
\newblock \showarticletitle{SeamlessNeRF: Stitching Part NeRFs with Gradient Propagation}. In \bibinfo{booktitle}{\emph{SIGGRAPH Asia 2023 Conference Papers}}. \bibinfo{pages}{1--10}.
\newblock


\bibitem[Gu(2022)]%
        {zipnerf-pytorch}
\bibfield{author}{\bibinfo{person}{Chun Gu}.} \bibinfo{year}{2022}\natexlab{}.
\newblock \bibinfo{title}{ZipNeRF-PyTorch}.
\newblock
\newblock
\newblock
\shownote{https://github.com/SuLvXiangXin/zipnerf-pytorch}.


\bibitem[Haque et~al\mbox{.}(2023)]%
        {instructnerf2023}
\bibfield{author}{\bibinfo{person}{Ayaan Haque}, \bibinfo{person}{Matthew Tancik}, \bibinfo{person}{Alexei Efros}, \bibinfo{person}{Aleksander Holynski}, {and} \bibinfo{person}{Angjoo Kanazawa}.} \bibinfo{year}{2023}\natexlab{}.
\newblock \showarticletitle{Instruct-NeRF2NeRF: Editing 3D Scenes with Instructions}. In \bibinfo{booktitle}{\emph{Proceedings of the IEEE/CVF International Conference on Computer Vision}}.
\newblock


\bibitem[Hasinoff et~al\mbox{.}(2016)]%
        {hasinoff2016burst}
\bibfield{author}{\bibinfo{person}{Samuel~W Hasinoff}, \bibinfo{person}{Dillon Sharlet}, \bibinfo{person}{Ryan Geiss}, \bibinfo{person}{Andrew Adams}, \bibinfo{person}{Jonathan~T Barron}, \bibinfo{person}{Florian Kainz}, \bibinfo{person}{Jiawen Chen}, {and} \bibinfo{person}{Marc Levoy}.} \bibinfo{year}{2016}\natexlab{}.
\newblock \showarticletitle{Burst photography for high dynamic range and low-light imaging on mobile cameras}.
\newblock \bibinfo{journal}{\emph{ACM Transactions on Graphics (ToG)}} \bibinfo{volume}{35}, \bibinfo{number}{6} (\bibinfo{year}{2016}), \bibinfo{pages}{1--12}.
\newblock


\bibitem[He et~al\mbox{.}(2012)]%
        {he2012guided}
\bibfield{author}{\bibinfo{person}{Kaiming He}, \bibinfo{person}{Jian Sun}, {and} \bibinfo{person}{Xiaoou Tang}.} \bibinfo{year}{2012}\natexlab{}.
\newblock \showarticletitle{Guided image filtering}.
\newblock \bibinfo{journal}{\emph{IEEE transactions on pattern analysis and machine intelligence}} \bibinfo{volume}{35}, \bibinfo{number}{6} (\bibinfo{year}{2012}), \bibinfo{pages}{1397--1409}.
\newblock


\bibitem[Huang et~al\mbox{.}(2022)]%
        {huang2022hdr}
\bibfield{author}{\bibinfo{person}{Xin Huang}, \bibinfo{person}{Qi Zhang}, \bibinfo{person}{Ying Feng}, \bibinfo{person}{Hongdong Li}, \bibinfo{person}{Xuan Wang}, {and} \bibinfo{person}{Qing Wang}.} \bibinfo{year}{2022}\natexlab{}.
\newblock \showarticletitle{Hdr-nerf: High dynamic range neural radiance fields}. In \bibinfo{booktitle}{\emph{Proceedings of the IEEE/CVF Conference on Computer Vision and Pattern Recognition}}. \bibinfo{pages}{18398--18408}.
\newblock


\bibitem[Jang and Agapito(2021)]%
        {jang2021codenerf}
\bibfield{author}{\bibinfo{person}{Wonbong Jang} {and} \bibinfo{person}{Lourdes Agapito}.} \bibinfo{year}{2021}\natexlab{}.
\newblock \showarticletitle{Codenerf: Disentangled neural radiance fields for object categories}. In \bibinfo{booktitle}{\emph{Proceedings of the IEEE/CVF International Conference on Computer Vision}}. \bibinfo{pages}{12949--12958}.
\newblock


\bibitem[Jin et~al\mbox{.}(2023)]%
        {jin2023tensoir}
\bibfield{author}{\bibinfo{person}{Haian Jin}, \bibinfo{person}{Isabella Liu}, \bibinfo{person}{Peijia Xu}, \bibinfo{person}{Xiaoshuai Zhang}, \bibinfo{person}{Songfang Han}, \bibinfo{person}{Sai Bi}, \bibinfo{person}{Xiaowei Zhou}, \bibinfo{person}{Zexiang Xu}, {and} \bibinfo{person}{Hao Su}.} \bibinfo{year}{2023}\natexlab{}.
\newblock \showarticletitle{TensoIR: Tensorial Inverse Rendering}. In \bibinfo{booktitle}{\emph{Proceedings of the IEEE/CVF Conference on Computer Vision and Pattern Recognition}}. \bibinfo{pages}{165--174}.
\newblock


\bibitem[Kingma and Ba(2015)]%
        {KingBa15}
\bibfield{author}{\bibinfo{person}{Diederik Kingma} {and} \bibinfo{person}{Jimmy Ba}.} \bibinfo{year}{2015}\natexlab{}.
\newblock \showarticletitle{Adam: A Method for Stochastic Optimization}. In \bibinfo{booktitle}{\emph{International Conference on Learning Representations (ICLR)}}. \bibinfo{address}{San Diega, CA, USA}.
\newblock


\bibitem[Kobayashi et~al\mbox{.}(2022)]%
        {kobayashi2022decomposing}
\bibfield{author}{\bibinfo{person}{Sosuke Kobayashi}, \bibinfo{person}{Eiichi Matsumoto}, {and} \bibinfo{person}{Vincent Sitzmann}.} \bibinfo{year}{2022}\natexlab{}.
\newblock \showarticletitle{Decomposing nerf for editing via feature field distillation}.
\newblock \bibinfo{journal}{\emph{Advances in Neural Information Processing Systems}}  \bibinfo{volume}{35} (\bibinfo{year}{2022}), \bibinfo{pages}{23311--23330}.
\newblock


\bibitem[Kolda and Bader(2009)]%
        {kolda2009tensor}
\bibfield{author}{\bibinfo{person}{Tamara~G Kolda} {and} \bibinfo{person}{Brett~W Bader}.} \bibinfo{year}{2009}\natexlab{}.
\newblock \showarticletitle{Tensor decompositions and applications}.
\newblock \bibinfo{journal}{\emph{SIAM review}} \bibinfo{volume}{51}, \bibinfo{number}{3} (\bibinfo{year}{2009}), \bibinfo{pages}{455--500}.
\newblock


\bibitem[Kopf et~al\mbox{.}(2007)]%
        {kopf2007joint}
\bibfield{author}{\bibinfo{person}{Johannes Kopf}, \bibinfo{person}{Michael~F Cohen}, \bibinfo{person}{Dani Lischinski}, {and} \bibinfo{person}{Matt Uyttendaele}.} \bibinfo{year}{2007}\natexlab{}.
\newblock \showarticletitle{Joint bilateral upsampling}.
\newblock \bibinfo{journal}{\emph{ACM Transactions on Graphics (ToG)}} \bibinfo{volume}{26}, \bibinfo{number}{3} (\bibinfo{year}{2007}), \bibinfo{pages}{96--es}.
\newblock


\bibitem[Kuang et~al\mbox{.}(2023)]%
        {kuang2023palettenerf}
\bibfield{author}{\bibinfo{person}{Zhengfei Kuang}, \bibinfo{person}{Fujun Luan}, \bibinfo{person}{Sai Bi}, \bibinfo{person}{Zhixin Shu}, \bibinfo{person}{Gordon Wetzstein}, {and} \bibinfo{person}{Kalyan Sunkavalli}.} \bibinfo{year}{2023}\natexlab{}.
\newblock \showarticletitle{Palettenerf: Palette-based appearance editing of neural radiance fields}. In \bibinfo{booktitle}{\emph{Proceedings of the IEEE/CVF Conference on Computer Vision and Pattern Recognition}}. \bibinfo{pages}{20691--20700}.
\newblock


\bibitem[Li et~al\mbox{.}(2022)]%
        {li2022languagedriven}
\bibfield{author}{\bibinfo{person}{Boyi Li}, \bibinfo{person}{Kilian~Q Weinberger}, \bibinfo{person}{Serge Belongie}, \bibinfo{person}{Vladlen Koltun}, {and} \bibinfo{person}{Rene Ranftl}.} \bibinfo{year}{2022}\natexlab{}.
\newblock \showarticletitle{Language-driven Semantic Segmentation}. In \bibinfo{booktitle}{\emph{International Conference on Learning Representations}}.
\newblock
\urldef\tempurl%
\url{https://openreview.net/forum?id=RriDjddCLN}
\showURL{%
\tempurl}


\bibitem[Ling et~al\mbox{.}(2024)]%
        {ling2023dl3dv}
\bibfield{author}{\bibinfo{person}{Lu Ling}, \bibinfo{person}{Yichen Sheng}, \bibinfo{person}{Zhi Tu}, \bibinfo{person}{Wentian Zhao}, \bibinfo{person}{Cheng Xin}, \bibinfo{person}{Kun Wan}, \bibinfo{person}{Lantao Yu}, \bibinfo{person}{Qianyu Guo}, \bibinfo{person}{Zixun Yu}, \bibinfo{person}{Yawen Lu}, {et~al\mbox{.}}} \bibinfo{year}{2024}\natexlab{}.
\newblock \showarticletitle{DL3DV-10K: A Large-Scale Scene Dataset for Deep Learning-based 3D Vision}. In \bibinfo{booktitle}{\emph{Proceedings of the IEEE/CVF conference on computer vision and pattern recognition}}.
\newblock


\bibitem[Liu et~al\mbox{.}(2023)]%
        {liu2023stylerf}
\bibfield{author}{\bibinfo{person}{Kunhao Liu}, \bibinfo{person}{Fangneng Zhan}, \bibinfo{person}{Yiwen Chen}, \bibinfo{person}{Jiahui Zhang}, \bibinfo{person}{Yingchen Yu}, \bibinfo{person}{Abdulmotaleb El~Saddik}, \bibinfo{person}{Shijian Lu}, {and} \bibinfo{person}{Eric~P Xing}.} \bibinfo{year}{2023}\natexlab{}.
\newblock \showarticletitle{StyleRF: Zero-shot 3D Style Transfer of Neural Radiance Fields}. In \bibinfo{booktitle}{\emph{Proceedings of the IEEE/CVF Conference on Computer Vision and Pattern Recognition}}. \bibinfo{pages}{8338--8348}.
\newblock


\bibitem[Liu et~al\mbox{.}(2021)]%
        {liu2021editing}
\bibfield{author}{\bibinfo{person}{Steven Liu}, \bibinfo{person}{Xiuming Zhang}, \bibinfo{person}{Zhoutong Zhang}, \bibinfo{person}{Richard Zhang}, \bibinfo{person}{Jun-Yan Zhu}, {and} \bibinfo{person}{Bryan Russell}.} \bibinfo{year}{2021}\natexlab{}.
\newblock \showarticletitle{Editing conditional radiance fields}. In \bibinfo{booktitle}{\emph{Proceedings of the IEEE/CVF international conference on computer vision}}. \bibinfo{pages}{5773--5783}.
\newblock


\bibitem[Martin-Brualla et~al\mbox{.}(2021)]%
        {martin2021nerf}
\bibfield{author}{\bibinfo{person}{Ricardo Martin-Brualla}, \bibinfo{person}{Noha Radwan}, \bibinfo{person}{Mehdi~SM Sajjadi}, \bibinfo{person}{Jonathan~T Barron}, \bibinfo{person}{Alexey Dosovitskiy}, {and} \bibinfo{person}{Daniel Duckworth}.} \bibinfo{year}{2021}\natexlab{}.
\newblock \showarticletitle{Nerf in the wild: Neural radiance fields for unconstrained photo collections}. In \bibinfo{booktitle}{\emph{Proceedings of the IEEE/CVF Conference on Computer Vision and Pattern Recognition}}. \bibinfo{pages}{7210--7219}.
\newblock


\bibitem[Mertens et~al\mbox{.}(2007)]%
        {mertens2007exposure}
\bibfield{author}{\bibinfo{person}{Tom Mertens}, \bibinfo{person}{Jan Kautz}, {and} \bibinfo{person}{Frank Van~Reeth}.} \bibinfo{year}{2007}\natexlab{}.
\newblock \showarticletitle{Exposure fusion}. In \bibinfo{booktitle}{\emph{15th Pacific Conference on Computer Graphics and Applications (PG'07)}}. IEEE, \bibinfo{pages}{382--390}.
\newblock


\bibitem[Mildenhall et~al\mbox{.}(2022)]%
        {mildenhall2022nerf}
\bibfield{author}{\bibinfo{person}{Ben Mildenhall}, \bibinfo{person}{Peter Hedman}, \bibinfo{person}{Ricardo Martin-Brualla}, \bibinfo{person}{Pratul~P Srinivasan}, {and} \bibinfo{person}{Jonathan~T Barron}.} \bibinfo{year}{2022}\natexlab{}.
\newblock \showarticletitle{Nerf in the dark: High dynamic range view synthesis from noisy raw images}. In \bibinfo{booktitle}{\emph{Proceedings of the IEEE/CVF Conference on Computer Vision and Pattern Recognition}}. \bibinfo{pages}{16190--16199}.
\newblock


\bibitem[Mildenhall et~al\mbox{.}(2021)]%
        {mildenhall2021nerf}
\bibfield{author}{\bibinfo{person}{Ben Mildenhall}, \bibinfo{person}{Pratul~P Srinivasan}, \bibinfo{person}{Matthew Tancik}, \bibinfo{person}{Jonathan~T Barron}, \bibinfo{person}{Ravi Ramamoorthi}, {and} \bibinfo{person}{Ren Ng}.} \bibinfo{year}{2021}\natexlab{}.
\newblock \showarticletitle{Nerf: Representing scenes as neural radiance fields for view synthesis}.
\newblock \bibinfo{journal}{\emph{Commun. ACM}} \bibinfo{volume}{65}, \bibinfo{number}{1} (\bibinfo{year}{2021}), \bibinfo{pages}{99--106}.
\newblock


\bibitem[M{\"u}ller et~al\mbox{.}(2022)]%
        {muller2022instant}
\bibfield{author}{\bibinfo{person}{Thomas M{\"u}ller}, \bibinfo{person}{Alex Evans}, \bibinfo{person}{Christoph Schied}, {and} \bibinfo{person}{Alexander Keller}.} \bibinfo{year}{2022}\natexlab{}.
\newblock \showarticletitle{Instant neural graphics primitives with a multiresolution hash encoding}.
\newblock \bibinfo{journal}{\emph{ACM Transactions on Graphics (ToG)}} \bibinfo{volume}{41}, \bibinfo{number}{4} (\bibinfo{year}{2022}), \bibinfo{pages}{1--15}.
\newblock


\bibitem[Munkberg et~al\mbox{.}(2022)]%
        {munkberg2022extracting}
\bibfield{author}{\bibinfo{person}{Jacob Munkberg}, \bibinfo{person}{Jon Hasselgren}, \bibinfo{person}{Tianchang Shen}, \bibinfo{person}{Jun Gao}, \bibinfo{person}{Wenzheng Chen}, \bibinfo{person}{Alex Evans}, \bibinfo{person}{Thomas M{\"u}ller}, {and} \bibinfo{person}{Sanja Fidler}.} \bibinfo{year}{2022}\natexlab{}.
\newblock \showarticletitle{Extracting triangular 3d models, materials, and lighting from images}. In \bibinfo{booktitle}{\emph{Proceedings of the IEEE/CVF Conference on Computer Vision and Pattern Recognition}}. \bibinfo{pages}{8280--8290}.
\newblock


\bibitem[Nam et~al\mbox{.}(2022)]%
        {nam2022learning}
\bibfield{author}{\bibinfo{person}{Seonghyeon Nam}, \bibinfo{person}{Abhijith Punnappurath}, \bibinfo{person}{Marcus~A Brubaker}, {and} \bibinfo{person}{Michael~S Brown}.} \bibinfo{year}{2022}\natexlab{}.
\newblock \showarticletitle{Learning srgb-to-raw-rgb de-rendering with content-aware metadata}. In \bibinfo{booktitle}{\emph{Proceedings of the IEEE/CVF Conference on Computer Vision and Pattern Recognition}}. \bibinfo{pages}{17704--17713}.
\newblock


\bibitem[Nishimura et~al\mbox{.}(2018)]%
        {nishimura2018automatic}
\bibfield{author}{\bibinfo{person}{Jun Nishimura}, \bibinfo{person}{Timo Gerasimow}, \bibinfo{person}{Rao Sushma}, \bibinfo{person}{Aleksandar Sutic}, \bibinfo{person}{Chyuan-Tyng Wu}, {and} \bibinfo{person}{Gilad Michael}.} \bibinfo{year}{2018}\natexlab{}.
\newblock \showarticletitle{Automatic ISP image quality tuning using nonlinear optimization}. In \bibinfo{booktitle}{\emph{2018 25th IEEE International Conference on Image Processing (ICIP)}}. IEEE, \bibinfo{pages}{2471--2475}.
\newblock


\bibitem[Paris et~al\mbox{.}(2011)]%
        {paris2011local}
\bibfield{author}{\bibinfo{person}{Sylvain Paris}, \bibinfo{person}{Samuel~W Hasinoff}, {and} \bibinfo{person}{Jan Kautz}.} \bibinfo{year}{2011}\natexlab{}.
\newblock \showarticletitle{Local Laplacian filters: Edge-aware image processing with a Laplacian pyramid.}
\newblock \bibinfo{journal}{\emph{ACM Trans. Graph.}} \bibinfo{volume}{30}, \bibinfo{number}{4} (\bibinfo{year}{2011}), \bibinfo{pages}{68}.
\newblock


\bibitem[Park et~al\mbox{.}(2021)]%
        {park2021nerfies}
\bibfield{author}{\bibinfo{person}{Keunhong Park}, \bibinfo{person}{Utkarsh Sinha}, \bibinfo{person}{Jonathan~T Barron}, \bibinfo{person}{Sofien Bouaziz}, \bibinfo{person}{Dan~B Goldman}, \bibinfo{person}{Steven~M Seitz}, {and} \bibinfo{person}{Ricardo Martin-Brualla}.} \bibinfo{year}{2021}\natexlab{}.
\newblock \showarticletitle{Nerfies: Deformable neural radiance fields}. In \bibinfo{booktitle}{\emph{Proceedings of the IEEE/CVF International Conference on Computer Vision}}. \bibinfo{pages}{5865--5874}.
\newblock


\bibitem[P{\'e}rez et~al\mbox{.}(2023)]%
        {perez2023poisson}
\bibfield{author}{\bibinfo{person}{Patrick P{\'e}rez}, \bibinfo{person}{Michel Gangnet}, {and} \bibinfo{person}{Andrew Blake}.} \bibinfo{year}{2023}\natexlab{}.
\newblock \showarticletitle{Poisson image editing}.
\newblock In \bibinfo{booktitle}{\emph{Seminal Graphics Papers: Pushing the Boundaries, Volume 2}}. \bibinfo{pages}{577--582}.
\newblock


\bibitem[Radford et~al\mbox{.}(2021)]%
        {radford2021learning}
\bibfield{author}{\bibinfo{person}{Alec Radford}, \bibinfo{person}{Jong~Wook Kim}, \bibinfo{person}{Chris Hallacy}, \bibinfo{person}{Aditya Ramesh}, \bibinfo{person}{Gabriel Goh}, \bibinfo{person}{Sandhini Agarwal}, \bibinfo{person}{Girish Sastry}, \bibinfo{person}{Amanda Askell}, \bibinfo{person}{Pamela Mishkin}, \bibinfo{person}{Jack Clark}, {et~al\mbox{.}}} \bibinfo{year}{2021}\natexlab{}.
\newblock \showarticletitle{Learning transferable visual models from natural language supervision}. In \bibinfo{booktitle}{\emph{International conference on machine learning}}. PMLR, \bibinfo{pages}{8748--8763}.
\newblock


\bibitem[Sabour et~al\mbox{.}(2023)]%
        {sabour2023robustnerf}
\bibfield{author}{\bibinfo{person}{Sara Sabour}, \bibinfo{person}{Suhani Vora}, \bibinfo{person}{Daniel Duckworth}, \bibinfo{person}{Ivan Krasin}, \bibinfo{person}{David~J Fleet}, {and} \bibinfo{person}{Andrea Tagliasacchi}.} \bibinfo{year}{2023}\natexlab{}.
\newblock \showarticletitle{RobustNeRF: Ignoring Distractors with Robust Losses}. In \bibinfo{booktitle}{\emph{Proceedings of the IEEE/CVF Conference on Computer Vision and Pattern Recognition}}. \bibinfo{pages}{20626--20636}.
\newblock


\bibitem[Schonberger and Frahm(2016)]%
        {schonberger2016structure}
\bibfield{author}{\bibinfo{person}{Johannes~L Schonberger} {and} \bibinfo{person}{Jan-Michael Frahm}.} \bibinfo{year}{2016}\natexlab{}.
\newblock \showarticletitle{Structure-from-motion revisited}. In \bibinfo{booktitle}{\emph{Proceedings of the IEEE conference on computer vision and pattern recognition}}. \bibinfo{pages}{4104--4113}.
\newblock


\bibitem[Srinivasan et~al\mbox{.}(2021)]%
        {srinivasan2021nerv}
\bibfield{author}{\bibinfo{person}{Pratul~P Srinivasan}, \bibinfo{person}{Boyang Deng}, \bibinfo{person}{Xiuming Zhang}, \bibinfo{person}{Matthew Tancik}, \bibinfo{person}{Ben Mildenhall}, {and} \bibinfo{person}{Jonathan~T Barron}.} \bibinfo{year}{2021}\natexlab{}.
\newblock \showarticletitle{Nerv: Neural reflectance and visibility fields for relighting and view synthesis}. In \bibinfo{booktitle}{\emph{Proceedings of the IEEE/CVF Conference on Computer Vision and Pattern Recognition}}. \bibinfo{pages}{7495--7504}.
\newblock


\bibitem[Tancik et~al\mbox{.}(2022)]%
        {tancik2022block}
\bibfield{author}{\bibinfo{person}{Matthew Tancik}, \bibinfo{person}{Vincent Casser}, \bibinfo{person}{Xinchen Yan}, \bibinfo{person}{Sabeek Pradhan}, \bibinfo{person}{Ben Mildenhall}, \bibinfo{person}{Pratul~P Srinivasan}, \bibinfo{person}{Jonathan~T Barron}, {and} \bibinfo{person}{Henrik Kretzschmar}.} \bibinfo{year}{2022}\natexlab{}.
\newblock \showarticletitle{Block-nerf: Scalable large scene neural view synthesis}. In \bibinfo{booktitle}{\emph{Proceedings of the IEEE/CVF Conference on Computer Vision and Pattern Recognition}}. \bibinfo{pages}{8248--8258}.
\newblock


\bibitem[Tancik et~al\mbox{.}(2020)]%
        {tancik2020fourier}
\bibfield{author}{\bibinfo{person}{Matthew Tancik}, \bibinfo{person}{Pratul Srinivasan}, \bibinfo{person}{Ben Mildenhall}, \bibinfo{person}{Sara Fridovich-Keil}, \bibinfo{person}{Nithin Raghavan}, \bibinfo{person}{Utkarsh Singhal}, \bibinfo{person}{Ravi Ramamoorthi}, \bibinfo{person}{Jonathan Barron}, {and} \bibinfo{person}{Ren Ng}.} \bibinfo{year}{2020}\natexlab{}.
\newblock \showarticletitle{Fourier features let networks learn high frequency functions in low dimensional domains}.
\newblock \bibinfo{journal}{\emph{Advances in Neural Information Processing Systems}}  \bibinfo{volume}{33} (\bibinfo{year}{2020}), \bibinfo{pages}{7537--7547}.
\newblock


\bibitem[Tomasi and Manduchi(1998)]%
        {tomasi1998bilateral}
\bibfield{author}{\bibinfo{person}{Carlo Tomasi} {and} \bibinfo{person}{Roberto Manduchi}.} \bibinfo{year}{1998}\natexlab{}.
\newblock \showarticletitle{Bilateral filtering for gray and color images}. In \bibinfo{booktitle}{\emph{Sixth international conference on computer vision (IEEE Cat. No. 98CH36271)}}. IEEE, \bibinfo{pages}{839--846}.
\newblock


\bibitem[Tseng et~al\mbox{.}(2019)]%
        {tseng2019hyperparameter}
\bibfield{author}{\bibinfo{person}{Ethan Tseng}, \bibinfo{person}{Felix Yu}, \bibinfo{person}{Yuting Yang}, \bibinfo{person}{Fahim Mannan}, \bibinfo{person}{Karl~ST Arnaud}, \bibinfo{person}{Derek Nowrouzezahrai}, \bibinfo{person}{Jean-Fran{\c{c}}ois Lalonde}, {and} \bibinfo{person}{Felix Heide}.} \bibinfo{year}{2019}\natexlab{}.
\newblock \showarticletitle{Hyperparameter optimization in black-box image processing using differentiable proxies.}
\newblock \bibinfo{journal}{\emph{ACM Trans. Graph.}} \bibinfo{volume}{38}, \bibinfo{number}{4} (\bibinfo{year}{2019}), \bibinfo{pages}{27--1}.
\newblock


\bibitem[Tseng et~al\mbox{.}(2022)]%
        {tseng2022neural}
\bibfield{author}{\bibinfo{person}{Ethan Tseng}, \bibinfo{person}{Yuxuan Zhang}, \bibinfo{person}{Lars Jebe}, \bibinfo{person}{Xuaner Zhang}, \bibinfo{person}{Zhihao Xia}, \bibinfo{person}{Yifei Fan}, \bibinfo{person}{Felix Heide}, {and} \bibinfo{person}{Jiawen Chen}.} \bibinfo{year}{2022}\natexlab{}.
\newblock \showarticletitle{Neural photo-finishing}.
\newblock \bibinfo{journal}{\emph{ACM Transactions on Graphics (TOG)}} \bibinfo{volume}{41}, \bibinfo{number}{6} (\bibinfo{year}{2022}), \bibinfo{pages}{1--15}.
\newblock


\bibitem[Vaswani et~al\mbox{.}(2017)]%
        {vaswani2017attention}
\bibfield{author}{\bibinfo{person}{Ashish Vaswani}, \bibinfo{person}{Noam Shazeer}, \bibinfo{person}{Niki Parmar}, \bibinfo{person}{Jakob Uszkoreit}, \bibinfo{person}{Llion Jones}, \bibinfo{person}{Aidan~N Gomez}, \bibinfo{person}{{\L}ukasz Kaiser}, {and} \bibinfo{person}{Illia Polosukhin}.} \bibinfo{year}{2017}\natexlab{}.
\newblock \showarticletitle{Attention is all you need}.
\newblock \bibinfo{journal}{\emph{Advances in neural information processing systems}}  \bibinfo{volume}{30} (\bibinfo{year}{2017}).
\newblock


\bibitem[Verbin et~al\mbox{.}(2022)]%
        {verbin2022ref}
\bibfield{author}{\bibinfo{person}{Dor Verbin}, \bibinfo{person}{Peter Hedman}, \bibinfo{person}{Ben Mildenhall}, \bibinfo{person}{Todd Zickler}, \bibinfo{person}{Jonathan~T Barron}, {and} \bibinfo{person}{Pratul~P Srinivasan}.} \bibinfo{year}{2022}\natexlab{}.
\newblock \showarticletitle{Ref-nerf: Structured view-dependent appearance for neural radiance fields}. In \bibinfo{booktitle}{\emph{2022 IEEE/CVF Conference on Computer Vision and Pattern Recognition (CVPR)}}. IEEE, \bibinfo{pages}{5481--5490}.
\newblock


\bibitem[Wang et~al\mbox{.}(2022)]%
        {wang2022clip}
\bibfield{author}{\bibinfo{person}{Can Wang}, \bibinfo{person}{Menglei Chai}, \bibinfo{person}{Mingming He}, \bibinfo{person}{Dongdong Chen}, {and} \bibinfo{person}{Jing Liao}.} \bibinfo{year}{2022}\natexlab{}.
\newblock \showarticletitle{Clip-nerf: Text-and-image driven manipulation of neural radiance fields}. In \bibinfo{booktitle}{\emph{Proceedings of the IEEE/CVF Conference on Computer Vision and Pattern Recognition}}. \bibinfo{pages}{3835--3844}.
\newblock


\bibitem[Wang et~al\mbox{.}(2023a)]%
        {wang2023nerf}
\bibfield{author}{\bibinfo{person}{Can Wang}, \bibinfo{person}{Ruixiang Jiang}, \bibinfo{person}{Menglei Chai}, \bibinfo{person}{Mingming He}, \bibinfo{person}{Dongdong Chen}, {and} \bibinfo{person}{Jing Liao}.} \bibinfo{year}{2023}\natexlab{a}.
\newblock \showarticletitle{Nerf-art: Text-driven neural radiance fields stylization}.
\newblock \bibinfo{journal}{\emph{IEEE Transactions on Visualization and Computer Graphics}} (\bibinfo{year}{2023}).
\newblock


\bibitem[Wang et~al\mbox{.}(2023b)]%
        {wang2023implicit}
\bibfield{author}{\bibinfo{person}{Chao Wang}, \bibinfo{person}{Ana Serrano}, \bibinfo{person}{Xingang Pan}, \bibinfo{person}{Krzysztof Wolski}, \bibinfo{person}{Bin Chen}, \bibinfo{person}{Karol Myszkowski}, \bibinfo{person}{Hans-Peter Seidel}, \bibinfo{person}{Christian Theobalt}, {and} \bibinfo{person}{Thomas Leimk{\"u}hler}.} \bibinfo{year}{2023}\natexlab{b}.
\newblock \showarticletitle{An Implicit Neural Representation for the Image Stack: Depth, All in Focus, and High Dynamic Range}.
\newblock \bibinfo{journal}{\emph{ACM Transactions on Graphics (TOG)}} \bibinfo{volume}{42}, \bibinfo{number}{6} (\bibinfo{year}{2023}), \bibinfo{pages}{1--11}.
\newblock


\bibitem[Wang et~al\mbox{.}(2023c)]%
        {wang2023lighting}
\bibfield{author}{\bibinfo{person}{Haoyuan Wang}, \bibinfo{person}{Xiaogang Xu}, \bibinfo{person}{Ke Xu}, {and} \bibinfo{person}{Rynson~W.H. Lau}.} \bibinfo{year}{2023}\natexlab{c}.
\newblock \showarticletitle{Lighting up NeRF via Unsupervised Decomposition and Enhancement}. In \bibinfo{booktitle}{\emph{ICCV}}.
\newblock


\bibitem[Yu et~al\mbox{.}(2021)]%
        {yu2021reconfigisp}
\bibfield{author}{\bibinfo{person}{Ke Yu}, \bibinfo{person}{Zexian Li}, \bibinfo{person}{Yue Peng}, \bibinfo{person}{Chen~Change Loy}, {and} \bibinfo{person}{Jinwei Gu}.} \bibinfo{year}{2021}\natexlab{}.
\newblock \showarticletitle{Reconfigisp: Reconfigurable camera image processing pipeline}. In \bibinfo{booktitle}{\emph{Proceedings of the IEEE/CVF International Conference on Computer Vision}}. \bibinfo{pages}{4248--4257}.
\newblock


\bibitem[Zamir et~al\mbox{.}(2022)]%
        {zamir2022restormer}
\bibfield{author}{\bibinfo{person}{Syed~Waqas Zamir}, \bibinfo{person}{Aditya Arora}, \bibinfo{person}{Salman Khan}, \bibinfo{person}{Munawar Hayat}, \bibinfo{person}{Fahad~Shahbaz Khan}, {and} \bibinfo{person}{Ming-Hsuan Yang}.} \bibinfo{year}{2022}\natexlab{}.
\newblock \showarticletitle{Restormer: Efficient transformer for high-resolution image restoration}. In \bibinfo{booktitle}{\emph{Proceedings of the IEEE/CVF conference on computer vision and pattern recognition}}. \bibinfo{pages}{5728--5739}.
\newblock


\bibitem[Zhang et~al\mbox{.}(2022)]%
        {zhang2022arf}
\bibfield{author}{\bibinfo{person}{Kai Zhang}, \bibinfo{person}{Nick Kolkin}, \bibinfo{person}{Sai Bi}, \bibinfo{person}{Fujun Luan}, \bibinfo{person}{Zexiang Xu}, \bibinfo{person}{Eli Shechtman}, {and} \bibinfo{person}{Noah Snavely}.} \bibinfo{year}{2022}\natexlab{}.
\newblock \showarticletitle{Arf: Artistic radiance fields}. In \bibinfo{booktitle}{\emph{European Conference on Computer Vision}}. Springer, \bibinfo{pages}{717--733}.
\newblock


\bibitem[Zhang et~al\mbox{.}(2021)]%
        {zhang2021nerfactor}
\bibfield{author}{\bibinfo{person}{Xiuming Zhang}, \bibinfo{person}{Pratul~P Srinivasan}, \bibinfo{person}{Boyang Deng}, \bibinfo{person}{Paul Debevec}, \bibinfo{person}{William~T Freeman}, {and} \bibinfo{person}{Jonathan~T Barron}.} \bibinfo{year}{2021}\natexlab{}.
\newblock \showarticletitle{Nerfactor: Neural factorization of shape and reflectance under an unknown illumination}.
\newblock \bibinfo{journal}{\emph{ACM Transactions on Graphics (ToG)}} \bibinfo{volume}{40}, \bibinfo{number}{6} (\bibinfo{year}{2021}), \bibinfo{pages}{1--18}.
\newblock


\end{thebibliography}
